\documentclass{article}

\usepackage{microtype}
\usepackage{graphicx}
\usepackage{subcaption}
\usepackage{booktabs} 

\usepackage{hyperref}

 \usepackage[preprint]{icml2026}

\usepackage{amsmath}
\usepackage{amssymb}
\usepackage{mathtools}
\usepackage{amsthm}

\usepackage[capitalize,noabbrev]{cleveref}

\usepackage{tikz}
\usepackage{enumitem}
\usepackage{float}
\usepackage{tabularx}  
\newcolumntype{Y}{>{\centering\arraybackslash}X}
\usepackage{nicefrac}       
\usepackage{natbib}

\newcommand{\tr}{\mathrm{Tr}}
\newcommand{\jac}{\mathrm{Jac}}

\theoremstyle{plain}
\newtheorem{theorem}{Theorem}[section]

\theoremstyle{definition}
\newtheorem{definition}[theorem]{Definition}

\theoremstyle{remark}

\usepackage[textsize=tiny]{todonotes}

\icmltitlerunning{Metric Graph Kernels via the Tropical Torelli Map}

\begin{document}

\twocolumn[
  \icmltitle{Metric Graph Kernels via the Tropical Torelli Map}



  \icmlsetsymbol{equal}{*}

  \begin{icmlauthorlist}
    \icmlauthor{Yueqi Cao}{kth}
    \icmlauthor{Anthea Monod}{ic}
  \end{icmlauthorlist}

  \icmlaffiliation{kth}{Department of Mathematics, KTH Royal Institute of Technology, Stockholm, Sweden}
  \icmlaffiliation{ic}{Department of Mathematics, Imperial College London, London, UK}

  \icmlcorrespondingauthor{Yueqi Cao}{yueqic@kth.se}
  \icmlcorrespondingauthor{Anthea Monod}{a.monod@imperial.ac.uk}

  \icmlkeywords{Graph Kernels, Metric Graphs, Geometric Learning, Kernel Methods}

  \vskip 0.3in
]



\printAffiliationsAndNotice{}  

\begin{abstract}
We introduce the first graph kernels for metric graphs via tropical algebraic geometry. In contrast to conventional graph kernels based on graph combinatorics such as nodes, edges, and subgraphs, our metric graph kernels are purely based on the geometry and topology of the underlying metric space. A key characterizing property of our construction is its invariance under edge subdivision, making the kernels intrinsically well-suited for comparing graphs representing different underlying metric spaces. We develop efficient algorithms to compute our kernels and analyze their complexity, which depends primarily on the genus of the input graphs rather than their size. Through experiments on synthetic data and selected real-world datasets, we demonstrate that our kernels capture complementary geometric and topological information overseen by standard combinatorial approaches, particularly in label-free settings. We further showcase their practical utility with an urban road network classification task.
\end{abstract}

\section{Introduction}
Graph kernels are powerful tools for measuring similarities between graphs and networks and have become essential in many machine learning tasks. Traditional graph kernels typically rely on combinatorial aspects of graphs, such as nodes, edges, and subgraphs, with additional information from node and edge labels and attributes \citep{kriege2020survey,vishwanathan2010graph,borgwardt2020graph,nikolentzos2021graph,neumann2016propagation}. While such kernels have demonstrated success in many domains, they can be limited in scenarios where graphs arise from geometric data, and where labels or attributes are absent or incomparable. Such challenges are intrinsic to \emph{metric graphs} and render existing graph kernels inapplicable in this setting.  In this paper, we introduce the first kernel framework designed specifically for metric graphs.

Metric graphs are geometric realizations of graphs with a length function on their edges.  To develop our metric graph kernels, we leverage the known one-to-one correspondence between metric graphs and \emph{abstract tropical curves} from tropical algebraic geometry \citep{chan2021moduli,cao2025computingtropicalabeljacobitransform}.  This correspondence has been studied in tropical geometry via the \emph{tropical Torelli map} \citep{chan2012combinatorics,brannetti2011tropical}, which sends any metric graph to a flat torus.  In computational settings and especially machine learning, for a weighted graph with generic length function, we can compute a unique symmetric positive definite (SPD) matrix to represent the flat torus. Using the tropical Torelli map, we can effectively identify weighted graphs with SPD matrices, which contain the intrinsic geometric and topological information on the underlying metric graph. 

The space of SPD matrices is central in information geometry and can be endowed with various Riemannian metrics to capture its rich geometric structure  \citep{luo2021geometric,zhang2019new,thanwerdas2023n}. However, it is only applicable when the metric graphs are of the same genus. In practice, to compare metric graphs of different genus via the tropical Torelli map, we must embed the SPD matrices into a common ambient space, namely the space of \emph{positive semi-definite (PSD) matrices}. Motivated by the geometric structure of the tropical Jacobian, we seek a distance on the PSD space that is invariant under orthogonal transformations, which turns out to coincide with the well-known \emph{Bures--Wasserstein distance} from  information geometry \citep{bhatia2019bures,thanwerdas2023bures}. Leveraging this distance, we propose the \emph{tropical Torelli--Wasserstein (TTW) kernel} for metric graphs.  For simplicity and efficient computation, we also propose the \emph{tropical Torelli--Euclidean (TTE) kernel} based on the Euclidean distance. Our graph kernels are purely based on the geometry and topology of the underlying metric graphs, thus are well-suited for comparing graphs that represent different underlying spaces. 

\textbf{Contributions.} Our work is the first to propose kernels for metric graphs, which we achieve by merging tropical geometry and information geometry in the context of machine learning.  Our main contributions are the following:
\begin{itemize}
    \item We propose new graph kernels for metric graphs, leveraging the tropical Torelli map from tropical geometry. Our graph kernels are invariant under edge subdivisions and naturally extend to well-defined kernels on the underlying space of metric graphs.
    \item We develop algorithms to compute our metric graph kernels. We show that the computational complexity is dominated by graph \emph{genus}, a topological invariant that measures the number of independent cycles, rather than the number of nodes or edges. We give concrete analyses on different levels of graph sparsity.
    \item We empirically evaluate our proposed kernels on synthetic and real-world datasets to assess their behavior in label-free settings. Our experiments illustrate that our kernels capture geometric and topological information that is complementary to existing combinatorial graph kernels, and are particularly effective when node and edge labels are unavailable. 
    \item We further demonstrate the practical relevance of our approach through targeted case studies. On synthetic datasets, we validate the theoretical runtime analysis, while on real-world data we apply our kernels to an urban road network classification task. 
\end{itemize}

\section{From Graphs to Metric Graphs}\label{sec:metric-graph}
A graph consists of discrete sets of nodes and edges. A \emph{metric graph} is a 1-dimensional metric space that can be realized as the underlying space of a graph. Any graph-based quantity extends well to metric graphs if it is compatible with \emph{edge subdivisions}---a concept we will now introduce mathematically, which will then be motivated and illustrated concretely by road networks.

\textbf{Edge Subdivisions.} Let $G=(V,E)$ be a graph and $\ell:E\to\mathbb{R}_+$ be a length function. An \emph{edge subdivision} of $e=[u,v]\in E$ is the following operation on $G$: First, add a new node $w$ to $V$, and then replace $e$ by two edges $e'=[u,w]$ and $e''=[w,v]$ such that $\ell(e)=\ell(e')+\ell(e'')$. An edge subdivision increases both the number of nodes and number of edges by 1. $G'$ is called a \emph{refinement} of $G$, denoted as $G'\geq G$, if $G'$ can be obtained from $G$ by a sequence of edge subdivisions. By definition, all refinements of $G$ share the same underlying metric graph $|G|$.

Note that refinement introduces a new relation between graphs that differs from the usual notion of subgraph inclusion: If $G'$ is a refinement of $G$, then there exists an injection $V(G)\to V(G')$ and a surjection $E(G')\to E(G)$. However, if $G$ is a subgraph of $G'$, then $V(G)\to V(G')$ and $E(G)\to E(G')$ are both inclusions.

\textbf{Metric Graph Kernels.} Let $G_1, G_2$ be two graphs. A \emph{graph kernel} is a function whose value $k(x_1,x_2)$ quantifies the similarity between $G_1, G_2$, typically in terms of their combinatorial structure. For example, the Weisfeiler--Lehman kernel assigns a large value when $G_1$ and $G_2$ are highly similar from the perspective of graph isomorphism \citep{shervashidze2009efficient}. In contrast, the following definition of a \emph{metric graph kernel} aims to quantify similarity based on the underlying metric spaces. Specifically, the kernel $k(G_1,G_2)$ takes a large value when the metric realizations $|G_1|$ and $|G_2|$ are similar, for instance, from the perspective of isometry between metric spaces.

\begin{definition}
 A graph kernel $k$ is called a \emph{metric graph kernel} if $k(G_1',G_2')=k(G_1,G_2)$ for any refinements $G_1'\geq G_1$ and $G_2'\geq G_2$, that is, the kernel $k$ is invariant under graph refinements.   
\end{definition}

Existing graph kernels are often based on nodes, edges, subgraphs, and node/edge labels, which are not invariant under graph refinements. Therefore, they fail to be metric graph kernels. In particular, node/edge labels are not well-defined on metric graphs---instead, the appropriate analogue should be functions on metric spaces. Table~\ref{tab:notion} summarizes comparisons of key notions between graphs and metric graphs.

\begin{table}[htbp]
\caption{Key notions for graphs and their corresponding notions in the setting of metric graphs.}
\label{tab:notion}
  \centering
  \small
  \begin{tabularx}{\linewidth}{|X|X|}
\toprule
 \textbf{Graph} & \textbf{Metric Graph} \\
\midrule
Vertex/Node & Point\\
Edge & Line segment\\
Subgraph & Subspace\\
Label/Attribute &  Function\\
Graph distance & Geodesic distance\\
Graph isomorphism & Metric isometry\\
\bottomrule
\end{tabularx}

\end{table}

\begin{figure}[htbp]
    \centering
    \includegraphics[width=\linewidth]{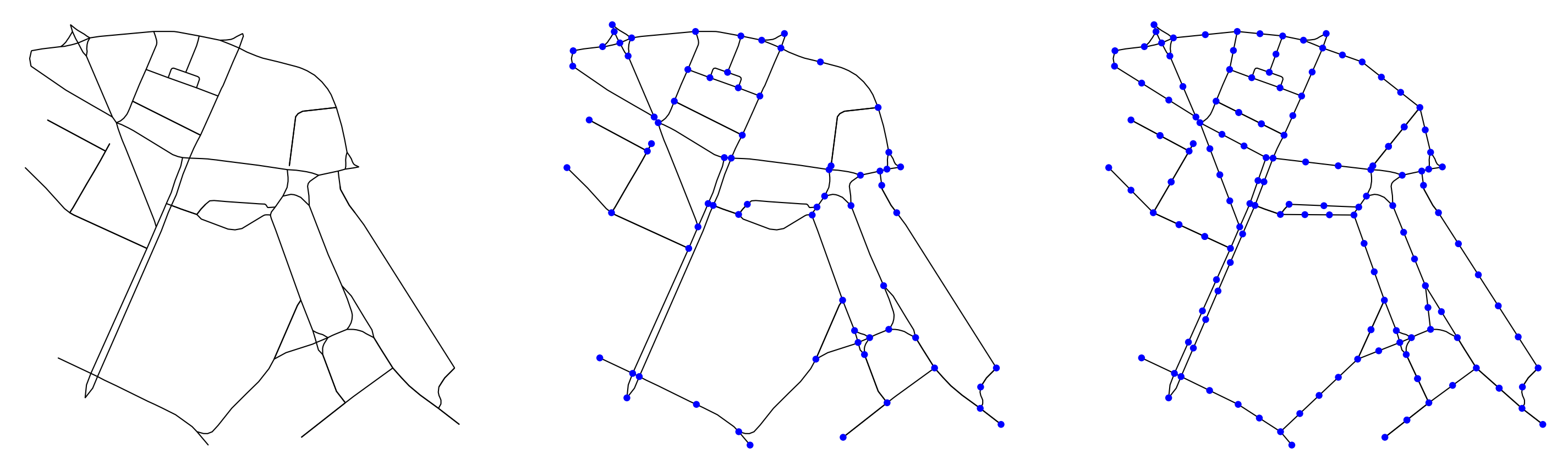}
    \caption{An illustration of a metric graph. The leftmost figure presents a local road network near the Victoria underground (tube) station in London. A graph representing the road network is given in the middle. The rightmost figure presents a refinement by adding artificial landmarks. }
    \label{fig:victoria}
\end{figure}

\textbf{Road Networks.} Typical real-world examples of metric graphs are road networks, which are often modeled using nodes to represent landmarks or intersections and edges to represent roads. However, the essential metric and topological properties of a road network should be independent of its combinatorial representation. For instance, refining a road network by inserting additional landmarks along existing roads should not alter the underlying geometry or structure of the network (Figure~\ref{fig:victoria}). 

\section{The Tropical Torelli Map}\label{sec:tropical-torelli}

In tropical geometry, metric graphs are known to be in one-to-one correspondence with \emph{abstract tropical curves} \citep{mikhalkin2008tropical}. The \emph{tropical Torelli map} is a fundamental construction that sends each tropical curve to a canonically defined flat torus, known as the \emph{tropical Jacobian variety}. It serves as a tropical analogue of the classical Torelli map for algebraic curves, which sends a curve to its Jacobian variety. Further background and details on algebraic and tropical geometry are given in Appendix~\ref{app:trop-torelli}.  

The tropical Torelli map is originally defined in an intrinsic, basis-free manner via moduli spaces of tropical curves and \emph{principally polarized Abelian varieties} \citep{brannetti2011tropical}. While this formulation is conceptually natural, it is not well suited for explicit computations. For computational purposes, it is necessary to instead express the map in a vector form by fixing bases for the relevant vector spaces and lattices of the tropical Jacobian varieties. This consideration motivates the main objective of this section.  We first introduce the \emph{tropical Torelli map for weighted graphs}; this is the map we use to send a graph to a matrix. For weighted graphs with generic length functions, we provide an algorithm for its practical computation, where the uniqueness of the resulting matrix representation is required but challenging to achieve. Under the generic assumption on length functions, our algorithm outputs a unique SPD matrix for any weighted graph. 


\subsection{Graph Homology}\label{sec:definition}

A concept that is central to our construction is the \emph{1-homology group} of a graph. The concept arises from algebraic topology and intuitively corresponds to cycles in a graph.  We now give mathematical details on the construction of the 1-homology group of a graph and how we obtain a matrix representation of the graph in terms of 1-cycles via the tropical Torelli map.

We first outline the 1-homology group as follows. Let $G=(V,E)$ be a connected graph with $n$ nodes and $m$ edges. Let $C_0(G;\mathbb{R})$ be the $n$-dimensional vector space spanned by $V$ and $C_1(G;\mathbb{R})$ be the $m$-dimensional vector space spanned by $E$. Fix an arbitrary orientation on $G$. The \emph{boundary map} $\partial: C_1(G;\mathbb{R})\to C_0(G;\mathbb{R})$ is a linear map given by $\partial([u,v])=v-u$. The \emph{1-homology group} is defined as the kernel of the boundary map $H_1(G;\mathbb{R})=\mathrm{ker}(\partial)$. Its dimension $g=\dim(H_1(G;\mathbb{R}))$ is called the \emph{genus} of $G$ and satisfies $g=m-n+1$. Any element $\sigma\in H_1(G;\mathbb{R})$ is called a \emph{1-cycle}. Let $\sigma(e)$ be the coefficient of $e$ in $\sigma$. The \emph{support} of a 1-cycle is the set $\{e\in E: \sigma(e)\neq 0\}$, which forms a cycle in the sense of subgraph.
 
Next, we describe how to represent the graph as a matrix in terms of 1-cycles. Let $\ell:E\to\mathbb{R}_+$ be a length function. Define an inner product $\mathcal{Q}_G$ on $C_1(G;\mathbb{R})$ by assigning $\mathcal{Q}_G(e_i,e_j) = \delta_{ij}\sqrt{\ell(e_i)\ell(e_j)}$
on the edges and extending bilinearly to the whole space $C_1(G;\mathbb{R})$. Notice that under the inner product $\mathcal{Q}_G$, an element $e_i\in E$ has norm $\sqrt{\ell(e_i)}$ when viewed as a 1-chain in $C_1(G;\mathbb{R})$, in contrast to its length $\ell(e_i)$ when viewed as an edge in $G$ \citep{ji2012complete}. The inner product $\mathcal{Q}_G$ is compatible with edge subdivision: if $e$ is subdivided into $e'$ and $e''$, then
\begin{equation*}
\|e'+e''\|^2_{\mathcal{Q}} = \ell(e')+\ell(e'')=\ell(e)=\|e\|^2_{\mathcal{Q}}.
\end{equation*}
The 1-homology group $H_1(G;\mathbb{R})$ inherits the inner product $\mathcal{Q}_G$ as a closed subspace of $C_1(G;\mathbb{R})$. For any graph, choosing a 1-cycle basis $\sigma_1,\ldots,\sigma_g$ for $H_1(G;\mathbb{R})$ is equivalent to choosing a spanning forest \citep{cao2025computingtropicalabeljacobitransform}. Once a 1-cycle basis is fixed, the inner product $\mathcal{Q}_G$ is represented by a SPD matrix $Q$. The \emph{tropical Torelli map for weighted graphs} sends graph $G$ to the matrix $Q(G)$.

\subsection{Main Algorithm}\label{sec:algorithm}

We now describe the implementation of the tropical Torelli map to send a graph to its matrix representation as described above.  Specifically, we show how to compute a 1-cycle basis for $H_1(G;\mathbb{R})$ and the SPD matrix $Q(G)$. 

For a connected graph $G$, we first compute a minimal spanning tree $T$. Fix an arbitrary orientation for $G$. For any edge $e=[u,v]\in G\backslash T$, it generates a 1-cycle $\sigma$ such that
$
\sigma = e+\sum_{e'\in \gamma}\sigma(e')e',
$
where $\gamma$ is the unique path in $T$ connecting $v$ to $u$. Since there are $m-(n-1)=g$ edges not in $T$, they generate a 1-cycle basis $\sigma_1,\ldots,\sigma_g$ for $H_1(G;\mathbb{R})$. Construct a $g\times m$ matrix $M$ by setting $M_{ij}=\sigma_i(e_j)$. The matrix $M$ is called the \emph{cycle--edge incidence matrix} \citep{cao2025computingtropicalabeljacobitransform}. 

We reorder the edge set as follows: the first $n-1$ edges $e_1,\ldots,e_{n-1}$ correspond to the minimal spanning tree $T$, while the remaining $g$ edges are sorted in ascending order by length. Then each 1-cycle $\sigma_i$  corresponds to the edge $e_{n-1+i}$. After reordering, the cycle--edge incidence matrix is in the block matrix form $M=[M_T, I_g]$ where $I_g$ is the $g\times g$ identity matrix. Let $L=\mathrm{diag}\{\ell(e_1),\ldots,\ell(e_m)\}$ be the diagonal matrix of edge lengths and $L_T=\mathrm{diag}\{\ell(e_1),\ldots,\ell(e_{n-1})\}$ and $L_g=\{\ell(e_n),\ldots,\ell(e_m)\}$ be its blocks. Then the inner product $\mathcal{Q}_G$ on $H_1(G;\mathbb{R})$ is represented by the matrix
\begin{equation}\label{eq:matrix-Q}
Q = MLM^\top = M_TL_TM_T^\top + L_g.
\end{equation}
For disconnected graphs, we compute $Q$ component-wise and then form a block matrix. We summarize the computation as pseudocode in Algorithm~\ref{alg:mat-Q}.

\begin{algorithm}[t]
\caption{Computing the tropical Torelli map}
\label{alg:mat-Q}
\begin{algorithmic}[1]

\INPUT weighted graph $G$
\OUTPUT matrix $Q$

\STATE $Q \gets [\,]$
\FOR{each connected component $G_i=(E_i,V_i)$}
    \STATE compute a minimal spanning tree $T$
    \STATE sort $E_i$ so that $e_1,\ldots,e_{n_i-1}\in T$ and 
    $\ell(e_{n_i}) < \cdots < \ell(e_{m_i})$
    \STATE $M_T \gets \mathrm{zeros}(g_i, n_i-1)$
    \FOR{$j = 1$ to $g_i$}
        \STATE $[u,v] \gets e_{n_i-1+j}$
        \STATE find path $\gamma$ from $v$ to $u$ in $T$
        \FOR{$k = 1$ to $n_i-1$}
            \STATE $M_{jk} \gets \gamma(e_k)$
        \ENDFOR
    \ENDFOR
    \STATE $Q \gets \mathrm{diag}\{Q,\, M_T L_T M_T^\top + L_{g_i}\}$
\ENDFOR
\end{algorithmic}
\end{algorithm}

\subsection{Generic Length Functions}

The output of Algorithm \ref{alg:mat-Q}, which is our matrix representation of the graph, depends on the choice of minimal spanning trees and ordering of edges, which means that the matrix representation of the graph need not be unique. We now discuss how to achieve the uniqueness of the output. Specifically, we show that the output becomes unique under the genericity assumption of length functions.

\begin{definition}\label{def:generic}
    Let $G$ be a connected graph. A length function $\ell:E:\to\mathbb{R}_+$ is \emph{generic} if $G$ has a unique minimal spanning tree $T$ and the lengths of edges are distinct.
\end{definition}

If $G$ is equipped with a generic length function, it admits a canonical orientation. We prove our main theorem on uniqueness and refinement-invariance of the resulting matrix under the tropical Torelli map for weighted graphs in Appendix~\ref{app:proof}.

\begin{theorem}\label{thm:def-Q}
    Let $G=(V,E)$ be a connected graph with a generic length function $\ell$. Under the canonical orientation, the matrix $Q$ computed by Algorithm \ref{alg:mat-Q} is unique, and invariant under any refinement of $G$.
\end{theorem}

A random length function is generic with probability one: suppose $G$ is a connected graph and for each edge $e_i\in E$, we independently assign a random variable $x_i$ with uniform distribution on $[0,1]$, then we have $\mathbb{P}[\exists i\neq j, x_i=x_j]=0$. In real-world data such as road networks, this assumption holds naturally: edge lengths are subject to measurement noise and variability, making exact equality between lengths highly unlikely. As a result, the length function in such settings is generic almost surely.

\subsection{Computational Complexity}

For a connected graph $G$ with $n$ nodes and $m$ edges, the time complexity to compute the reduced cycle--edge incidence matrix $M_T$ is $O(gn\log n)$. Computing the matrix $Q$ runs in $O(g^2n)$ time. Therefore, the overall time complexity of Algorithm \ref{alg:mat-Q} is $O(gn(g+\log n))$.

The genus $g$ is a fundamental quantity that reflects the topological complexity of a graph and is closely related to its sparsity. We thus propose the following categorizations of graphs into the following \emph{sparsity classes} based on their genus: (i) \textbf{Sparse graphs}: $m\asymp n+c$ for some constant $c>0$; in this case, we have $g=O(1)$, and the total time complexity is $O(n\log n)$. (ii) \textbf{Semi-sparse graphs}: $m\asymp cn$ for some constant $c>0$; in this case, we have $g=O(n)$, and the total time complexity is $O(n^3)$. (iii) \textbf{Dense graphs}: $m\asymp n^{1+c}$ for some constant $0<c\le 1$; in this case we have $g=O(n^{1+c})$ and the total time complexity is $O(n^{3+2c})$. 

Notice that the time complexity ranges from almost linear to quintic, which is in line with the densest scenario for graphs with $O(n^2)$ edges. We remark that the algorithm is particularly fast on sparse graphs, making it well-suited for real-world datasets such as URNs.

\section{Tropical Metric Graph Kernels}\label{sec:tw-kernel}

A common approach to constructing a kernel function on a geometric space is to start with a distance function that captures meaningful dissimilarities between elements. Given a distance $d(x,y)$ on the space, a typical strategy is to convert it into a similarity measure using a radial basis function (RBF), $k(x,y)=\exp(-\gamma d^2(x,y))$, where $\gamma>0$ is a scaling parameter \citep{que2016back,scholkopf1997comparing,zhu2020distance}.
We now construct our metric graph kernel by first enlarging the embedding space to positive semi-definite (PSD) matrices to enable comparisons between graphs of different genus. Then we construct graph kernels using distances on the space of PSD matrices. We present algorithms to compute these kernels and discuss their computational complexity. 

\subsection{The Space of Positive Semi-Definite Matrices.}\label{sec:psd-subsampling}

The tropical Torelli map sends a weighted graph of genus $g$ to a $g\times g$ symmetric positive definite (SPD) matrix. To compare graphs with different genus, we embed the matrices to a common ambient space. Given a sufficiently large upper bound $g_0$, this can be done by simply padding each matrix with zeros, which results in a positive semi-definite (PSD) matrix of fixed size. In this way, graphs are mapped into the same space of PSD matrices $\mathrm{PSD}(g_0)$.

Unlike the space of SPD matrices, which is a smooth manifold that admits various Riemannian metrics, the space of PSD matrices lacks a manifold structure due to the presence of boundary points corresponding to rank-deficient matrices. It is known that $\mathrm{PSD}(g_0)$ is a stratified space which contains $\mathrm{SPD}(g_0)$ as its maximal strata \citep{thanwerdas2022geometry,vandereycken2013riemannian}. Many well-studied Riemannian distances on the SPD manifold can \textbf{not} be extended naturally to PSD space. For example, the log-Euclidean and affine-invariant metrics are not well-defined on the PSD space \citep{bonnabel2010riemannian}.

On the other hand, for computational efficiency, the ambient space should not be of extremely high dimension. To address this issue, we manually set a target dimension $g_0$ and seek to reduce the size of any high-dimensional SPD matrix. In principle, the optimal submatrix should be obtained by minimizing the approximation error in the Frobenius norm. However, the exact solution can only be found by enumerating all permutations of rows and columns, which is computational infeasible. Thus we adopt a simple approach by randomly selecting a principal submatrix of size $g_0$. Notice that here the goal is to obtain a small matrix of size $g_0$ rather than a large matrix of rank $g_0$, which excludes standard low-rank approximation techniques based on the Eckart--Young theorem \citep{eckart1936approximation} or the Nystr\"om method \citep{drineas2005nystrom}. Further discussions and empirical results on stability are provided in Appendix~\ref{app:verify}.


\subsection{Two Tropical Kernels for Metric Graphs}\label{sec:ttw}

We construct our metric graph kernels using the RBF approach and thus now turn to proposing a geometric distance on the space of PSD matrices. 

\textbf{The Bures--Wasserstein Distance.} In the original definition of the tropical Torelli map, a metric graph is sent to a $g$-dimensional flat torus which is the quotient space of $\mathbb{R}^g$ by the lattice spanned by the column vectors of $\sqrt{Q}$. The flat torus can be represented as $(\mathbb{R}^g/\sqrt{Q},\|\cdot\|_2)$, which allows for isometries under orthogonal transformations. That is, given any orthogonal matrix $U\in O(g_0)$, the tori $\mathbb{R}^g/\sqrt{Q}$ and $\mathbb{R}^g/(U\sqrt{Q})$ are equivalent up to isometry. Therefore, we consider the following distance for any $Q_1,Q_2\in \mathrm{PSD}(g_0)$ modulo the orthogonal transformations:
\begin{align}
d_W(Q_1,Q_2)= \min_{U_1,U_2\in O(g_0)} \big\|U_1\sqrt{Q_1}-U_2\sqrt{Q_2}\big\|_F\tag{$*$} 
\end{align}
It turns out that $(*)$ is exactly the \emph{Bures--Wasserstein distance} on $\mathrm{PSD}(g_0)$ and can be computed from the following closed form \citep{bhatia2019bures,thanwerdas2023bures,altschuler2021averaging}:
\begin{equation*}
\begin{aligned}
(*)=\left(\tr(Q_1)+\tr(Q_2)-2\tr\bigg(\sqrt{\sqrt{Q_1}Q_2\sqrt{Q_1}}\bigg)\right)^{\frac{1}{2}}.
\end{aligned}
\end{equation*}
Due to page limits, we defer discussing the motivation from tropical geometry to Appendix \ref{app:trop-jac}. For completeness, we also include a derivation of $(*)$ in Appendix~\ref{app:bw-derivation}.

\textbf{The Tropical Torelli--Wasserstein Kernel.}
Based on the tropical Torelli map for weighted graphs and the Bures--Wasserstein distance on the space of PSD matrices, we define the \emph{tropical Torelli--Wasserstein (TTW) kernel} as
\begin{equation*}
k_{\mathrm{TTW}}(G_1,G_2) = \exp(-\gamma d_W^2(Q(G_1),Q(G_2))).
\end{equation*}
As a consequence of Theorem~\ref{thm:def-Q}, the TTW kernel is invariant to graph refinement, thus, it is a well-defined metric graph kernel. 

To further specify what the TTW kernel captures, we require the notion of 2-isomorphism between weighted graphs \citep{brannetti2011tropical}.
\begin{definition}
    Two weighted graphs $(G_1,\ell_1)$ and $(G_2,\ell_2)$ are \emph{2-isomorphic} if there exists a bijection $\phi:E(G_1)\to E(G_2)$ such that $\ell_2(\phi(e))=\ell_1(e)$ for all $e\in E(G_1)$ and $\phi$ induces a bijection between cycles of $G_1$ and cycles of $G_2$.
\end{definition}
Roughly speaking, a 2-isomorphism preserves the combinatorial structure of a graph $G$ at the edge level and induces an isometry on the underlying metric space $|G|$. The following theorem shows that the TTW kernel measures the similarity between two metric graphs in terms of this isometry.
\begin{theorem}\label{thm:ttw-kernel}
    Let $(G_1,\ell_1)$ and $(G_2,\ell_2)$ be two weighted graphs without bridges and equipped with generic length functions. If $(G_1,\ell_1)$ is 2-isomorphic to $(G_2,\ell_2)$, then $k_{\mathrm{TTW}}(G_1,G_2)=1$; conversely, if $k_{\mathrm{TTW}}(G_1,G_2)=1$, then  there exist refinements $G_1'\geq G_1$ and $G_2'\geq G_2$ such that $G_1'$ is 2-isomorphic to $G_2'$.
\end{theorem}

\textbf{The Tropical Torelli--Euclidean Kernel.} Although the TTW kernel enjoys a clear theoretical interpretation via Theorem~\ref{thm:ttw-kernel} and captures underlying tropical geometry by construction, we make the important remark here that the TTW kernel is not positive definite in the kernel sense, since the 2-Wasserstein distance is not conditionally negative definite on $\mathrm{PSD}(g_0)$ (Appendix~\ref{app:kernel-background}). Therefore, we introduce a simple, alternative kernel obtained by replacing the Bures--Wasserstein distance with the Euclidean distance, which we call the \emph{tropical Torelli--Euclidean (TTE) kernel}:
\begin{equation*}
k_{\mathrm{TTE}}(G_1,G_2) = \exp\big(-\gamma\|Q(G_1)-Q(G_2)\|_F^2\big),
\end{equation*}
Similar to the TTW kernel, the TTE kernel is also a valid metric graph kernel.

Although $k_{\mathrm{TTE}}$ does not take into account the inherent geometry of the space of PSD matrices, it has two key advantages: first, the computation is efficient since it does not require computing the matrix square root, and second, a kernel on $\mathbb{R}^n$ induced by the Euclidean distance is known to be positive definite \citep{haasdonk2004learning}, hence we also have the positive definiteness of the TTE kernel.

\begin{theorem}\label{thm:tte}
    Let $\mathcal{M}_{\leq g}$ be the space of weighted graphs with generic length functions of genus no greater than $g$. Then $k_{\mathrm{TTE}}$ is a positive definite kernel on $\mathcal{M}_{\leq g}$.
\end{theorem}

\begin{algorithm}[htbp]
\caption{Computing the kernel matrix}
\label{alg:mat-kernel}

\begin{algorithmic}[1]
     \INPUT{list of $N$ weighted graph $G$, genus bound $g_0$}
    \OUTPUT{kernel matrix $K$}
    
    \FOR{$i\gets 1$ to $N$}
    \STATE
    compute the tropical Torelli matrix $Q_i$ for $G_i$
    \IF{$\mathrm{dim}(Q_i)<g_0$}
    \STATE
    $Q_i\gets \mathrm{diag}\{Q_i,0\}$
    
    \ELSE
    \STATE
    $Q_i\gets Q_i[\mathrm{rand}(g_0),\mathrm{rand}(g_0)]$
    
    \STATE $K\gets \exp(-\gamma\cdot \mathrm{pairwise\_dist}^2([Q_i]))$
    \ENDIF
    \ENDFOR
\end{algorithmic}
\end{algorithm}

\subsection{Computational Complexity}
The computation of the TTW kernel and the TTE kernel are given in Algorithm~\ref{alg:mat-kernel}.  For a list of $N$ matrices $Q_i$ of dimension $g_0$, the time complexity of computing the full kernel matrix is $O(N^2g_0^2)$ for the TTE kernel and $O(N^2 g_0^3)$ for the TTW kernel if we use eigenvalue decomposition to compute the square root of a matrix. In practice, we find the computation of the TTE kernel matrix is more efficient for sparse graphs since the matrices $Q_i$ can be stored in sparse form which has less memory cost than $O(g_0^2)$. 

\section{Experiments}\label{sec:experiments}

We evaluate our tropical graph kernels with three experimental studies: (i) \textbf{Simulations}: We simulate graphs of different sparsity and test empirical computational time and kernel stability. (ii) \textbf{Comparisons}: We compare the performance of our kernels with classical label-free graph kernels on \emph{de-labeled} benchmark datasets. (iii) \textbf{Real data application}: We construct datasets of local urban road networks and demonstrate the applicability of our methods on  classification tasks.

\textbf{Implementation and Computing Infrastructure.} We use the open source Python library \texttt{GraKeL} \citep{siglidis2020grakel}, which includes all the graph kernels used in in Section \ref{sec:benchmark-test}. Unless otherwise specified, we always use the default hyperparameters provided by \texttt{GraKeL}. All our experiments are performed on a shared server running with 8 CPUs (Intel Xeon Platinum 8358 (Ice Lake) 2.60GHz), 128 GB of RAM, and a walltime limit of 8 hours.  Code is available upon request.

\begin{figure*}[htbp]
    \centering
    \begin{subfigure}{0.18\linewidth}
        \includegraphics[width=\linewidth]{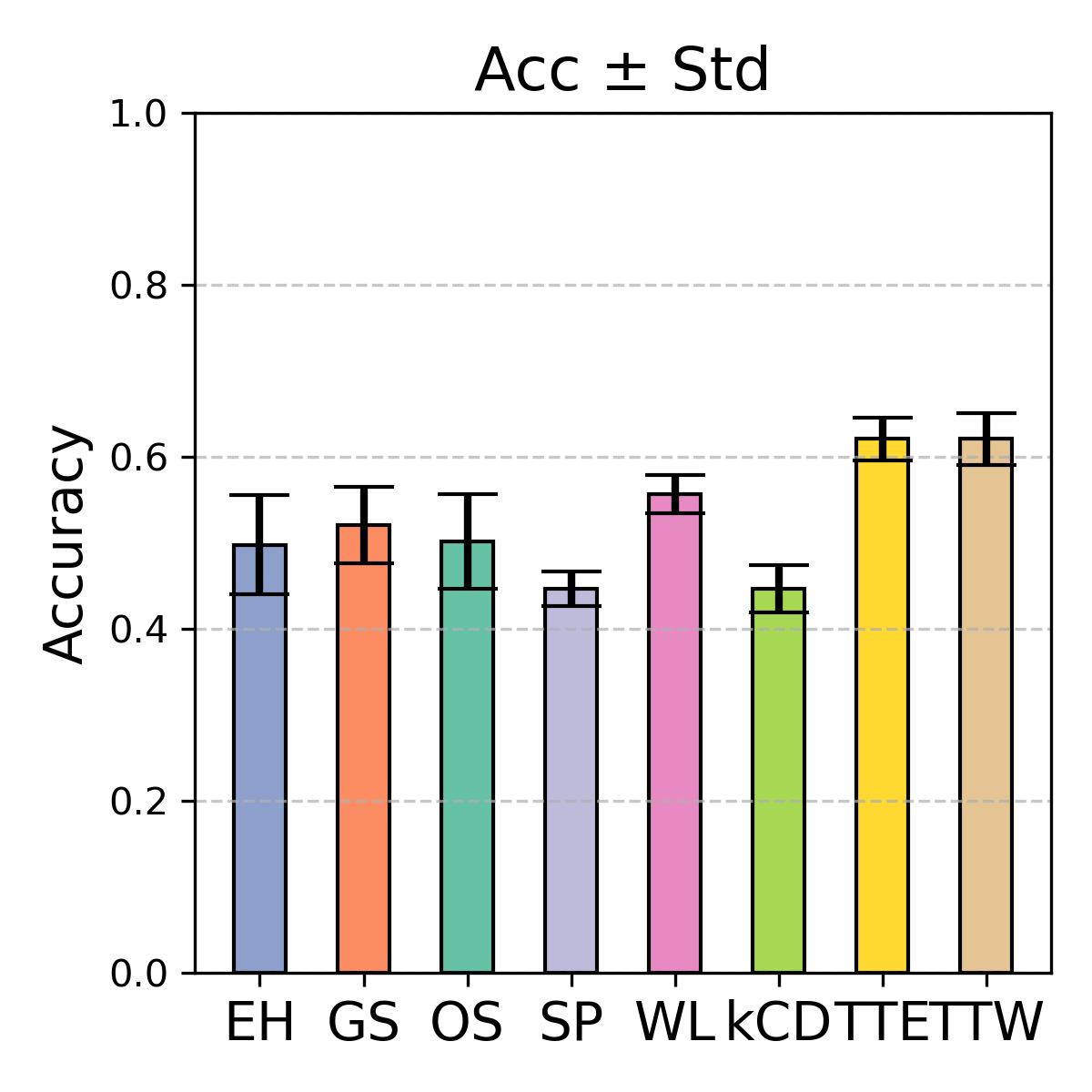}
        \caption{\texttt{FRANKENSTEIN}}
    \end{subfigure}
        \begin{subfigure}{0.18\linewidth}
        \includegraphics[width=\linewidth]{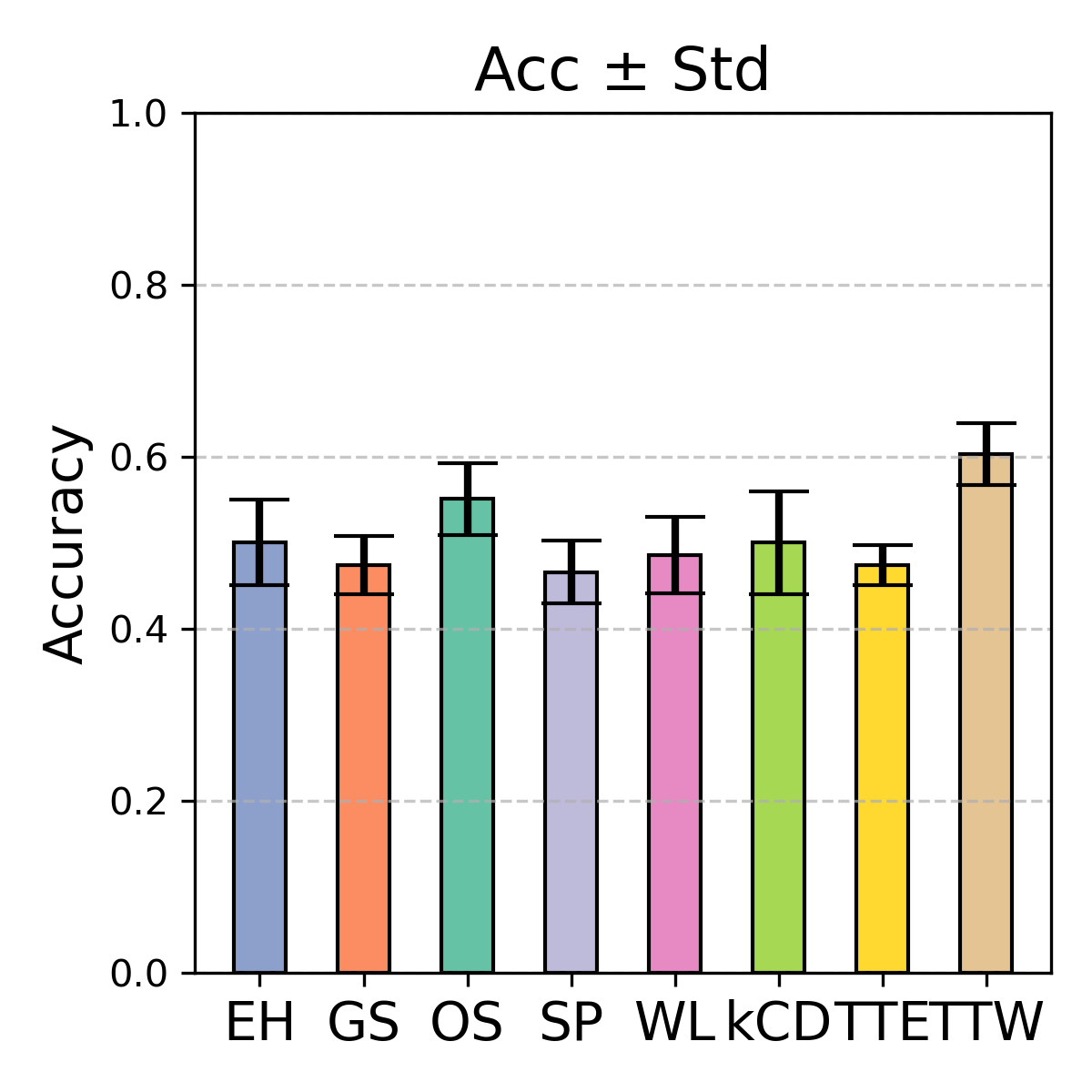}
        \caption{\centering\texttt{IMDB-BINARY}}
    \end{subfigure}
        \begin{subfigure}{0.18\linewidth}
        \includegraphics[width=\linewidth]{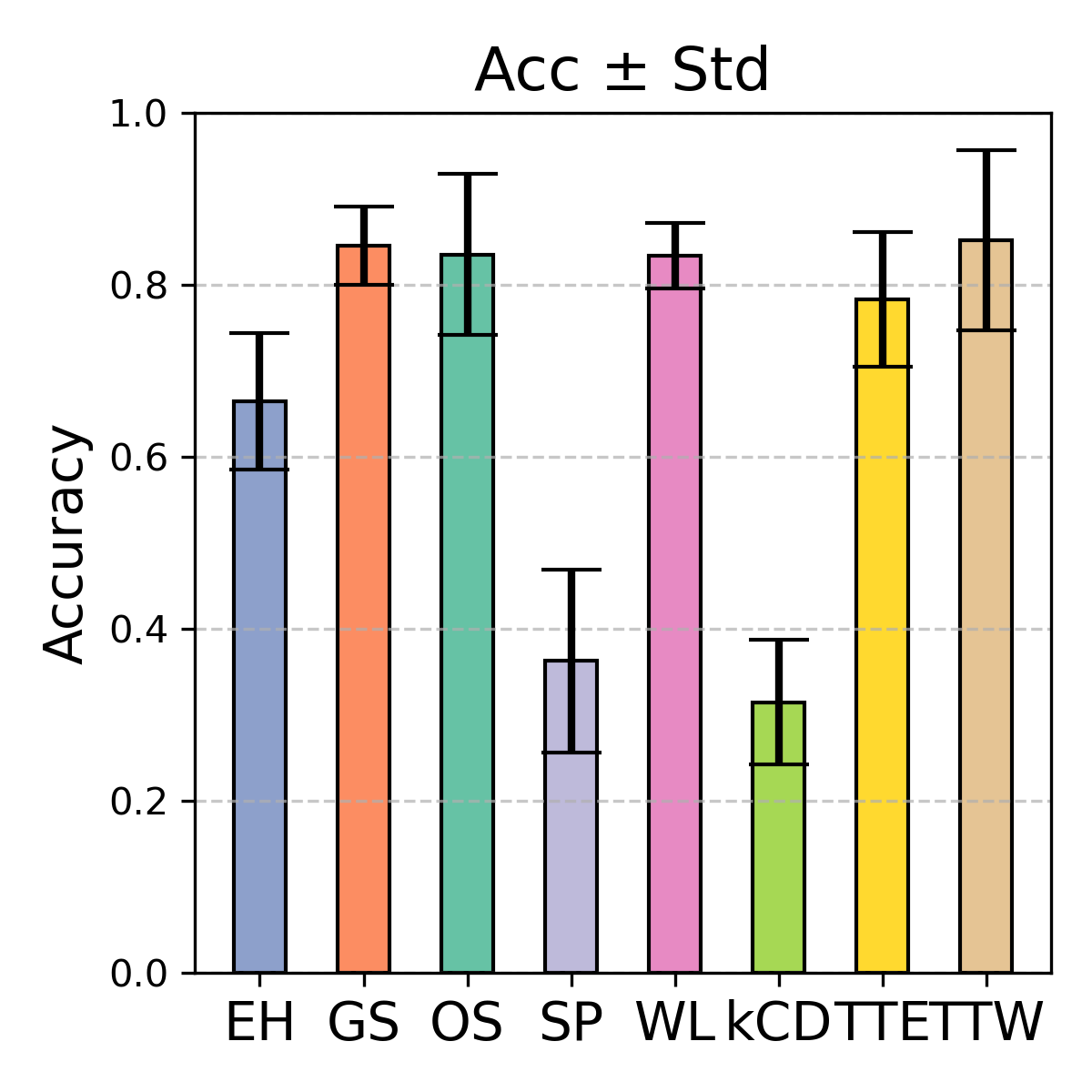}
        \caption{\texttt{MUTAG}}
    \end{subfigure}
        \begin{subfigure}{0.18\linewidth}
        \includegraphics[width=\linewidth]{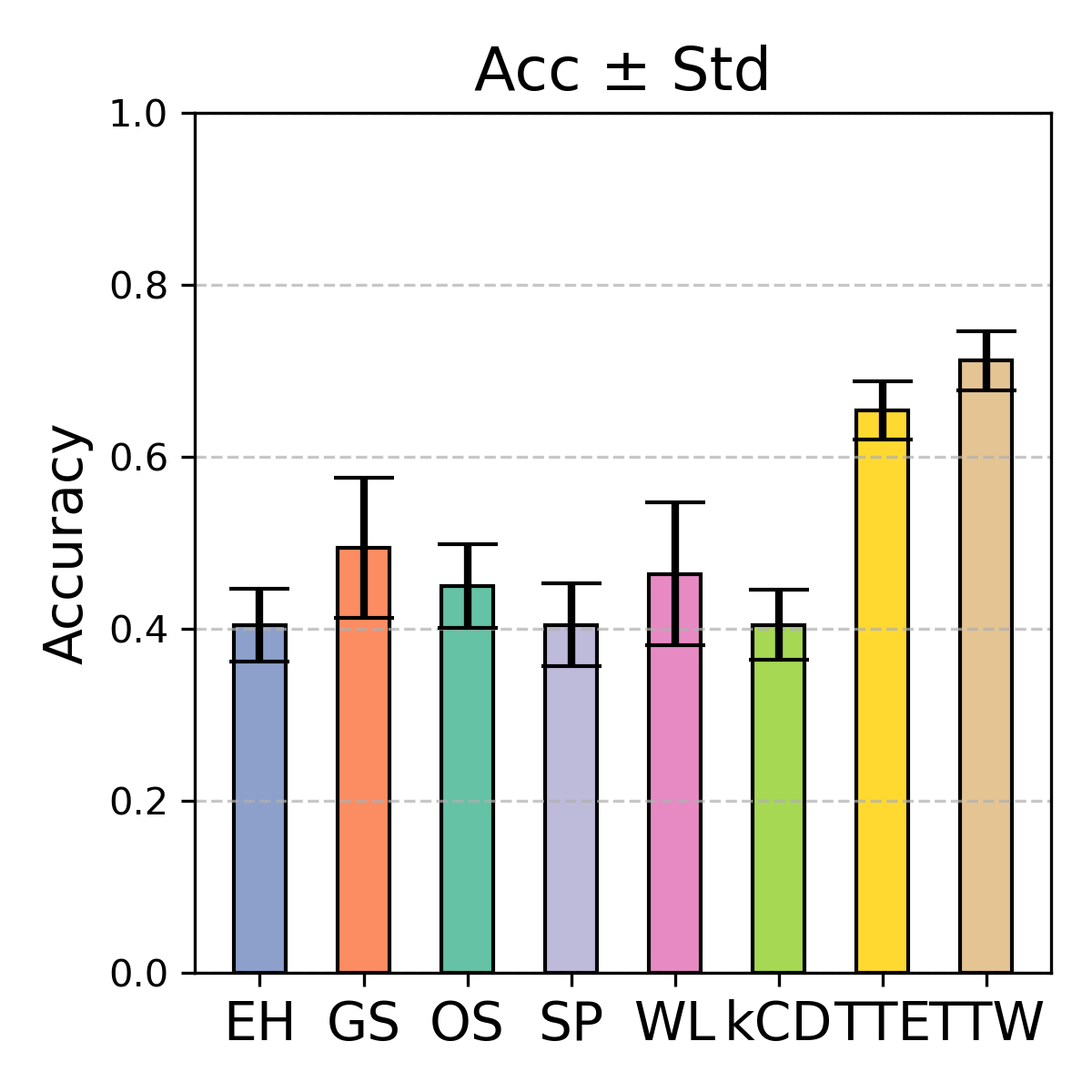}
        \caption{\texttt{PROTEIN}}
    \end{subfigure}
        \begin{subfigure}{0.18\linewidth}
        \centering\includegraphics[width=\linewidth]{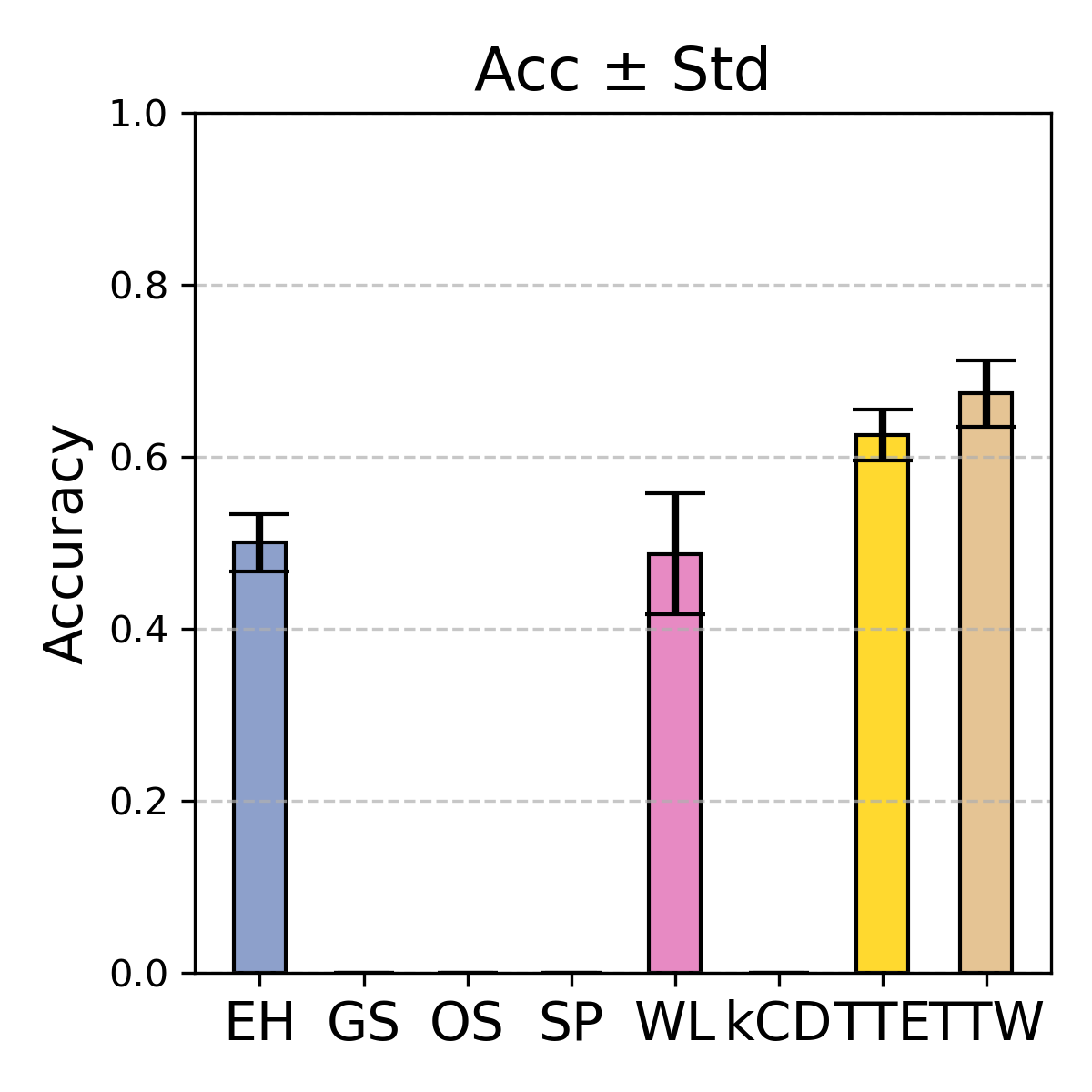}
        \caption{\texttt{REDDIT-BINARY}}
    \end{subfigure}

    \caption{Classification accuracy of 10-fold cross-validation on benchmark datasets \textbf{without} labels (partial).}
    \label{fig:acc-8}
\end{figure*}

\subsection{Simulations on Synthetic Datasets}\label{sec:simulation-synthetic}

\textbf{Verifying Computation Runtime.} Given node number $n$ and genus $g$, we first generated a line graph with $n$ nodes, and then randomly connected two nodes until the edge number achieves $m=g+n-1$. We set the range of $n$ from $50$ to $140$. For each $n$, we generated $N=10$ synthetic graphs corresponding to each of the following settings of sparsity: (i) \textbf{Sparse graphs} such that $g = c_s$ for $c_s = 5,10,20$; (ii) \textbf{Semi-sparse graphs} such that $g=\lfloor c_{ss}n\rfloor$ for $c_{ss}=1.0, 1.5, 2.0$; and (iii) \textbf{Dense graphs} such that $g =\lfloor n^{1+c_d}\rfloor$ for $c_d=0.1,0.2,0.3$. On each synthetic dataset, we repeated the computation of kernel matrices 10 times. Figure \ref{fig:time-test} in Appendix~\ref{app:computation-time} shows how the computation time of the TTE and TTW kernels scales with the number of nodes $n$ across datasets with varying sparsity levels. We observe that the practical computation time for the TTE kernel closely follows the growth pattern of the graph genus, whereas the TTW kernel shows faster growth for denser graphs. Moreover, we see that the computation time for both kernels remains nearly constant on sparse datasets regardless of the change of number of nodes, which highlights their potential for applications involving large sparse graphs.

\textbf{Verifying Empirical Stability.}  Using the same strategy for generating synthetic graphs, we evaluate the empirical stability of our kernels with respect to the constructions described in Section~\ref{sec:tw-kernel}. 

First, we test the stability of random truncation of SPD matrices when reducing a large SPD matrix to a fixed size $g_0$. The results are shown in Appendix~\ref{app:subsampling-stability}, from which we see the random truncation method does not induce large variations in the resulting kernel values. 

Second, we test the positive definiteness of the TTW and TTE kernels by examining the smallest nonzero eigenvalue of the corresponding kernel matrices. The results are shown in Appendix~\ref{app:kernel-stability}. In particular, the results are consistent with Theorem~\ref{thm:tte} that the TTE kernel is positive definite, and they show that in practice we do not observe significant instability due to the indefinite nature of the TTW kernel. 

Finally, we test the sensitivity of the kernel values with respect to perturbations of edge lengths. We assume an additive noise model on edge length, and we measure the difference of kernel matrices under Frobenius norm. The results are given in Appendix~\ref{app:length-stability} and we see that our kernels are robust under edge length perturbations.

\subsection{Comparison to Existing Kernels }\label{sec:benchmark-test}

We now compare the performance of our graph kernels to that of existing graph kernels for graph classification tasks.  Since ours are the first graph kernels designed for metric graphs, there are no prior baselines available in this setting; thus, we compare against classical label-free graph kernels, on benchmark datasets with labels removed, which represent the most relevant existing alternatives for comparison.

\textbf{Label-Free Graph Kernels.} Given that metric graph kernels operate on weighted unlabeled graphs, we focus on classical graph kernels that are likewise independent of node/edge labels and attributes. In particular, we select kernels that can capture global graph structure and are naturally suited to weighted settings. For certain kernels which can be applied to both labeled or unlabeled graphs, we set constant node/edge labels to provide a fair comparison, ensuring that observed performance differences arise from differences in the underlying graph geometry and topology. 


We compare our kernels against the following six label-free graph kernels: the Edge Histogram (EH) kernel \citep{sugiyama2015halting}, the Graphlet Sampling (GS) kernel \citep{shervashidze2009efficient}, the ODD-STh (OS) kernel \citep{da2012tree}, the Shortest Path (SP) kernel \citep{borgwardt2005shortest}, the Weisfeiler--Lehman (WL) kernel \citep{shervashidze2011weisfeiler}, and the $k$-Core Decomposition ($k$CD) kernel \citep{nikolentzos2018degeneracy}. The widely used WL kernel  is based on iterative refinement of node labels and therefore is not designed to respect invariances such as edge subdivision that are central to our setting of metric graphs. As a result, WL addresses a different comparison regime than the one we study. \emph{Nevertheless, due to its prominence in the literature, we include the WL kernel in our experimental evaluation using a constant initial node labeling.} A detailed description of all comparison kernels is provided in Appendix~\ref{app:compare-kernels}.

\textbf{Benchmark Graph Datasets without Labels.} We compare the performance of graph kernels on 23 benchmark graph datasets taken from the TUDataset~\citep{Morris+2020}. To adapt to the setting of metric graphs, we initialize the node/edge labels and weights of graphs by replacing the original node/edge labels by constants. For unweighted graphs, we add random weights to edges from the uniform distribution $\mathrm{Unif}(0,1)$. To compare computational time, we classify the sparsity of benchmark datasets as follows: Let $\bar{n}$ and $\bar{g}$ be the average number of nodes and genus of a dataset. We identify a dataset as (i) sparse, if $\log(\bar{g}/\bar{n})<0$; (ii) semi-sparse, if $0\le \log(\bar{g}/\bar{n})<1$; (iii) dense, if $\log(\bar{g}/\bar{n})\ge 1$. A complete description of dataset statistics can be found in Table~\ref{tab:data-stats-full} in Appendix~\ref{app:data-stat}.

\textbf{Test Results.} We performed classification using the C-SVM solver in \texttt{sklearn} to test the graph kernels. We used the default hyperparameters in \texttt{sklearn} and evaluated the SVM model through a 10-fold cross-validation. Figure~\ref{fig:acc-8} presents the classification accuracy on 5 sample datasets; a complete list of classification accuracy can be found in Table~\ref{tab:acc-full} in Appendix~\ref{app:full-test}. We see that the TTE and TTW kernels outperform other graph kernels on most benchmark datasets. Since we excluded the label information from the graphs, the classification accuracy of some classical graph kernels are lower than the results reported in previous studies \citep{borgwardt2020graph,kriege2020survey}.








We also tested the computation time of feature maps (\texttt{.fit} function in \texttt{GraKeL}) and full kernel matrices (\texttt{.transform} function in \texttt{GraKeL}). We excluded the EH kernel since it is the simplest and fastest. Complete results of computational time can be found in Table~\ref{tab:fit-time-full} and \ref{tab:transform-time-full} in Appendix~\ref{app:full-test}. We see that the TTE and TTW kernel are particularly efficient on sparse and semi-sparse graph datasets. For the computation of full kernel matrices, TTE is the most efficient method and gives results on all 23 datasets.

\subsection{Urban Road Network Classification}

\begin{figure}[htbp]
    \centering
    \includegraphics[width=\linewidth]{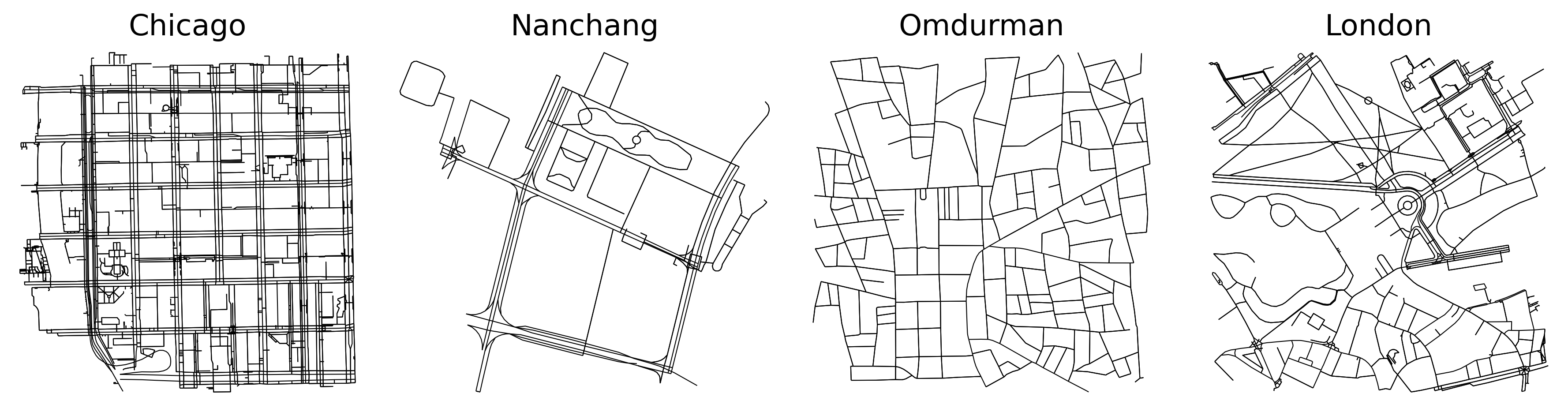}
    \caption{Illustration of URN patterns. From left to right, URNs are taken from Chicago (gridiron), Nanchang (linear), Omdurman (chaotic), and London (tributary).}
    \label{fig:urn-type}
\end{figure}


Urban road networks (URNs) are naturally modeled as metric graphs, where a combinatorial model, i.e., a weighted graph, is built on top of landmarks (nodes) and roads (edges). These networks encode both geometric and topological information, capturing the structural organization of cities. Global URNs present dominant patterns across multiple scales, which are closely related to continents and regions. Based on the study of \citet{chen2024global}, we focus on URN patterns at small scale ($<$ 500 meters), grouped into 4 categories: gridiron, linear, chaotic, tributary.  Figure~\ref{fig:urn-type} presents sample cities from each category.

\begin{table*}[htbp]
  \caption{Classification results for URN datasets. In the table, $\bar{n},\bar{g}$ are the average numbers of nodes and genus. The accuracy is measured in percentage and time is measured in seconds. ``M'' indicates out of memory. The maximal memory is set to 128GB.}
  \label{tab:10fold-urns}
  \centering
  \small
  \begin{tabularx}{\linewidth}{XYYXXXXXXXXX}
    \toprule
    \multicolumn{3}{c}{}&\multicolumn{2}{c}{OS}&\multicolumn{2}{c}{SP}&\multicolumn{2}{c}{TTE} &\multicolumn{2}{c}{TTW}\\
    \cmidrule{4-11}
Name & $\bar{n}$\linebreak (\textit{std.})  & $\bar{g}$\linebreak (\textit{std.}) &Acc\linebreak (\textit{std.}) & Time&Acc\linebreak (\textit{std.}) & Time& Acc\linebreak (\textit{std.}) & Time & Acc\linebreak(\textit{std.}) & Time  \\
\midrule
\texttt{2-1S} & 120.40\linebreak (\textit{66.73})  & 54.26 \linebreak (\textit{34.47}) &84.00\linebreak(\textit{12.21}) &188.70 &92.25\linebreak(\textit{5.30}) &21.01 & 87.50 \linebreak (\textit{5.24}) & 14.96 & \textbf{93.50} \linebreak (\textit{4.90}) & 114.63 \\ 

\texttt{2-1M} & 362.35 \linebreak (\textit{203.81}) & 187.96 \linebreak (\textit{123.03}) & M & M &M&M & 67.10 \linebreak (\textit{7.48}) & 105.43 & \textbf{89.75} \linebreak (\textit{3.61}) & 1357.85 \\

\texttt{2-2S} & 96.40 \linebreak (\textit{77.20}) & 38.84 \linebreak (\textit{36.89}) & 90.50\linebreak(\textit{1.98}) &252.25&\textbf{94.67}\linebreak(\textit{3.23})&39.06 & 91.50 \linebreak (\textit{3.11}) & 17.82 & 92.67 \linebreak (\textit{3.43}) & 166.45 \\ 

\texttt{2-2M} & 289.60 \linebreak (\textit{239.37}) & 123.00 \linebreak (\textit{121.43}) &M &M&M&M & 92.67 \linebreak (\textit{3.82}) & 119.28 & \textbf{94.50} \linebreak (\textit{2.99}) & 1553.34 \\ 

\texttt{3-S} & 55.02 \linebreak (\textit{51.09}) & 22.31 \linebreak (\textit{24.06}) &62.50\linebreak(\textit{15.69}) &99.02&84.01\linebreak(\textit{3.43})&13.27 & 87.67 \linebreak (\textit{4.03}) & 7.25 & \textbf{93.17} \linebreak (\textit{2.83}) & 79.36 \\ 

\texttt{3-M} & 250.49 \linebreak (\textit{206.92}) & 131.61 \linebreak (\textit{128.22}) &M &M&M&M & 75.60 \linebreak (\textit{2.80}) & 143.54 & \textbf{84.13} \linebreak (\textit{2.63}) & 2750.13 \\ 

\texttt{4-S} & 81.82 \linebreak (\textit{74.89}) & 37.68 \linebreak (\textit{38.39}) &\textbf{84.38}\linebreak(\textit{3.80}) &365.67&83.00\linebreak(\textit{4.58})& 67.85& 81.12 \linebreak (\textit{4.12}) & 22.07 & 81.00 \linebreak (\textit{4.36}) & 287.45 \\ 

\texttt{4-M} & 283.36 \linebreak (\textit{238.34}) & 133.03 \linebreak (\textit{125.64}) &M &M&M&M & 68.12 \linebreak (\textit{3.80}) & 155.03 & \textbf{77.25} \linebreak (\textit{3.70}) & 3187.99 \\ 
    \bottomrule
  \end{tabularx}
\end{table*}

\textbf{Constructing Datasets.} The spatial data of URNs were retrieved from the OpenStreetMap (OSM) database \citep{OpenStreetMap}, queried by the open source Python library OSMnx \citep{boeing2024modeling}. We constructed 8 datasets for classification: First, we picked a city for each category. For a city with latitude--longitude coordinate $C=(x,y)$, we retrieved a large road network centered at $C$ with global radius $R$ and then randomly sampled $L$ landmarks from the large network. For each landmark, we retrieved a small road network with local radius $r$. See Table~\ref{tab:urn-data} in Appendix~\ref{app:urn-construct} for full details.

\begin{figure}[htbp]
    \centering
    \begin{subfigure}{0.49\linewidth}
        \includegraphics[width=\linewidth]{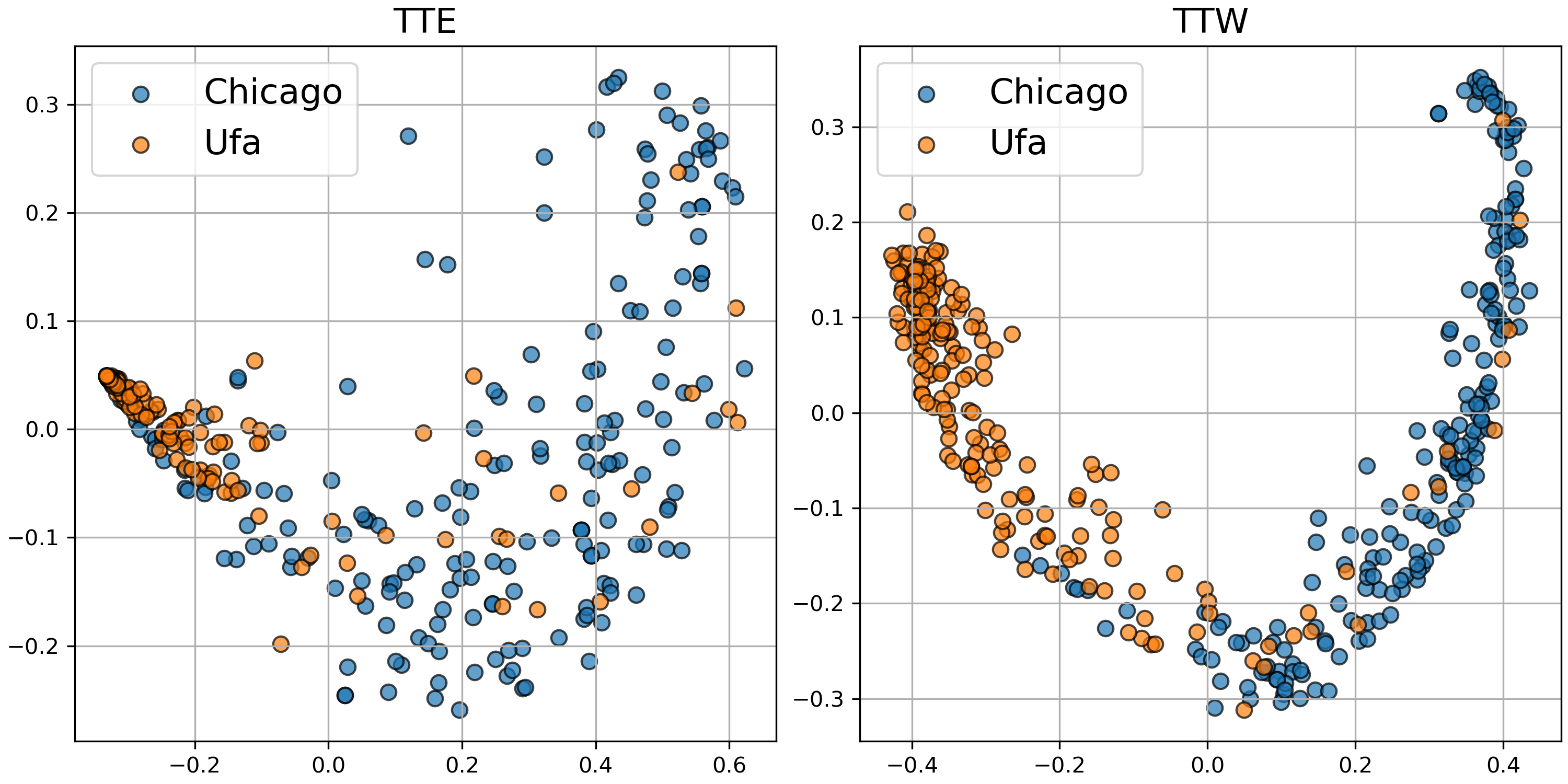}
        \caption{\texttt{URN2-1S}}
    \end{subfigure}
    \begin{subfigure}{0.49\linewidth}
        \includegraphics[width=\linewidth]{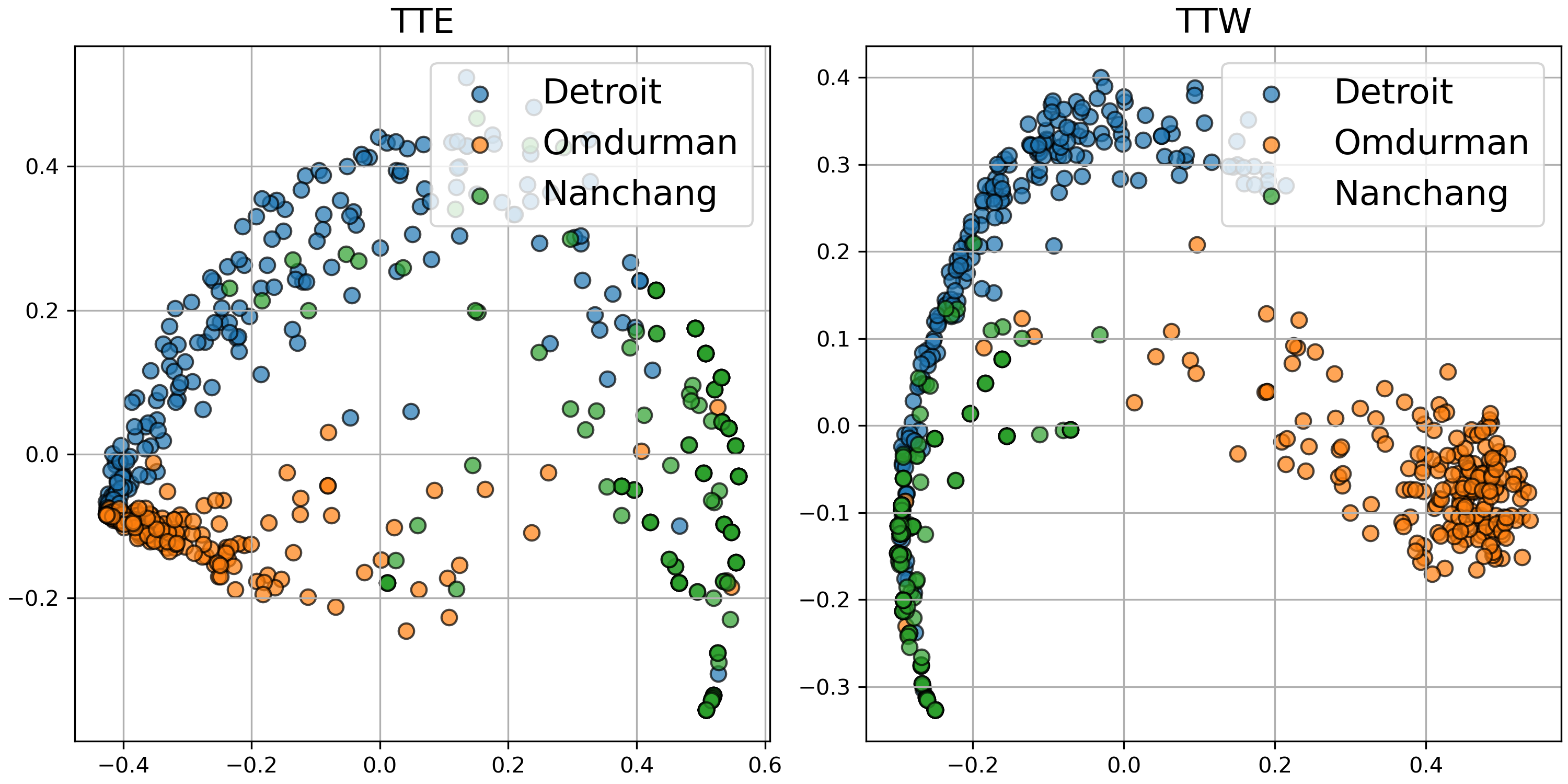}
        \caption{\texttt{URN3-S}}
    \end{subfigure}
    \caption{Dimension reduction by kernel PCA. The low dimensional representations indicate that TTE and TTW kernels are able classify different URNs.}
    \label{fig:kernel-pca}
\end{figure}

\textbf{Classification Results.} We performed classification using C-SVM for 4 graph kernels: OS, SP, TTE, and TTW. Figure~\ref{fig:kernel-pca} displays an exploratory dimension reduction analysis via kernel PCA.  Table~\ref{tab:10fold-urns} shows the complete list of dataset statistics, classification accuracy and computational time. The TTE and TTW kernels outperform the other graph kernels on most datasets. Notably, URNs are typically sparse, which makes the TTE and TTW kernels particularly well-suited for these datasets, where other kernels fail due to memory limitations. Additional results can be found in Appendix~\ref{app:confusion}.

\section{Discussion}\label{sec:discuss}

We proposed new graph kernels which are extendable to the space of metric graphs. We adapted the classical Torelli map from tropical geometry to weighted graphs, embedding them into the space of PSD matrices. We constructed our kernels using two distance functions. Through comprehensive experiments, we demonstrated the effectiveness and applicability of our metric graph kernels.

\textbf{Limitations and Future Research.} Graphs that are trees have genus 0, so the tropical Torelli map is trivial. Consequently, the resulting kernels identify all trees. There are several possible directions to modify either the graph or the graph kernel to incorporate tree-like information, which we discuss further in Appendix~\ref{app:discuss}.

Our method  does not incorporate label or attribute information. However, as discussed in Section~\ref{sec:metric-graph}, the correct analogue of labels in the setting of metric graphs should be functions defined on the underlying metric space.  Developing a principled way to define and integrate such definition on metric graphs remains an open challenge.

Finally, we emphasize that there currently exist no standard nor appropriate benchmark datasets for evaluating methods for metric graphs. Many real-world graphs are not naturally endowed with a meaningful metric structure: for example, molecular graphs encode chemical bonds between atoms, but these bonds do not correspond to geometric edges with interpretable lengths. Tested datasets in this work, such as \texttt{AIDS}, \texttt{BZR}, and \texttt{COX2} fall into this category. Similarly, edge weights or distances in social networks lack a coherent geometric interpretation. Building principled metric graph benchmark datasets is thus an important line for future work.


\vfill\eject

\section*{Acknowledgments}

Y.C.~is supported by Digital Futures Postdoctoral Fellowship. A.M.~is supported by the EPSRC AI Hub on Mathematical Foundations of Intelligence:
An ``Erlangen Programme'' for AI No.~EP/Y028872/1.

\section*{Impact Statement}

This paper presents work whose goal is to advance the field
of Machine Learning. There are many potential societal
consequences of our work, none which we feel must be
specifically highlighted here.

\bibliography{icml2026_ref}
\bibliographystyle{icml2026}

\appendix

\newpage
\appendix
\onecolumn

\section{Background on the Tropical Torelli Map}\label{app:trop-torelli}

In this section of the Appendix, we provide further background on the tropical Torelli map. We present its original definition in both the coordinate-free version and matrix versions. We also provide further discussions on the tropical Torelli map for weighted graphs, especially its motivations.

\subsection{The Coordinate-Free Version}

In classical algebraic geometry over the complex numbers, one of the fundamental results is that every smooth projective algebraic curve $X$ of genus $g$ encodes rich geometric information in a special complex torus. This object is known as the \emph{Jacobian variety} of the curve $\jac(X)$. The classical Torelli map is a natural and powerful construction that assigns to each curve $X$ its Jacobian $\jac(X)$. To study this map, appropriate \emph{moduli spaces} for its domain and codomain are required. In classical algebraic geometry, the construction and study of such moduli spaces has been a central theme of research. More precisely, the Torelli map is defined as a morphism $\mathcal{M}_g\to\mathcal{A}_g$, where  $\mathcal{M}_g$ is the moduli space of smooth complex projective curves of genus $g$, and $\mathcal{A}_g$ is the moduli space of principally polarized abelian varieties of dimension $g$. A fundamental result, known as the \emph{Torelli theorem}, asserts that this map is injective, meaning that the Jacobian retains enough information to uniquely recover a smooth projective curve. The Torelli map thus plays a central role in the study of algebraic curves by linking their intrinsic geometry to the rich structure of abelian varieties \citep{miranda1995algebraic}.

In tropical geometry, an (abstract) topical curve $\Gamma$ is defined as a metric graph. It can be obtained from a particular degeneration process of a one-parameter family of smooth curves, where the edge lengths measure the speed of such degeneration \citep{chan2021moduli}. Analagous to the classical setting, every tropical curve is canonically associated with a real torus called its \emph{tropical Jacobian} $\jac(\Gamma)$. Using graph homology, overviewed in Section~\ref{sec:definition}, the tropical Jacobian is defined as the quotient $H_1(\Gamma;\mathbb{R})/H_1(\Gamma;\mathbb{Z})$, equipped with a canonical positive definite quadratic form induced by the edge lengths of $\Gamma$, which plays the role of a tropical principal polarization. 

The tropical Torelli map assigns to each tropical curve $\Gamma$ its corresponding tropical Jacobian $\jac(\Gamma)$. As in the classical case, a meaningful study of this map requires the introduction of appropriate moduli spaces for its domain and codomain. In contrast to the classical construction over the complex numbers, which relies heavily on complex geometry, the moduli space $\mathcal{M}_g^{\mathrm{trop}}$ of tropical curves  of genus $g$ and the moduli space $\mathcal{A}_g^{\mathrm{trop}}$ of principally polarized tropical abelian varieties of dimension $g$ are constructed using combinatorial and polyhedral methods. With these constructions, the tropical Torelli map is defined as a morphism $\mathcal{M}_g^{\mathrm{trop}}\to\mathcal{A}_g^{\mathrm{trop}}$. It can be shown that the tropical Torelli map is injective on the locus of 3-edge-connected tropical curves \citep{brannetti2011tropical}.

\subsection{The Matrix Version}\label{app:trop-jac}

Notice that, in the coordinate-free definition, the tropical Torelli map assigns to a tropical curve a \emph{torus}, which is not well suited for explicit computation since it is an abstract geometric object. For the purpose of computation and using this construction in applications, as is our goal in this paper, it is therefore necessary to fix a concrete representation of the tropical Jacobian. At the same time, we must be mindful of the situation when different representations correspond to isomorphic tropical Jacobians, and hence determine the same point in the target of the tropical Torelli map.

For any tropical Jacobian, there exist bases under which it can be represented in the standard form $(\mathbb{R}^g/\sqrt{Q}, \|\cdot\|_2)$, where the lattice is spanned by the column vectors of $\sqrt{Q}$, and the tropical polarization is represented by the standard Euclidean inner product. Under this form, the squared distance function is given by
    \begin{equation*}
    d^2([x],[y])=\min_{w\in\mathbb{Z}^g}\big\|x-y-\sqrt{Q}w\big\|_2^2.
    \end{equation*}
    Another torus $(\mathbb{R}^g/\sqrt{Q'},\|\cdot\|_2)$ is isomorphic to $(\mathbb{R}^g/\sqrt{Q},\|\cdot\|_2)$ if and only if there exists  an invertible integer matrix $A\in\mathrm{GL}_g(\mathbb{Z})$ and an orthogonal matrix $U\in O(g)$ such that $\sqrt{Q'}=U\sqrt{Q}A$. This representation of the tropical Jacobian is the motivation for many constructions of our metric graph kernel.
Given this observation, we are now able to define the tropical Torelli map by sending each tropical curve $\Gamma$ to the equivalence class $[Q(\Gamma)]$, where the equivalence relation is defined by the isomorphism described above. This definition could be further simplified if we can always select a unique representative from the class $[Q(\Gamma)]$, however, such a choice is not available in general without imposing additional assumptions.

\subsection{The Tropical Torelli Map for Weighted Graphs}

From a computational perspective, there are two important restrictions arising from the original definition of the tropical Torelli map:
\begin{itemize}
    \item The map is defined on the moduli space of tropical curves of fixed genus $g$. Although the setting is natural for theoretical considerations, it is not well suited if we want to compare metric graphs of different genus.
    \item Even in the matrix version of the tropical Torelli map, working with entire equivalence classes of matrices is computationally prohibitive, as it involves the arithmetic group $\mathrm{GL}_g(\mathbb{Z})$, which is discrete and infinite, and therefore standard continuous optimization methods cannot be applied.
\end{itemize}
The first restriction can be disregarded if the goal is simply to compute the image of the tropical Torelli map for a given graph. To address the second restriction, a practical approach is to fix a basis a priori, which is equivalent to choosing a spanning forest of a weighted graph (or a spanning tree if the graph is connected) and ordering the remaining non-tree edges. With such a choice given, the tropical Torelli map can be well defined for a weighted graph $G$ with a \emph{distinguished spanning forest} $\mathrm{SF}_G$. 

\begin{definition}
    Let $\mathcal{M}^*_{\leq g}$ be the set of  weighted graphs of genus at most $g$, equipped with distinguished spanning forests. The \emph{tropical Torelli map for weighted graphs} is given by
    $
    \mathcal{T}:\mathcal{M}^*_{\leq g}\to\mathrm{SPD}(g),\, G\mapsto Q(G)
    $, where $Q(G)$ is determined by the spanning forest $\mathrm{SF}_G$.
\end{definition}

A generic length function on a weighted graph canonically determines a spanning forest, namely the minimal one (Definition~\ref{def:generic}). In this sense, the condition of a generic length function on a weighted graph  is stronger and implies the condition of a distinguished spanning forest.

Since the bases of the tropical Jacobian are fixed a priori, there is no need to consider equivalence under the arithmetic group $\mathrm{GL}_g(\mathbb{Z})$, which significantly simplifies the computation. Notice that without specifying the bases, a possible well-defined distance between $[Q_1]$ and $[Q_2]$ would be
\begin{equation*}
    d([Q_1],[Q_2]) = \min_{\substack{U_1,U_2\in O(g)\\ A_1,A_2\in\mathrm{GL}_g(\mathbb{Z})}}\|U_1\sqrt{Q_1}A_1-U_2\sqrt{Q_2}A_2\|,
\end{equation*}
for which it is difficult to obtain a closed-form expression or to minimize numerically. However, in the absence of $\mathrm{GL}_g(\mathbb{Z})$, the resulting distance is exactly the Bures--Wasserstein distance (Section~\ref{sec:ttw}).

\subsection{Example}

\begin{figure}[htbp]
    \centering
    \begin{subfigure}{0.4\linewidth}
        \includegraphics[width=\linewidth]{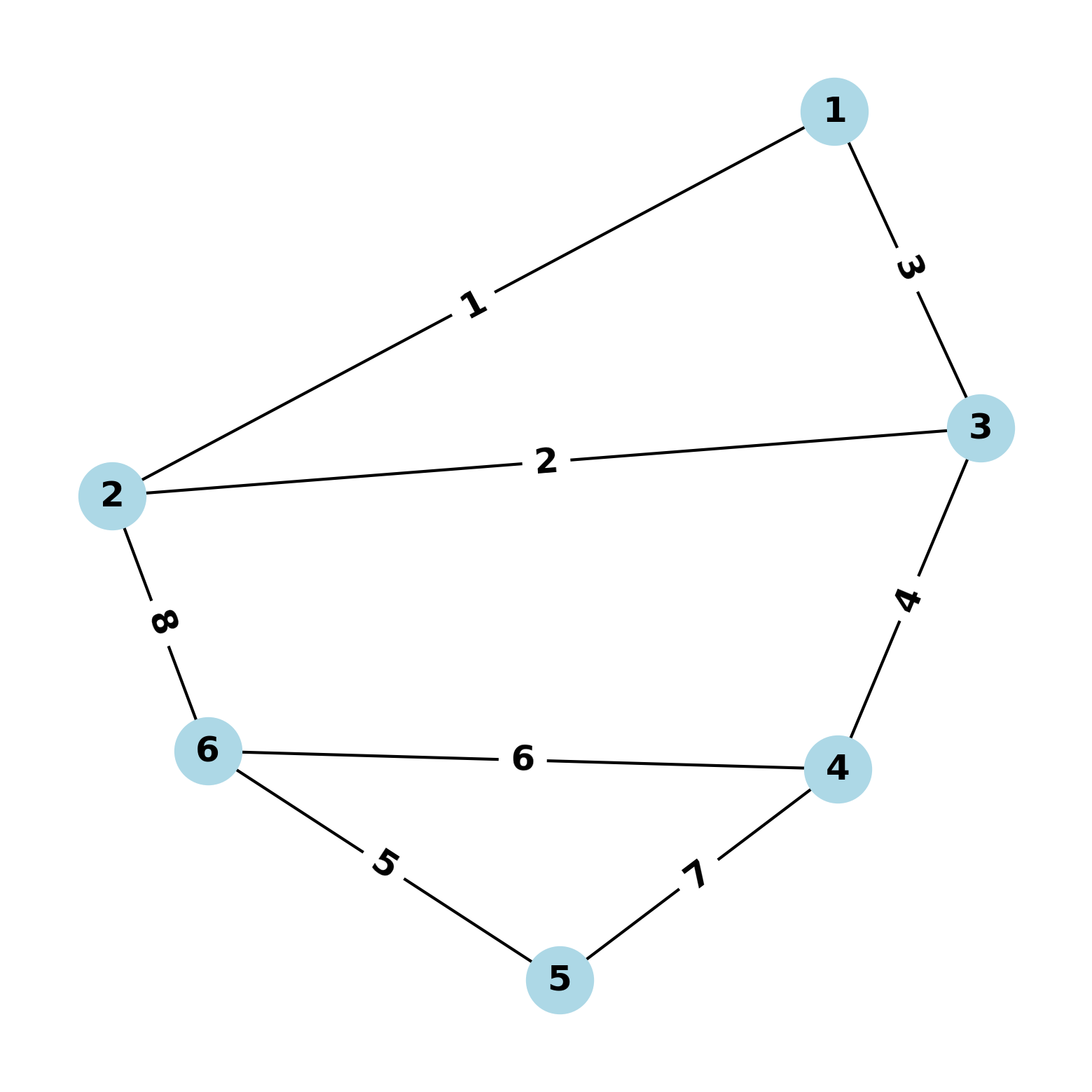}
        \caption{Example graph}
        \label{fig:example}
    \end{subfigure}
    \begin{subfigure}{0.4\linewidth}
        \includegraphics[width=\linewidth]{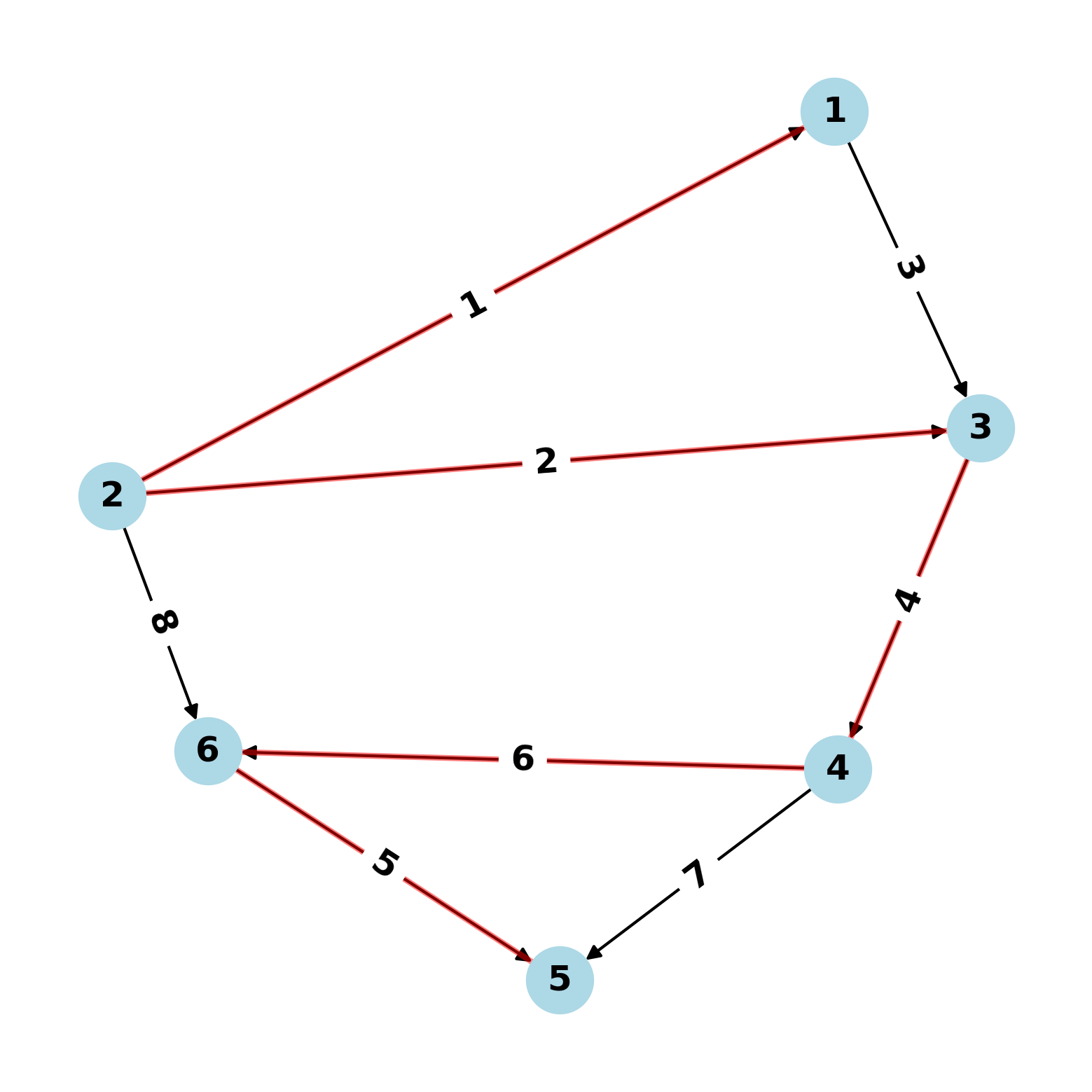}
        \caption{Orientation and minimal spanning tree}
        \label{fig:mst}
    \end{subfigure}
    \caption{Computation of the tropical Torelli map for a  weighted graph. On the right panel, the orientation is indicated by arrow and the minimal spanning tree is colored in red.}
    \label{fig:example-mst}
\end{figure}

We demonstrate the computation of the tropical Torelli map for weighted graphs on a toy example shown in Figure \ref{fig:example}. The graph $G$ has 6 nodes and 8 edges. The length of each edge is defined as $\ell(e_i)=i$. We compute the minimal spanning tree as $T=\{e_1,e_2,e_4,e_5,e_6\}$, which is shown in Figure \ref{fig:mst}. After fixing an orientation, each edge not in $T$ determines a 1-cycle as follows:
\begin{equation*}
\begin{aligned}
    \sigma_1 &= e_3 - e_2 + e_1,\\
    \sigma_2 &= e_7 - e_5 - e_6,\\
    \sigma_3 &= e_8 - e_6 - e_4 - e_2.
\end{aligned}
\end{equation*}
Thus the cycle-edge incidence matrix is given by
\begin{equation*}
M = \begin{bmatrix}
    1 & -1 & 1 & 0 & 0 & 0 & 0 & 0\\
    0 & 0 & 0 & 0 & -1 & -1 & 1 & 0\\
    0 & -1 & 0 & -1 & 0 & -1 & 0 & 1
\end{bmatrix}.
\end{equation*}
The matrix of the tropical Torelli map can be computed by
\begin{equation*}
Q = MLM^\top = \begin{bmatrix}
    6 & 0 & 2\\
    0 & 18 & 6\\
    2 & 6 & 20
\end{bmatrix}.
\end{equation*}

\section{Background on Kernel Support Vector Machines}\label{app:graph-kernel}

In this section, we review the fundamentals of kernel Support Vector Machines (SVMs), which are used in our graph classification experiments. We also provide background on kernel functions, in particular on their construction from distance functions and their positive definiteness.

\subsection{A Short Review on Support Vector Machines}\label{app:svm-intro}

\textbf{General Setup.} Support Vector Machines (SVMs) are supervised learning methods for classification and regression that construct decision functions based on margin maximization \citep{steinwart2008support}. In the binary classification setting, we are given training data $(x_1,y_1),\ldots,(x_n,y_n)$ where $x_i\in\mathbb{R}^m$ and $y_i\in\{+1,-1\}$, the central idea of SVM is to find a hyperplane $f(x)=w^\top x+b$ that separates the two classes with the maximum margin, i.e., the maximum distance between the hyperplane and the closest data points. For linearly separable data, we have the so-called \emph{hard-margin} maximization, which can be formulated as 
\begin{equation*}
    \min_{w,b} \frac{1}{2}\|w\|^2 \quad \text{ subject to}\quad y_i(w^\top x_i+b)\ge 1, \forall i\in\{1,\cdots,n\}.
\end{equation*}

\textbf{Soft-Margin Formulation.} In practice, exact separability rarely holds. This motivates the soft-margin formulation, which is commonly known as \emph{C-SVM}. By introducing slack variables $\xi_i\ge 0,i=1,\ldots,n$, the C-SVM optimization problem is
\begin{equation}\label{eq:original-opt}
    \min_{w,b,\xi} \frac{1}{2}\|w\|^2 + C\sum_{i=1}^n\xi_i
    \quad \text{subject to} \quad
    \left\{\begin{array}{ll}
       y_i(w^\top x_i+b)\ge 1-\xi_i  & \forall\ i\in\{1,\cdots,n\} \\
       \xi_i\ge 0,  & 
    \end{array}\right.
\end{equation}
which is equivalent to the following unconstrained problem with hinge loss:
\begin{equation*}
    \min_{w,b} \frac{1}{2}\|w\|^2 + C\bigg[\sum_{i=1}^n\max(0,1-y_i(w^\top x_i+b))\bigg].
\end{equation*}
In the case of C-SVM, the hyperparameter $C>0$ is a regularization parameter controlling the trade-off between margin maximization and constraint violations. Large $C$ adds more penalty on misclassification while small $C$ yields wider margins at the expense of allowing more violations.

\textbf{Prediction.} The Lagrangian dual of \eqref{eq:original-opt} is given by
\begin{equation}\label{eq:dual}
     \max_{\alpha} \sum_{i=1}^n\alpha_i - \frac{1}{2} \sum_{i,j=1}^n\alpha_i\alpha_jy_iy_jx_i^\top x_j
    \quad \text{subject to} \quad \left\{\begin{array}{ll}
       0\le\alpha_i\le C  & \forall\ i\in\{1,\cdots,n\} \\
       \displaystyle\sum^n_{i=1}y_i\alpha_i=0.  & 
    \end{array}\right.
\end{equation}
The optimal multipliers $\alpha^*$ can be obtained by solving \eqref{eq:dual}, and the primal-dual relationship gives
\begin{equation*}
w=\sum_{i=1}^n\alpha^*_iy_ix_i.
\end{equation*}
Only those samples with $\alpha^*_i>0$ contributes to $w$, which are called \emph{support vectors}. The bias term can be obtained using any support vector by setting $b=y_i-\sum_{j=1}^n\alpha_j^*y_jx^\top_j x_i$. Then the decision function is given by
\begin{equation*}
    f(x) = w^\top x+b = \sum_{i=1}^n\alpha^*_iy_ix_i^\top x + b.
\end{equation*}
For any new data $x_{\mathrm{new}}$, the prediction is $\hat{y} = \mathrm{sign}(f(x_{\mathrm{new}}))$.

\textbf{The Kernel Trick.} The dual formulation \eqref{eq:dual} only depends on the inner product $x_i^\top x_j$. This observation leads to the so-called \emph{kernel trick} which essentially replaces $x_i^\top x_j$ by a kernel function $k(x_i,x_j)$. Therefore the kernel SVM formulation is given by
\begin{equation}\label{eq:kernel}
     \max_{\alpha} \sum_{i=1}^n\alpha_i - \frac{1}{2} \sum_{i,j=1}^n\alpha_i\alpha_jy_iy_jk(x_i,x_j)
    \quad \text{subject to} \quad \left\{\begin{array}{ll}
       0\le\alpha_i\le C  & \forall\  i\in\{1,\cdots,n\} \\
       \displaystyle\sum^n_{i=1}y_i\alpha_i=0.  & 
    \end{array}\right.
\end{equation}
The decision function then becomes
\begin{equation*}
    f(x) =\sum_{i=1}^n\alpha^*_iy_i k(x_i,x) + b,
\end{equation*}
with the optimal multipliers $\alpha^*$ solved from \eqref{eq:kernel}.

\subsection{Background on Kernels}\label{app:kernel-background}

Kernels form a foundational class of similarity functions in machine learning and data analysis, 
which are often used to capture nonlinear relationships by embedding data into high-dimensional feature spaces. 
Let $\mathcal{X}$ be a set and $k:\mathcal{X}\times \mathcal{X}\to\mathbb{R}$ be a symmetric function.
Commonly $k$ is required to be positive definite in the sense that for any finite set of points $x_1, \ldots, x_n \in \mathcal{X}$ the matrix $[K_{ij}=k(x_i,x_j)]_{n\times n}$
is positive semi-definite, i.e., for any real numbers $c_1,\ldots, c_n$, we have
\begin{equation*}
\sum_{i=1}^n \sum_{j=1}^n c_i c_j k(x_i, x_j) \geq 0.
\end{equation*}
In the kernel SVM formulation, a positive semi-definite matrix $[K_{ij}]_{n\times n}$ ensures that \eqref{eq:kernel} is a quadratic program. 

For any positive definite kernel $k$, there exists a unique Hilbert space $\mathcal{H}$ known as the
\emph{reproducing kernel Hilbert space (RKHS)}, along with a map $\phi: \mathcal{X} \to \mathcal{H}$ called the \emph{feature map}, such that the value of the kernel is given by the inner product of the feature map \citep{wainwright2019high}:
\begin{equation*}
k(x,y)=\langle\phi(x),\phi(y)\rangle_{\mathcal{H}}.
\end{equation*}
A typical way to construct a kernel involves the distance on the underlying space. Let $d(x,y)$ be a distance function on $\mathcal{X}$, the distance is said to be of \emph{conditionally negative type} if for any finite set of points $x_1, \ldots, x_n \in \mathcal{X}$ and any real numbers $c_1, \ldots, c_n$ such that $\sum_{i=1}^n c_i = 0$, we have
\begin{equation*}  
\sum_{i=1}^n \sum_{j=1}^n c_i c_j d(x_i, x_j) \leq 0.
\end{equation*}
If $d(x,y)$ is of conditionally negative type, then the kernel defined by $k(x,y) = \exp(-\gamma d(x,y)^2)$ is positive definite for any $\gamma>0$; this result is known as \emph{Schoenberg's theorem} \citep{sejdinovic2013equivalence}. 
The Euclidean distance $d(x,y)=\|x-y\|_2$ for $x,y\in\mathbb{R}^m$ is a typical example of a distance of conditionally negative type. However, the 2-Wasserstein distance is not of conditionally negative type on the space of Gaussians of dimension greater than 1  \citep{peyre2019computational}, and thus the induced kernel is indefinite. In practice, indefinite kernels are also widely used, and the corresponding learning methodology is called \emph{indefinite learning}. In general, an indefinite kernel corresponds to a \emph{Reproducing Kernel Kre\u{\i}n Space (RKKS)}, on which many RKHS-based techniques can be extended \citep{oglic2018learning,togninalli2019wasserstein}.

\section{Technical Proofs}

\subsection{Proof of Theorem~\ref{thm:def-Q}}\label{app:proof}

\begin{proof}
We first show that if $G$ has a generic length function, then $G$ can be oriented in a canonical way through the following operations: (1) for any $e\in T$ with endpoints $u,v$, define $\omega(u)$ to be the length of the other tree edge incident to $u$.  We set $\omega(u)=0$ if no such edge exists; if there is more than one incident edge, we choose $\omega(u)$ to be the minimum length among all incident tree edges. Then we orient $e$ from low to high $\omega$ value; (2) for any $e\notin T$ with endpoints $u,v$, define $w(u)$ as the length of the unique tree edge incident to $u$ that lies on the path from $u$ to $v$ in $T$. Then we orient $e$ from low to high $\omega$ value.

In Algorithm \ref{alg:mat-Q}, the edges not in the minimal spanning $T$ are sorted by length. With a generic length function $\ell$, the minimal spanning tree $T$ is unique and the edge lengths are distinct, thus the 1-cycles $\sigma_1,\ldots,\sigma_g$ are uniquely determined. It suffices to show that the cycle-edge matrix is independent of the ordering of edges in $T$. Suppose $e'_1,\ldots,e'_{n-1}$ is another ordering of edges in $T$. Then there exists a permutation matrix $P\in S_{n-1}$ such that $M_T'=M_TP$ and $L_T'=P^\top L_TP$, and 
\begin{equation*}
Q'= M_T'L_T'M_T'^\top+L_{g}=M_TPP^\top L_T P P^\top M_T^\top + L_g = Q.
\end{equation*}
We now show that the matrix $Q$ is invariant under edge subdivision. Without loss of generality we assume that $e_{s}$ is subdivided into $e'_{s}$ and $e_{s+1}'$. For the remaining edges, we set $e'_k=e_k$ if $k<s$, and $e'_{k}=e_{k-1}$ if $k>s+1$. For cycle-edge incidence matrices we have
    \begin{equation*}
    \left\{\begin{array}{ll}
        M'_{ik} = M_{ik}, &  1\le i \le g\,, 1\le k\le s,\\
        M'_{ik} = M_{i(k-1)}, & 1\le i\le g,\, s+2\le k\le m+1,\\
        M'_{i(s+1)} = M_{is},& 1\le i\le g.
    \end{array}\right.
    \end{equation*}
    For edge-length matrices we have
    \begin{equation*}
    \left\{\begin{array}{ll}
        L'_{kk} = L_{kk},& 1\le k<s,\\
        L'_{kk} = L_{(k-1)(k-1)}, & s+1< k \le m+1,\\
        L'_{ss}+L'_{(s+1)(s+1)} = L_{ss}.&
    \end{array}\right.
    \end{equation*}
    Therefore, for any $1\le i,j\le g$, we have
    \begin{equation*}
    \begin{aligned}
        Q'_{ij} =& \sum_{k=1}^{m+1} M'_{ik}L'_{kk}M'_{jk}\\
        =& \sum_{k=1}^{s-1}M_{ik}L_{kk}M_{jk} + M'_{is}L'_{ss}M'_{js} + M'_{i(s+1)}L'_{(s+1)(s+1)}M'_{j(s+1)}+ \sum_{k=s+2}^{m+1}M'_{ik}L'_{kk}M'_{jk}\\
        =& \sum_{k=1}^{s-1}M_{ik}L_{kk}M_{jk} + M_{is}(L'_{(s+1)(s+1)}+L'_{ss})M_{js} +\sum_{k=s+1}^{m}M_{ik}L_{kk}M_{jk}\\
        =& \sum_{k=1}^m M_{ik}L_{kk}M_{jk} = Q_{ij} .
    \end{aligned}
    \end{equation*}
Finally, for any graph refinement $G'\geq G$, we can decompose it as $G'=G_k\geq G_{k-1}\geq\ldots\geq G_0=G$ where each $G_{i+1}\geq G_i$ is given by one edge subdivision. Due to the invariance under edge subdivision we have $Q(G')=Q(G)$, as desired. 
\end{proof}

\subsection{Derivation of the Bures--Wasserstein Distance}\label{app:bw-derivation}

We now provide a brief proof of the following formula:
\begin{equation}\label{eq:bw-eq}
    \min_{U_1,U_2\in O(g)}\|U_1\sqrt{Q_1}-U_2\sqrt{Q_2}\|_F^2  = \tr(Q_1)+\tr(Q_2)-2\tr\bigg(\sqrt{\sqrt{Q_1}Q_2\sqrt{Q_1}}\bigg)
\end{equation}
for positive semi-definite real matrices $Q_1,Q_2\in\mathrm{PSD}(g)$. A general proof for positive semi-definite Hermitian matrices can be found in \citet{bhatia2019bures}.

\begin{proof}
First, notice that
\begin{equation*}
    \begin{aligned}
        \min_{U_1,U_2\in O(g)}\|U_1\sqrt{Q_1}-U_2\sqrt{Q_2}\|_F^2&= \min_{U\in O(g)}\|U\sqrt{Q_1}-\sqrt{Q_2}\|_F^2\\
        &=\min_{U\in O(g)} \tr\big(\big(U\sqrt{Q_1}-\sqrt{Q_2}\big)^\top\big(U\sqrt{Q_1}-\sqrt{Q_2}\big)\big)\\
        &=\min_{U\in O(g)}\tr(Q_1)+\tr(Q_2)-\tr\big(\sqrt{Q_2}\sqrt{Q_1}U^\top\big)-\tr\big(\sqrt{Q_1}\sqrt{Q_2}U\big)\\
        &=\tr(Q_1)+\tr(Q_2)-2\max_{U\in O(g)}\tr\big(\sqrt{Q_1}\sqrt{Q_2}U\big).
    \end{aligned}
\end{equation*}
Now consider the SVD decomposition $\sqrt{Q_1}\sqrt{Q_2}=W\Lambda V^\top$. We have
\begin{equation*}
    \tr\big(\sqrt{Q_1}\sqrt{Q_2}U\big)=\tr(W\Lambda V^\top U)=\tr(\Lambda V^\top UW)=\sum_{i=1}^g\lambda_i \theta_i,
\end{equation*}
where $\theta_i$'s are the diagonal elements of $V^\top UW\in O(g)$. Since $|\theta_i|\le 1$, we have $\displaystyle\sum_{i=1}^g\lambda_i\theta_i\le\sum_{i=1}^g\lambda_i=\tr(\Lambda)$, where equality holds if and only if $V^\top UW = I$. Since the singular values of $\sqrt{Q_1}\sqrt{Q_2}$ are the square roots of the eigenvalues of $(\sqrt{Q_1}\sqrt{Q_2})(\sqrt{Q_1}\sqrt{Q_2})^\top=\sqrt{Q_1}Q_2\sqrt{Q_1}$, we have \begin{equation*}
\tr(\Lambda) = \tr\bigg(\sqrt{\sqrt{Q_1}Q_2\sqrt{Q_1}}\bigg),
\end{equation*}
which is the last term of \eqref{eq:bw-eq}.
\end{proof}

\subsection{Proof of Theorem~\ref{thm:ttw-kernel}}

Recall that a graph $G$ is said to be \emph{$k$-edge-connected} if the graph obtained from $G$ by removing any set of $s\le k-1$ edges is connected. An edge $e\in E(G)$ is a \emph{bridge} if removing $e$ will increase the number of connected components by 1. Thus, a graph $G$ is 2-edge-connected if $G$ has no bridges. A graph $G$ is \emph{3-edge-connected} if removing any two edges still leaves $G$ connected. In particular, the characterization of 3-edge-connected graphs is given in terms of \emph{C1-sets} \citep{caporaso2010torelli}: suppose $\widetilde{G}$ is obtained from $G$ by contracting all bridges. A subset $S\subseteq E(\widetilde{G})$ is a \emph{C1-set} of $G$ if $\widetilde{G}\backslash S$ has no bridges and contracting all edges not in $S$ yields a cycle. For a weighted graph $(G,\ell)$, we have the following operation of \emph{3-edge-connectivization}:

\begin{definition}\label{def:3-edge}
    Let $G$ be a graph equipped with length function $\ell:E(G)\to\mathbb{R}_+$. A \emph{3-edge-connectivization} of $(G,\ell)$ is a new weighted graph $(G^3,\ell^3)$ such that
    \begin{itemize}
        \item $G^3$ is obtained by contracting all bridges of $G$, and for each C1-set $S$ of $G$, contracting all but one edge of $S$; and
        \item For any $e\in E(G^3)$, let $S_e$ be the corresponding C1-set of $e$ in $G$, then $\ell^3(e)=\displaystyle\sum_{e'\in S_e}\ell(e')$.
    \end{itemize}
\end{definition}

Defintion~\ref{def:3-edge} is a simplified version of 3-edge-connectivization of a tropical curve $\Gamma$ \citep[Lemma-Definition 5.3.2]{brannetti2011tropical}. The following injectivity result for the tropical Torelli map applies directly to our setting.

\begin{theorem}{\citep[Theorem 5.3.3]{brannetti2011tropical}}\label{thm:inj}
    Let $\Gamma$ and $\Gamma'$ be two tropical curves of genus $g$. Then $\jac(\Gamma)\cong\jac(\Gamma')$ if and only if $\Gamma^3$ is 2-isomorphic to $\Gamma'^3$.
\end{theorem}

We now proceed to our proof of Theorem~\ref{thm:ttw-kernel}.

\begin{proof}
    Assume $(G,\ell)$ is 2-isomorphic to $(G',\ell')$, and let $\phi:E(G)\to E(G')$ be a 2-isomorphism. Since $\ell$ is generic, it determines a unique minimal spanning forest $\mathrm{SF}_{G}$ for $G$. Similarly, $\ell'$ determines a unique minimal spanning forest $\mathrm{SF}_{G'}$ for $G'$. Since $\phi$ preserves edge lengths, it sends  $\mathrm{SF}_{G}$ to $\mathrm{SF}_{G'}$ and preserves the order of non-tree edges in $G$ and $G'$. By construction we have $Q(G)=Q(G')$ and $k_{\mathrm{TTW}}(G,G') = 1$.

    Conversely, if $k_{\mathrm{TTW}}(G,G')=1$, then we have $(\mathbb{R}^{g}/\sqrt{Q(G)},\,\|\cdot\|_2)\cong(\mathbb{R}^{g'}/\sqrt{Q(G')},\,\|\cdot\|_2)$. In particular it implies that $g=g'$ and $\jac(G)\cong\jac(G')$. By Theorem~\ref{thm:inj}, the 3-edge-connectivizations of $G$ and $G'$ are 2-isomorphic. Let  $\phi:E(G^3)\to E(G'^3)$ be such a 2-isomorphism. Since $G$ is a refinement of $G^3$, there is a surjection $j:E(G)\to E(G^3)$. The composition $\phi\circ j:E(G)\to E(G'^3)$ determines a new refinement $G'_*\geq G'^3$. Take a common refinement $\widehat{G'}$ such that $\widehat{G'}\geq G'_*$ and  $\widehat{G'}\geq G'$. Let $j':E(\widehat{G'})\to E(G'^3)$ be the associated surjection of $\widehat{G'}\ge G'^3$. Similarly, the composition $\phi^{-1}\circ j':E(\widehat{G'})\to E(G^3)$ determines a new refinement $\widehat{G}\geq G^3$. Since $\widehat{G'}$ is already a refinement of $G'_*$, it follows that $\widehat{G}$ is a refinement of $G$. Moreover, the last construction yields a 2-isomorphism $E(\widehat{G'})\to E(\widehat{G})$. In summary, we have shown that $\widehat{G}\ge G$, $\widehat{G'}\ge G'$, and $\widehat{G}$ and $\widehat{G'}$ are 2-isomorphic.
\end{proof}
\subsection{Proof of Theorem~\ref{thm:tte}}

\begin{proof}
    By Theorem~\ref{thm:def-Q}, for a weighted graph $G$ of genus $g$ equipped with a generic length function, the map $\mathcal{M}_{\le g}\to \mathrm{PSD}(g),\, G\mapsto Q(G)$ is well-defined. For any $G_1,\ldots,G_n\in\mathcal{M}_{\le g}$ and real numbers $c_1,\ldots,c_n\in\mathbb{R}$, since the Gaussian kernel $k(x,y)=\exp(-\gamma\|x-y\|^2)$ is known to be positive definite on $\mathbb{R}^{g^2}$, we have
    \begin{equation*}
        \sum_{i,j=1}^nc_ic_j\exp\big(-\gamma\|Q(G_i)-Q(G_j)\|_F^2\big)\ge 0,
    \end{equation*}
    which implies that $k_{\mathrm{TTE}}$ is a positive definite kernel on $\mathcal{M}_{\le g}$.
\end{proof}

\section{Additional Numerical Verifications}
\label{app:verify}

In this section we present additional numerical results in Section~\ref{sec:simulation-synthetic}. Specifically, we test practical computational runtime of our metric graph kernels. We further investigate the stability of kernel matrices under random subsampling, test their positive semi-definiteness, and analyze their sensitivity to perturbations in edge lengths.

\subsection{Verifying Computation Runtime.}\label{app:computation-time}

\begin{figure}[htbp]
    \centering
    \begin{subfigure}{0.32\linewidth}
        \includegraphics[width=\linewidth]{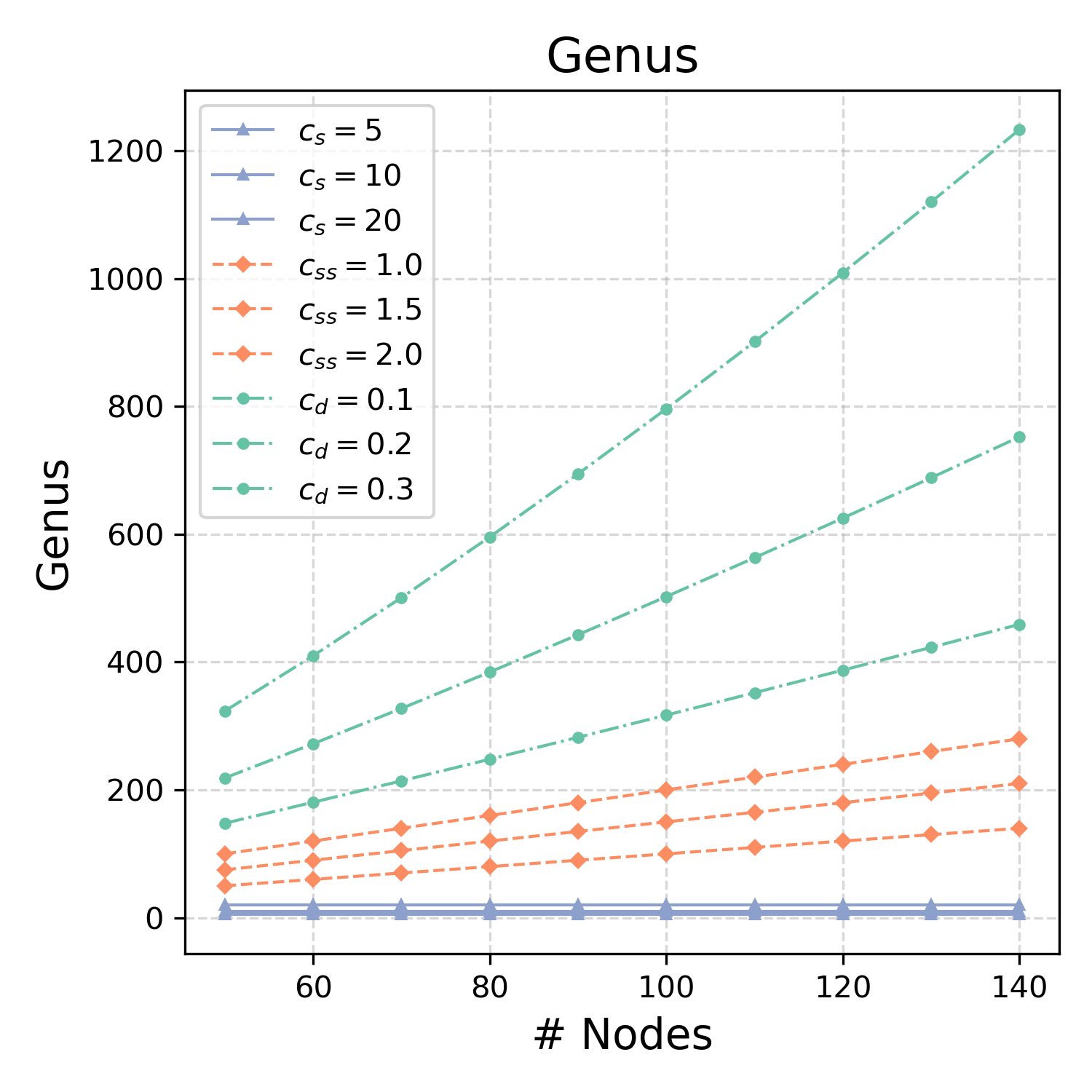}
    \end{subfigure}
\begin{subfigure}{0.32\linewidth}
        \includegraphics[width=\linewidth]{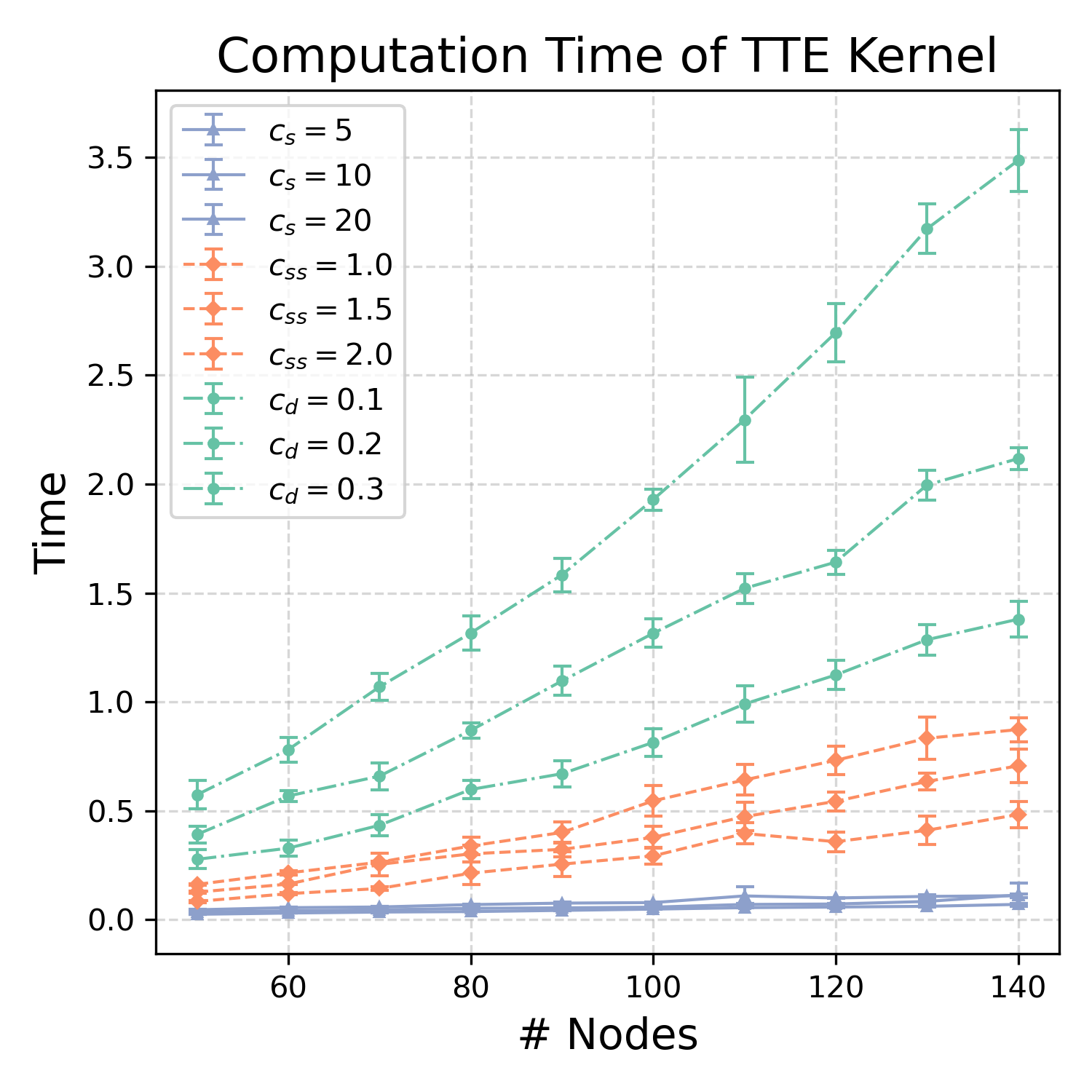}
    \end{subfigure}    
\begin{subfigure}{0.32\linewidth}
        \includegraphics[width=\linewidth]{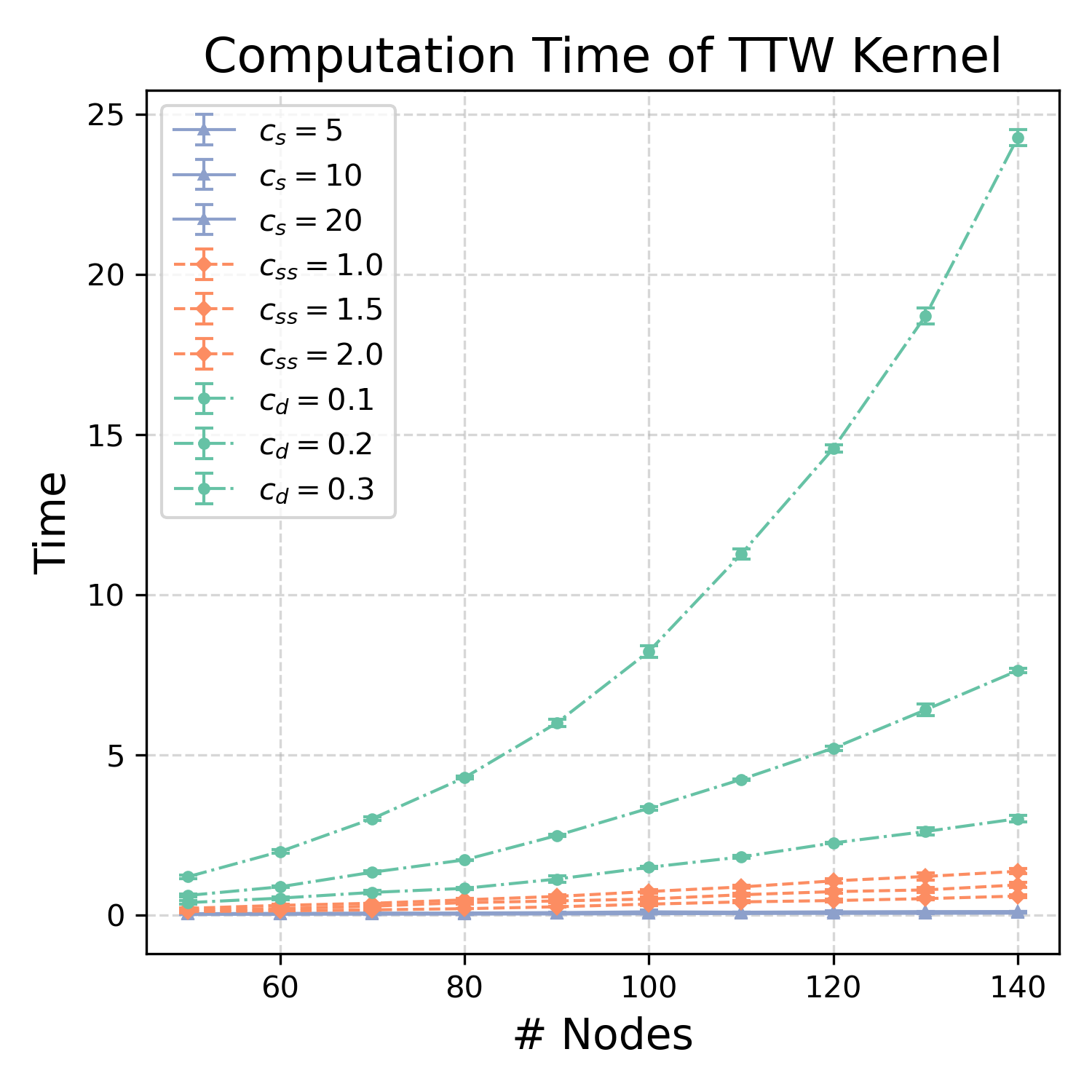}
    \end{subfigure}
    \caption{Verifying computation time. From left to right: the first figure plots genus as a function of number of nodes; the second and third figures show the computation time of kernel matrices.}
    \label{fig:time-test}
\end{figure}

We ran experiments on synthetic datasets to verify the computation runtime on our proposed classes of sparsity based on genus. The details are described in Section~\ref{sec:simulation-synthetic}.   Figure~\ref{fig:time-test} shows that the runtime to compute the TTE kernel is in alignment with the growth of the graph genus, while the runtime to compute the TTW kernel grows faster with graph density. 

\subsection{Subsampling Stability}\label{app:subsampling-stability}

We provide experimental verifications of our subsampling method described in Section~\ref{sec:psd-subsampling} to map high-dimensional SPD matrices to the same dimension by extracting a random principal submatrix.

As discussed in Section~\ref{sec:psd-subsampling}, it is necessary to consider the reduction loss in terms of the Frobenius norm since each matrix element contains topological information of the underlying metric graph. Suppose the original SPD matrix $Q$ is of size $g$ and the target size is $g_0$. Let $I\subseteq\{1,2,\ldots,g\}$ denote the index set of size $g_0$. Then in principle, we consider the following optimization problem
\begin{equation}\label{eq:max-principal}
    \max_{\substack{|I|=g_0\\ I\subseteq \{1,2,\ldots,g\}}} \sum_{i,j\in I} |Q_{ij}|^2.
\end{equation}
However, an exact solution of \eqref{eq:max-principal} can only be obtained by ranging over all index sets of size $g_0$, whose complexity is $\binom{g}{g_0}$. Therefore, instead of solving an exact solution, we simply subsample one index set $I$ randomly. 

We test the stability of the TTW and TTE kernel matrices with respect to random matrix reduction using the following experimental setting: We generate a set of $N=20$ graphs with genus $g\in\{100,200,300\}$. For each graph $G$, we compute its tropical Torelli matrices $Q(G)$ and randomly extract a principal submatrix $\widehat{Q}_\alpha(G)$ of size $g_0=[\alpha g]$, where the ratio has range $\alpha\in\{0,1,0.2,\ldots,0.9\}$. Then we compute the kernel matrix $K$ using original $Q(G_i)$, and the kernel matrix $\widehat{K}$ using reduced $\widehat{Q}_\alpha(G_i)$. We repeat the above computations for $B=10$ times and compute the following mean relative error:
\begin{equation*}
    \frac{1}{B}\sum_{i=1}^B\frac{\|\widehat{K}-K\|_F^2}{\|K\|_F^2}.
\end{equation*}
The mean relative error and standard deviation are displayed in Table~\ref{tab:subsample-table} and Table~\ref{tab:subsample-table-2}, from which we see that both kernels are stable with respect to random principal submatrix selection. We also see that the TTW kernel is more stable than the TTE kernel since it has smaller mean relative error for every reduction ratio.

\begin{table}[htbp]
  \caption{Numerical stability of random principal submatrix selection for the TTW kernel.}
  \label{tab:subsample-table}
  \centering
  \small
  \begin{tabularx}{\linewidth}{XXXXXXXXXXX}
    \toprule

$g=100$ & 0.870\linebreak(\textit{0.003}) &
0.756\linebreak(\textit{ 0.004})&
0.651\linebreak(\textit{0.003})&
0.551\linebreak(\textit{0.005})&
0.452\linebreak(\textit{0.004})&
0.355\linebreak(\textit{0.003})&
0.263\linebreak(\textit{0.004})&
0.170\linebreak(\textit{0.003})&
0.088\linebreak(\textit{0.003})\\
\midrule
$g = 200$ & 0.874\linebreak(\textit{0.002})&
0.760\linebreak(\textit{0.003})&
0.659\linebreak(\textit{0.003})&
0.555\linebreak(\textit{0.002})&
0.461\linebreak(\textit{0.001})&
0.365\linebreak(\textit{0.004})&
0.273\linebreak(\textit{0.003})&
0.181\linebreak(\textit{0.003})&
0.090\linebreak(\textit{0.002})\\
\midrule
$g=300$ &0.875\linebreak(\textit{0.002})&
0.766\linebreak(\textit{0.002})&
0.664\linebreak(\textit{0.002})&
0.562\linebreak(\textit{0.001})&
0.466\linebreak(\textit{0.002})&
0.372\linebreak(\textit{0.003})&
0.277\linebreak(\textit{0.002})&
0.183\linebreak(\textit{0.002})&
0.093\linebreak(\textit{0.001})\\

\bottomrule
\end{tabularx}
\end{table}

\begin{table}[htbp]
  \caption{Numerical stability of random principal submatrix selection for the TTE kernel.}
  \label{tab:subsample-table-2}
  \centering
  \small
  \begin{tabularx}{\linewidth}{XXXXXXXXXXX}
    \toprule

$g=100$ & 0.951\linebreak(\textit{0.004})&
0.916\linebreak(\textit{0.006})&
0.854\linebreak(\textit{0.009})&
0.771\linebreak(\textit{0.014})&
0.673\linebreak(\textit{0.020})&
0.548\linebreak(\textit{0.023})&
0.437\linebreak(\textit{0.022})&
0.309\linebreak(\textit{0.015})&
0.188\linebreak(\textit{0.025})
\\
\midrule
$g=200$ & 0.970\linebreak(\textit{0.003})&
0.932\linebreak(\textit{0.008})&
0.876\linebreak(\textit{0.006})&
0.793\linebreak(\textit{0.011})&
0.696\linebreak(\textit{0.012})&
0.600\linebreak(\textit{0.009})&
0.474\linebreak(\textit{0.016})&
0.335\linebreak(\textit{0.023})&
0.171\linebreak(\textit{0.012})\\
\midrule
$g=300$ & 0.972\linebreak(\textit{0.001})&
0.936\linebreak(\textit{0.003})&
0.864\linebreak(\textit{0.004})&
0.799\linebreak(\textit{0.012})&
0.698\linebreak(\textit{0.010})&
0.612\linebreak(\textit{0.012})&
0.474\linebreak(\textit{0.016})&
0.348\linebreak(\textit{0.020})&
0.161\linebreak(\textit{0.013})\\
\bottomrule
\end{tabularx}
\end{table}

\subsection{Kernel Positive Definiteness}\label{app:kernel-stability}

Recall that a kernel function $k$ is \emph{positive definite} if and only if for any $x_1,\ldots,x_N\in\mathcal{X}$, the kernel matrix $[k(x_i,x_j)]_{N\times N}$ is positive semi-definite (Section~\ref{app:kernel-background}). To test the positive definiteness of TTW and TTE kernels, we examine the smallest nonzero eigenvalues of their corresponding kernel matrices on different sets of data. Here we set a threshold $\mathrm{tol} = 10^{-6}$ and an eigenvalue $\lambda$ is considered as nonzero if $|\lambda|>\mathrm{tol}$.

\begin{figure}[htbp]
    \centering
    \begin{subfigure}{0.45\linewidth}
        \includegraphics[width=\linewidth]{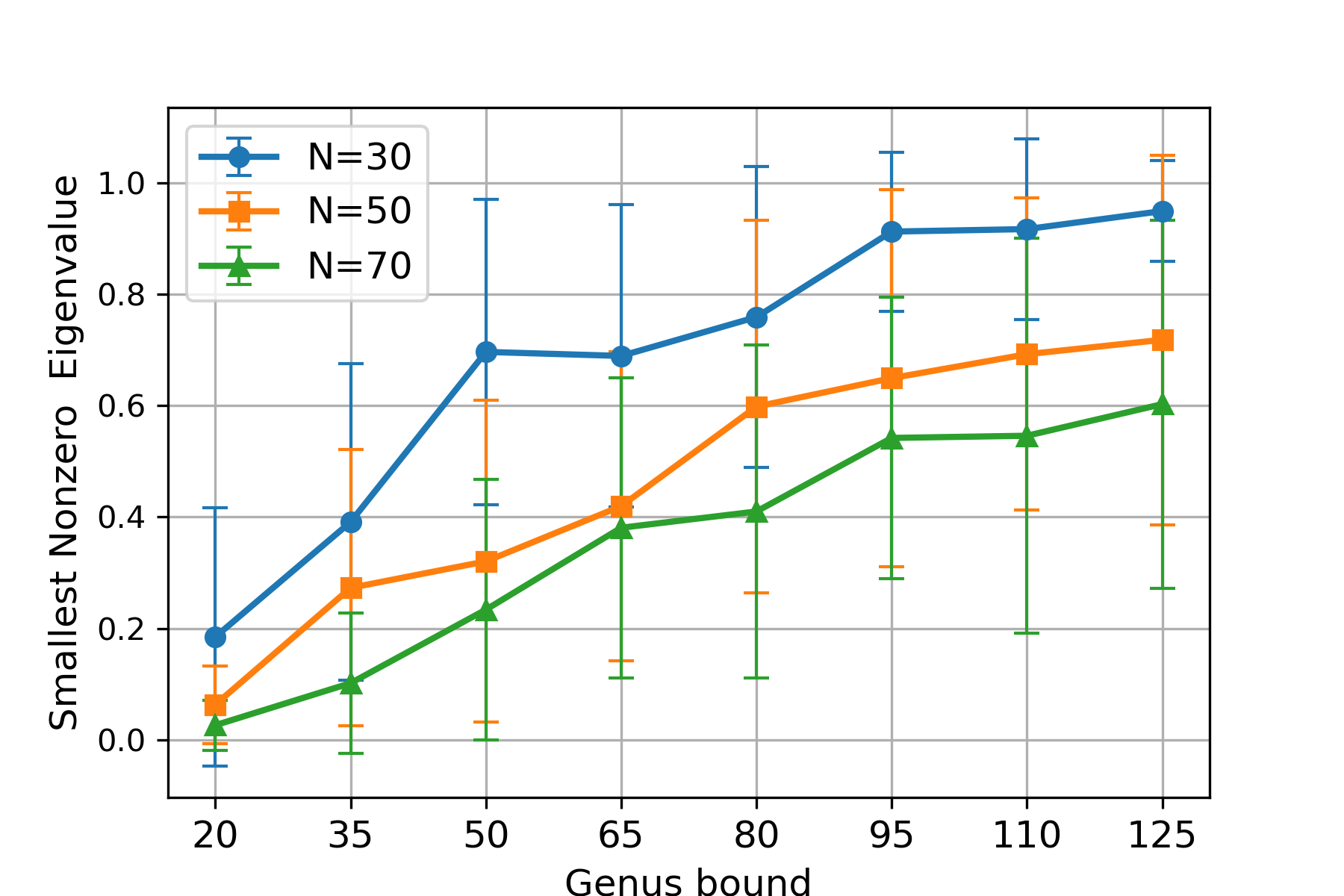}
        \caption{TTW}
    \end{subfigure}
    \begin{subfigure}{0.45\linewidth}
        \includegraphics[width=\linewidth]{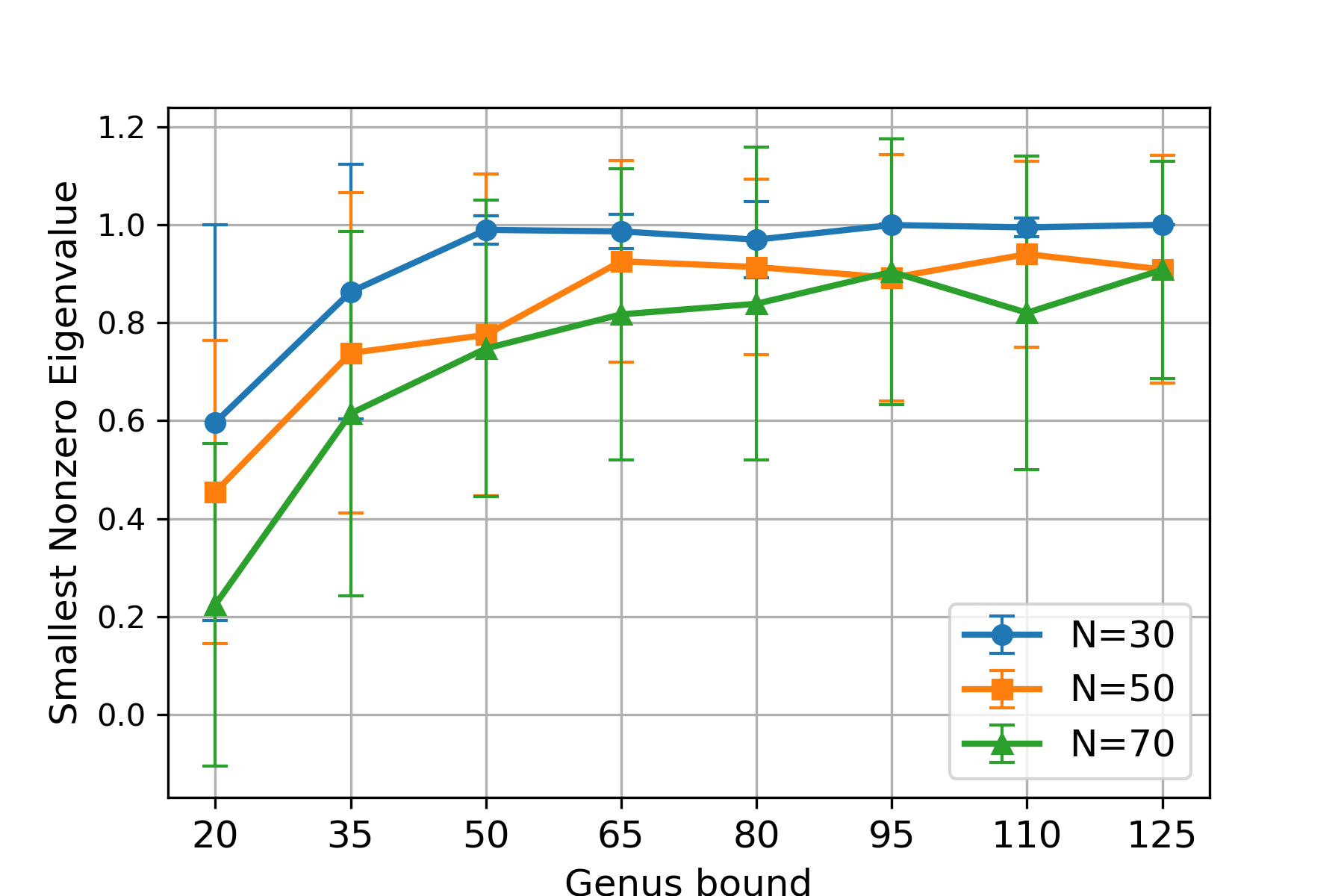}
        \caption{TTE}
    \end{subfigure}    

    \caption{Smallest nonzero eigenvalues of kernel matrices v.s. graph genus ($N\in\{30,50,70\}$)}
    \label{fig:small-vs-g}
\end{figure}

First we test how the smallest nonzero eigenvalue of the kernel matrices varies with the genus bound $g_0$. Specifically, for a fixed number of graphs $N\in\{30,50,70\}$, we generate $N$ random weighted graphs of genus $g\le g_0$, where $g_0$ ranges from $20$ to $125$. For each configuration, we compute the corresponding kernel matrix and its smallest nonzero eigenvalue. The experiment is repeated $B=20$ times for each value of $N$. The mean and standard deviation of the smallest nonzero eigenvalues are presented in Figure~\ref{fig:small-vs-g}. The results show that, for each  fixed $N$, the smallest nonzero eigenvalues of both kernel matrices tend to increase with respect to the genus bound $g_0$. Moreover, for large genus bounds, the smallest nonzero eigenvalues of both kernel matrices remain positive.    

\begin{figure}[htbp]
    \centering
    \begin{subfigure}{0.45\linewidth}
        \includegraphics[width=\linewidth]{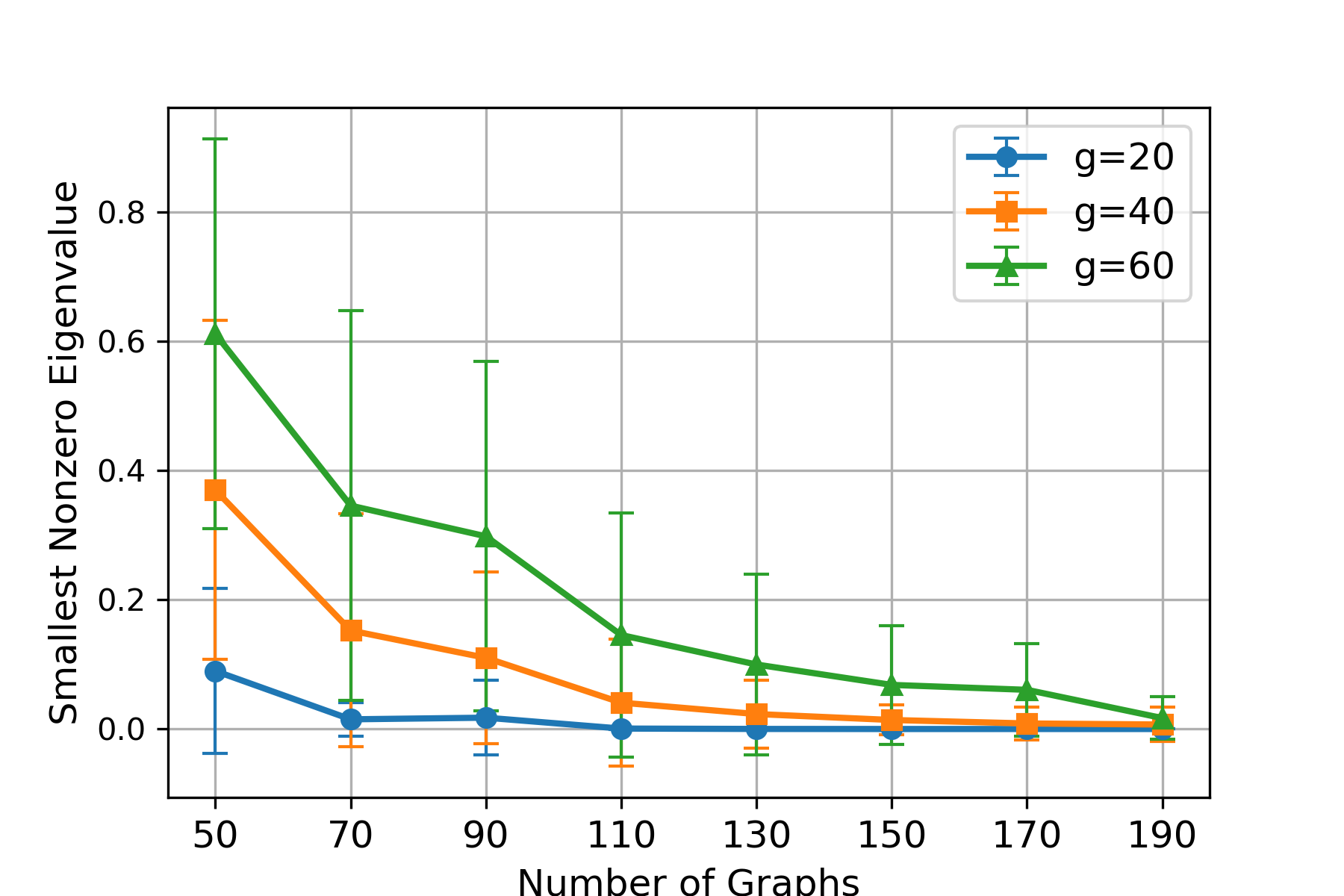}
        \caption{TTW}
    \end{subfigure}
    \begin{subfigure}{0.45\linewidth}
        \includegraphics[width=\linewidth]{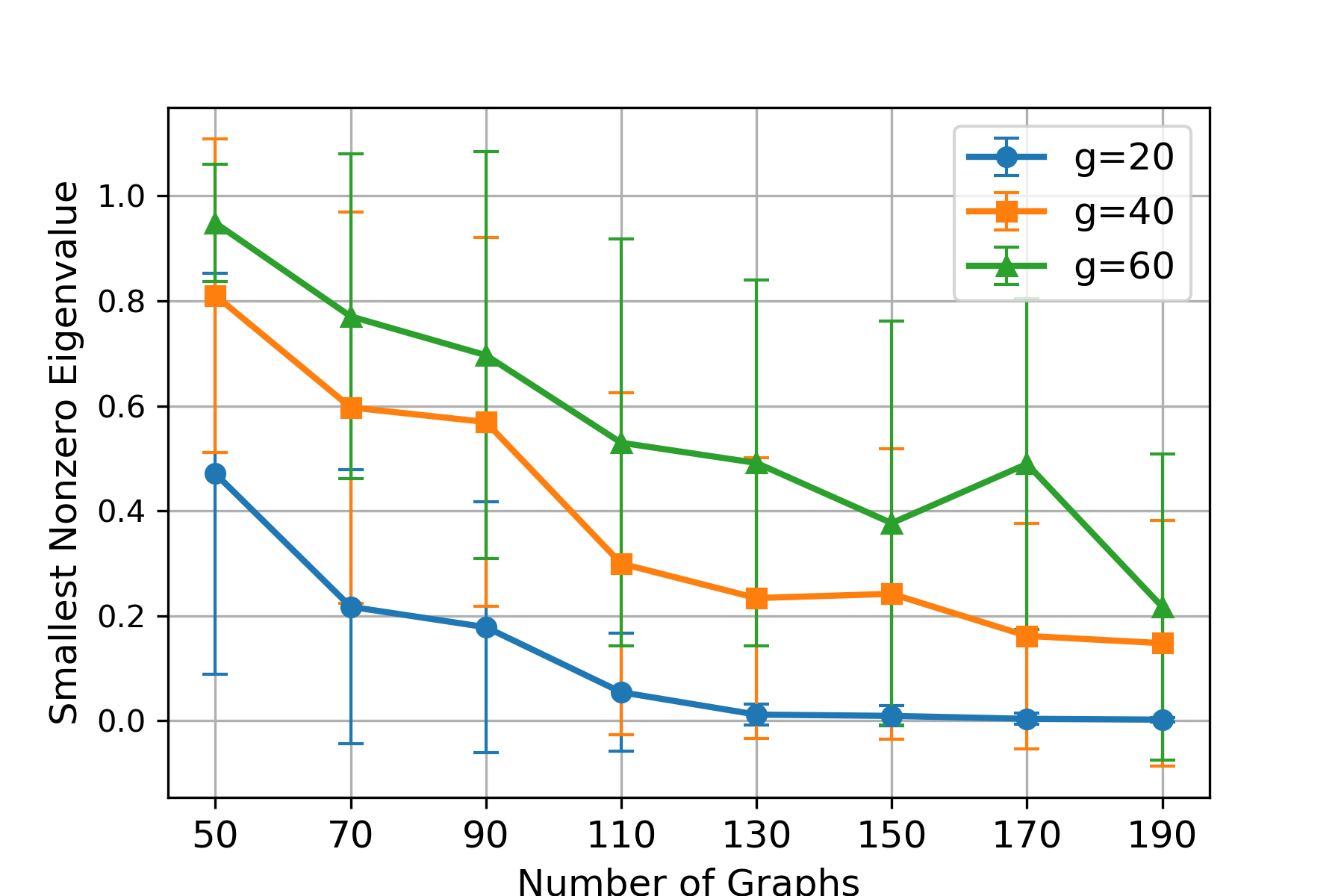}
        \caption{TTE}
    \end{subfigure}    

    \caption{Smallest nonzero eigenvalues of kernel matrices v.s. number of graphs ($g_0\in\{20,40,60\}$)}
    \label{fig:small-vs-N}
\end{figure}

We then test how the smallest nonzero eigenvalue of the kernel matrices varies with the number of graphs $N$. Similarly, for a fixed genus bound $g_0\in\{20,40,60\}$, we generate $N$ random weighted graphs of genus $g\le g_0$, where $N$ ranges from $50$ to $190$. The experiment is repeated $B=20$ times for each value of $g_0$. The mean and standard deviation of the smallest nonzero eigenvalues are presented in Figure~\ref{fig:small-vs-N}. The results show that, for each  fixed $g_0$, the smallest nonzero eigenvalues of both kernel matrices tend to decrease with respect to the number of graphs $N$ which is also the size of the kernel matrix. For the TTE kernel, the smallest nonzero eigenvalue is always positive; for the TTW kernel, we observe small negative eigenvalues when $g_0$ is small and $N$ is large. However, we do not observe significant instability of the TTW kernel.

\subsection{Stability Under Edge Length Perturbations}\label{app:length-stability}

We investigate the robustness of the TTW and TTE kernels with respect to edge length perturbations. Specifically, for a given graph $G$, we assume an additive noise model applied independently to each edge
\begin{equation}\label{eq:additive-model}
    \ell(e_i) = \bar{\ell}(e_i) + \epsilon_i, \quad \epsilon_i\sim\mathrm{Unif}(0,R).
\end{equation}
In general, suppose that the edge length matrix is perturbed by $L+\Delta L$. Then, as shown in \eqref{eq:matrix-Q} in the construction of the tropical Torelli matrix, we have
\begin{equation*}
    \Delta Q = M_T\Delta L_T M^\top_T + \Delta L_g.
\end{equation*}
Therefore the error satisfies $\|\Delta Q\|\le \|M_T\|^2\|\Delta L_T\|+\|\Delta L_g\|$, which is proportional to the noise  magnitude $R$ under \eqref{eq:additive-model}.

To evaluate the variation of kernel matrices with respect to edge length perturbations, we adopt the following experimental setting: for fixed number of graphs $N$ and fixed genus bound $g_0$, we randomly sample $N$ graphs with genus no greater than $g_0$. Then we compute the true kernel matrix $K_{\mathrm{true}}$. For the same set of weighted graphs, we add noise to every edge according to \eqref{eq:additive-model}, with noise magnitude $R$ ranging from $0.01$ to $5.12$. After obtaining the noisy kernel matrix $K_{\mathrm{noisy}}$, we compute the relative error $\|K_{\mathrm{noisy}}-K_{\mathrm{true}}\|^2_F/\|K_{\mathrm{true}}\|^2_F$.

\begin{figure}[htbp]
    \centering
    \begin{subfigure}{0.45\linewidth}
        \includegraphics[width=\linewidth]{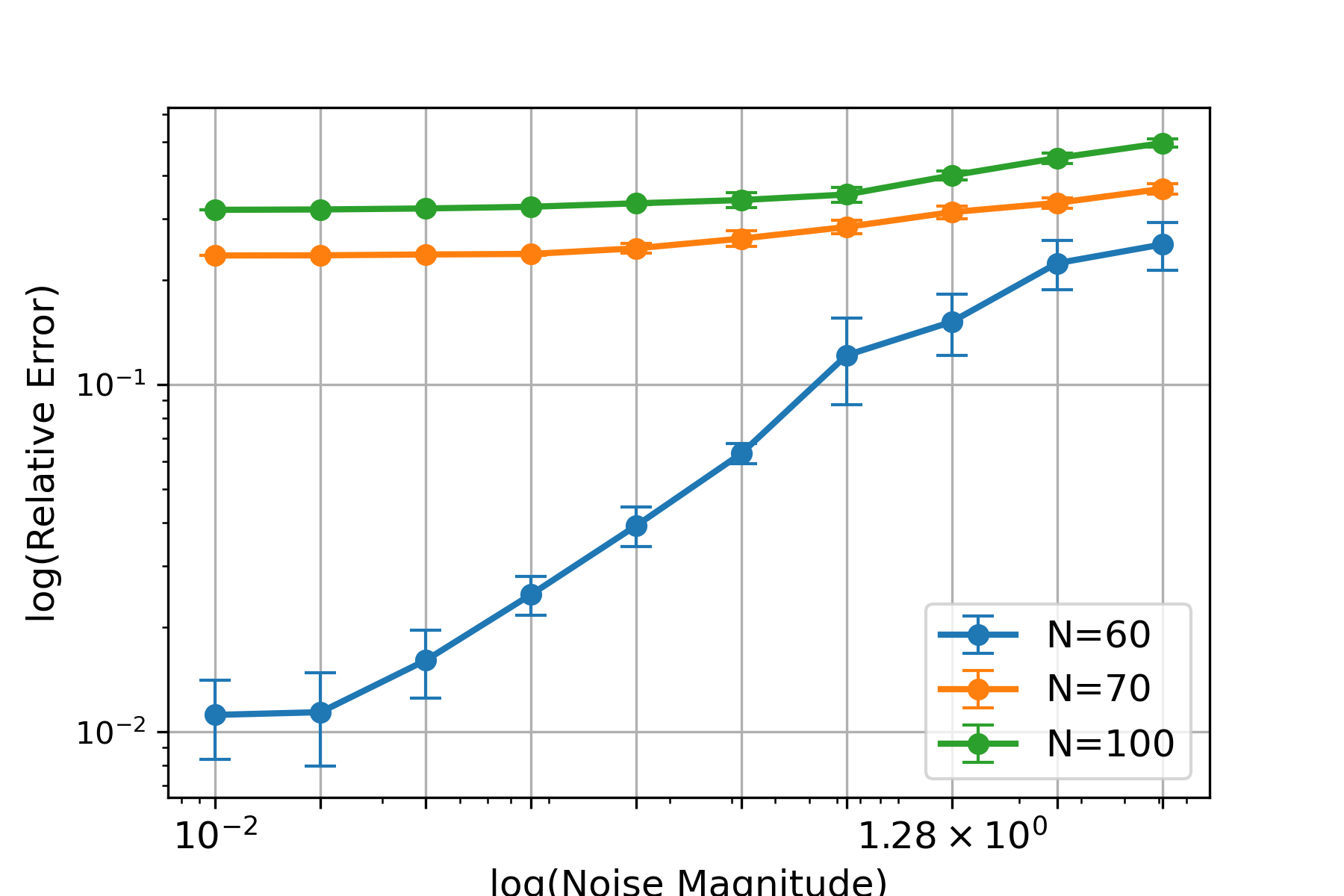}
        \caption{TTW}
    \end{subfigure}
    \begin{subfigure}{0.45\linewidth}
        \includegraphics[width=\linewidth]{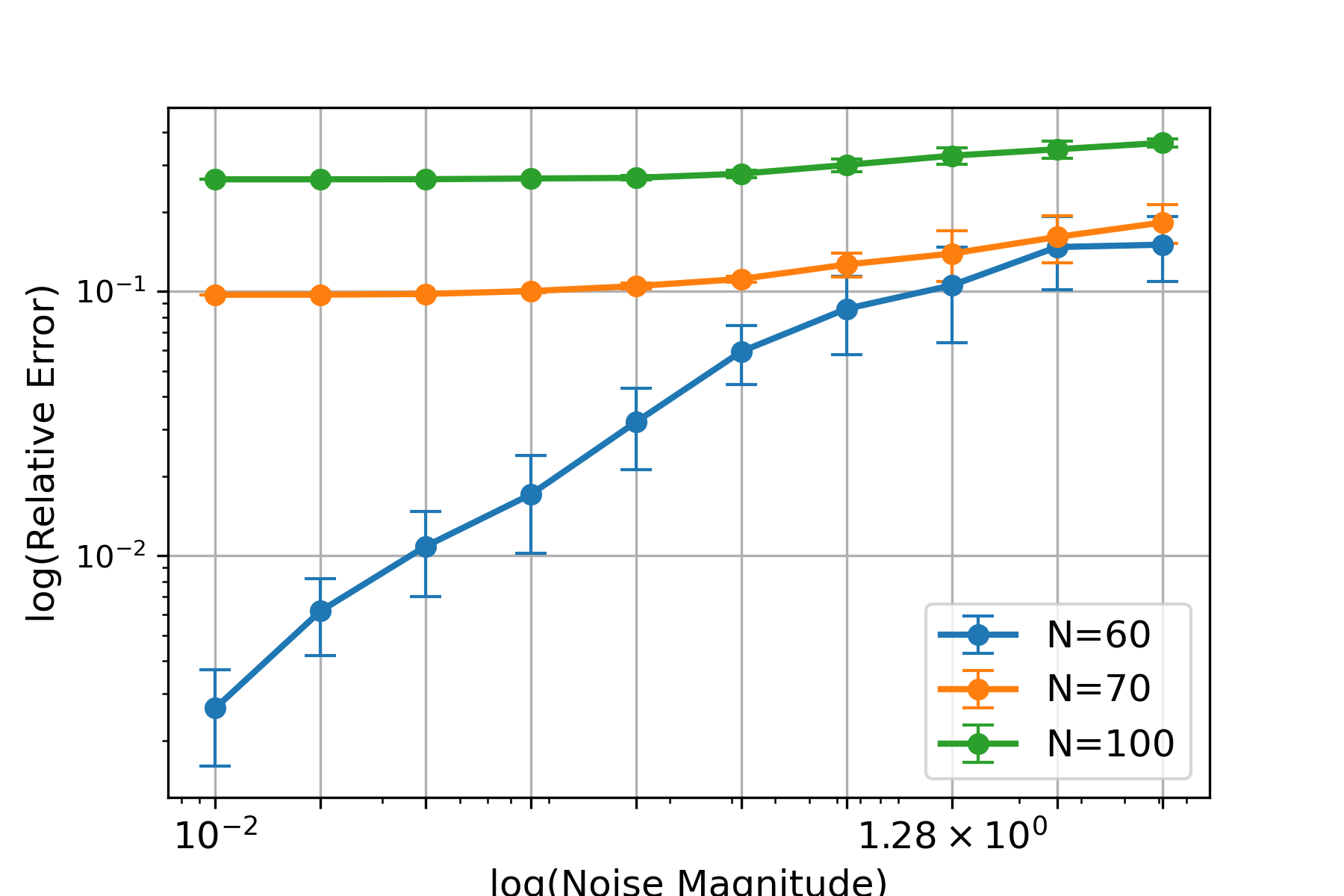}
        \caption{TTE}
    \end{subfigure}    

    \caption{Relative error of kernel matrices with respect to edge length perturbations ($N\in\{60,70,100\}$)}
    \label{fig:noise-vary-N}
\end{figure}

\begin{figure}[htbp]
    \centering
    \begin{subfigure}{0.45\linewidth}
        \includegraphics[width=\linewidth]{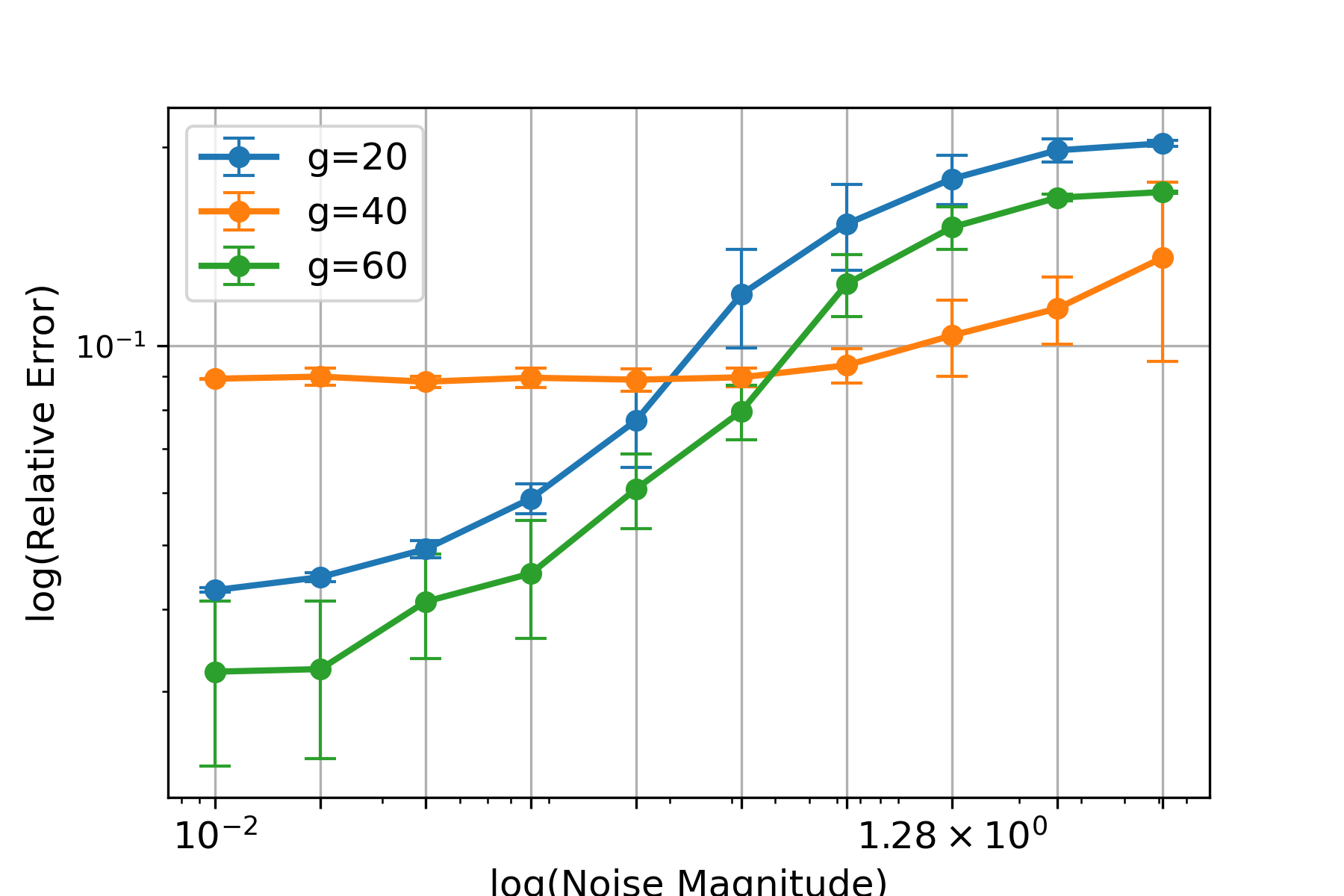}
        \caption{TTW}
    \end{subfigure}
    \begin{subfigure}{0.45\linewidth}
        \includegraphics[width=\linewidth]{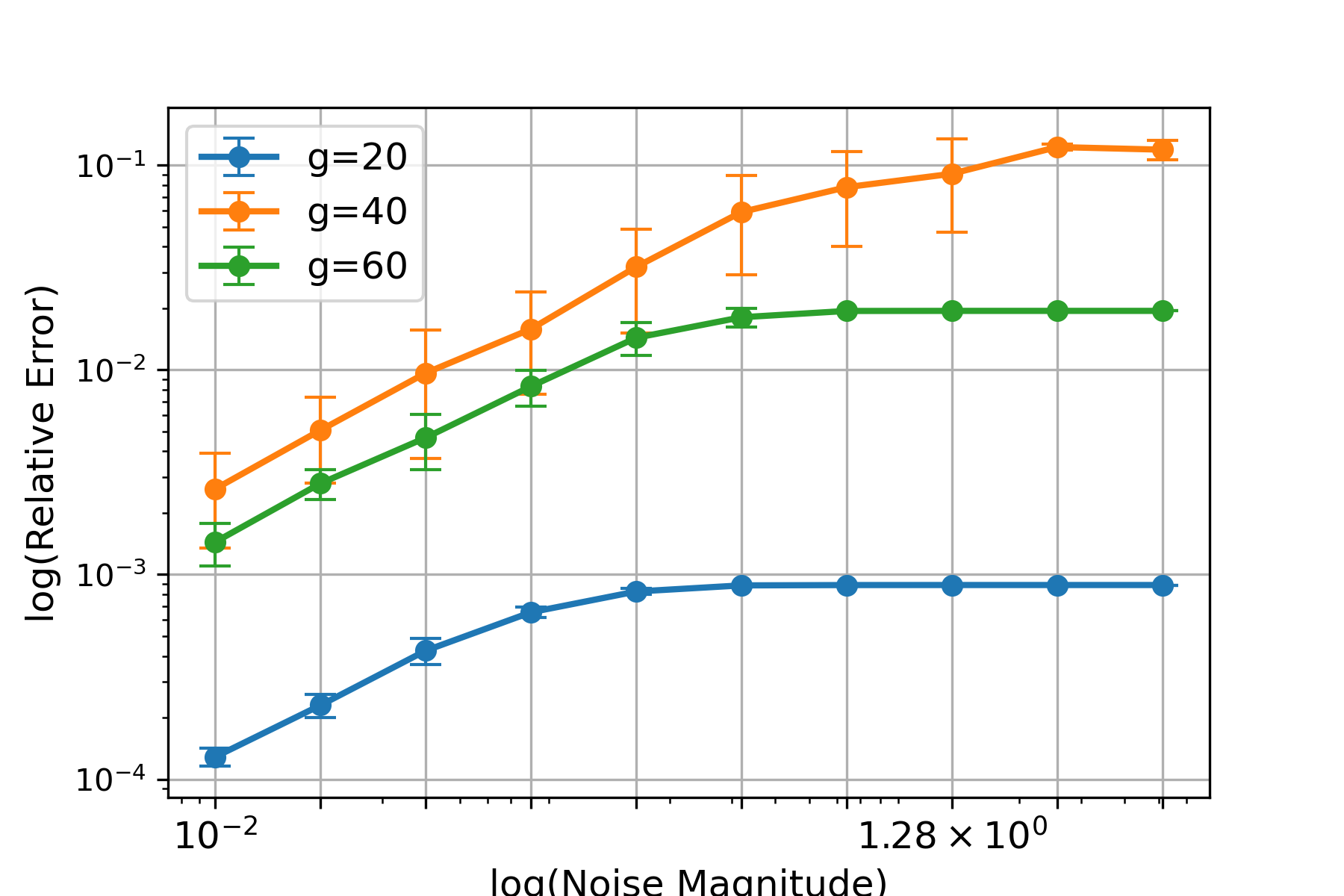}
        \caption{TTE}
    \end{subfigure}    

    \caption{Relative error of kernel matrices with respect to edge length perturbations ($g_0\in\{20,40,60\}$)}
    \label{fig:noise-vary-g}
\end{figure}

Figure~\ref{fig:noise-vary-N} shows the log-log plot of relative errors with respect to noise magnitudes for fixed $g_0=30$ and $N\in\{60,70,100\}$; Figure~\ref{fig:noise-vary-g} shows the corresponding log-log plot for fixed $N=30$ and $g_0\in\{20,40,60\}$. From both figures, we observe that the TTW and TTE kernels are robust to edge-length perturbations.

\newpage
\section{Additional Experimental Results on Benchmark Datasets}\label{app: benchmark-detail}

\subsection{Compared Graph Kernels}\label{app:compare-kernels}

We list the graph kernels used for comparison in the benchmark dataset experiments, explain their core ideas, and describe the hyperparameters that require tuning.

\begin{itemize}
   \item \textbf{The Edge Histogram (EH) kernel}. The EH kernel compares graphs based on the histogram of discrete edge weights or labels \citep{sugiyama2015halting}. It is the most computationally efficient graph kernel. However it cannot capture the graph topology. The EH kernel does not have any hyperparameters.
    \item \textbf{The Graphlet Sampling (GS) kernel}. The GS kernel measures the similarity between graphs via counting matching graphlets, i.e., small subgraphs with $k$ vertices \citep{shervashidze2009efficient}. Since enumerating all graphlets is computationally expensive, the counting and matching steps are commonly approximated by sampling a specified number of graphlets up to a given size. The hyperparameters for GS kernel are (1) maximal size of a graphlet, which is commonly set as $k=3\sim 6$; (2) the maximal number of samples which we set to be $B=300\sim 500$. 
    \item \textbf{The ODD-STh (OS) kernel} The OS kernel compares graphs based on the Directed Acyclic Graph (DAG) decomposition \citep{da2012tree}. For each node in a graph, a breadth-first search (BFS) tree or DAG is constructed up to a fixed depth, capturing the local neighborhood structure in a rooted format. For unlabeled (or constantly labeled) graphs, the OS kernel is able to capture the structural differences. The hyperparameter for the OS kernel is the maximal depth of the subtree pattern, which is commonly set to be $h=2\sim5$. 
    \item \textbf{The Shortest Path (SP) kernel}. The SP kernel compares graphs based on the shortest path lengths between all pairs of nodes \citep{borgwardt2005shortest}. The original SP kernel uses a base kernel function on the node labels at the path endpoints and the path lengths, which is computationally expensive. We use the simplified version using the histogram of shortest paths, which is also the implemented version in \texttt{GraKeL}. The SP kernel does not have any hyperparameters.
    \item \textbf{The Weisfeiler--Lehman (WL) Kernel}. The WL kernel is a fast and powerful graph kernel based on the Weisfeiler--Lehman graph isomorphism test algorithm \citep{shervashidze2011weisfeiler}. Its core idea is to iteratively relabel each node by aggregating the multiset of labels of its neighbors, thus capturing  local neighborhood structures. In particular, we will use the WL subtree kernel implemented in \texttt{GraKeL} which is widely used in the literature. The hyperparameter for the WL kernel is the number of refinement iterations, which is commonly set to be $h=5\sim 10$.
    \item \textbf{The $k$-Core Decomposition ($k$CD) kernel}. The $k$CD kernel compares graphs based on the $k$-core decomposition of a graph \citep{nikolentzos2018degeneracy}. The $k$-core of a graph is the maximal subgraph in which all nodes have degree at least $k$. This decomposition assigns a core number to each node, indicating its topological depth or centrality within the graph. The $k$CD kernel uses the distribution of these core numbers as the basis for comparison between graphs. The hyperparameter for the $k$CD kernel is the maximum core number, which is typically set to be $k=3\sim 6$.
\end{itemize}

We also list the theoretical computational complexity for aforementioned graph kernels in Table~\ref{tab:complexity}.

\subsection{SVM Settings}

For the graph classification task, we use SVM with precomputed kernels. Specifically, we use the \texttt{SVC} implementation from \texttt{scikit-learn}, which is based on the soft-margin SVM (Section~\ref{app:svm-intro}).  The solver is limited to a maximum of 10,000 iterations to ensure convergence without excessive computation time.

The evaluation of the classifier is performed using 10-fold cross-validation. Namely, the dataset is partitioned into ten disjoint subsets (folds), with each fold serving as a validation set once while the remaining nine folds form the training set. The samples are shuffled prior to splitting to reduce bias and improve the representativeness of each fold, and a fixed random seed ensures reproducibility of the splits. We use accuracy as the scoring metric.

\subsection{Dataset Statistics}\label{app:data-stat}

\begin{table}[H]
\caption{Benchmark dataset statistics. In the table, $N$ is the number of graphs in a dataset; $\bar{n},\bar{m},\bar{g}$ are the average numbers of nodes, edges, and genus. In the sparsity column, ``S'', ``SS'' and ``D'' are short for ``Sparse'', ``Semi-Sparse'' and ``Dense''.}
\label{tab:data-stats-full}
    \centering
\begin{tabularx}{\textwidth}{XYYYYYY}
\toprule
 Name & $N$ & $\bar{n}$ \linebreak (\textit{std.}) & $\bar{m}$\linebreak(\textit{std.}) & $\bar{g}$\linebreak (\textit{std.}) & $\log(\bar{g}/\bar{n})$& Sparsity \\
\midrule
\texttt{AIDS} & 2000 & 15.59 \linebreak (\textit{13.59}) & 16.20 \linebreak (\textit{15.01}) & 1.63 \linebreak (\textit{1.67}) & -2.26& S \\

\texttt{BZR} & 405 & 35.75 \linebreak (\textit{7.26}) & 38.36 \linebreak (\textit{7.69}) & 3.61 \linebreak (\textit{0.87}) & -2.29& S \\

\texttt{BZR-MD} & 306 & 21.30 \linebreak (\textit{4.19}) & 225.06 \linebreak (\textit{85.83}) & 204.75 \linebreak (\textit{81.70}) & 2.26& D \\

\texttt{COIL-DEL} & 3900 & 21.54 \linebreak (\textit{13.22}) & 54.24 \linebreak (\textit{38.47}) & 33.70 \linebreak (\textit{25.27}) & 0.45& SS \\

\texttt{COX2} & 467 & 41.22 \linebreak (\textit{4.03}) & 43.45 \linebreak (\textit{4.27}) & 3.22 \linebreak (\textit{0.42}) & -2.55& S \\

\texttt{COX2-MD} & 303 & 26.28 \linebreak (\textit{2.46}) & 335.12 \linebreak (\textit{65.19}) & 309.84 \linebreak (\textit{62.75}) & 2.47& D \\

\texttt{DD} & 1178 & 284.32 \linebreak (\textit{272.00}) & 715.66 \linebreak (\textit{693.91}) & 432.36 \linebreak (\textit{426.25}) & 0.42& SS \\

\texttt{DHFR} & 756 & 42.43 \linebreak (\textit{9.06}) & 44.54 \linebreak (\textit{9.25}) & 3.12 \linebreak (\textit{0.65}) & -2.61& S \\

\texttt{DHFR-MD} & 393 & 23.87 \linebreak (\textit{4.50}) & 283.02 \linebreak (\textit{106.85}) & 260.15 \linebreak (\textit{102.40}) & 2.39& D \\

\texttt{ENZYMES} & 600 & 32.46 \linebreak (\textit{14.87}) & 62.14 \linebreak (\textit{25.50}) & 30.75 \linebreak (\textit{13.01}) & -0.05 &S \\

\texttt{ER-MD} & 446 & 21.33 \linebreak (\textit{6.01}) & 234.85 \linebreak (\textit{133.47}) & 214.52 \linebreak (\textit{127.63}) & 2.31& D \\

\texttt{FRANKEN\linebreak-STEIN} & 4337 & 16.83 \linebreak (\textit{10.45}) & 17.88 \linebreak (\textit{11.60}) & 2.07 \linebreak (\textit{1.67}) & -2.10& S \\

\texttt{IMDB\linebreak-BINARY} & 1000 & 19.77 \linebreak (\textit{10.06}) & 96.53 \linebreak (\textit{105.60}) & 77.76 \linebreak (\textit{98.22}) & 1.37& D \\

\texttt{IMDB\linebreak-MULTI} & 1500 & 13.00 \linebreak (\textit{8.52}) & 65.94 \linebreak (\textit{110.78}) & 53.93 \linebreak (\textit{103.84}) & 1.42& D \\

\texttt{MSRC-9} & 221 & 40.58 \linebreak (\textit{5.27}) & 97.94 \linebreak (\textit{15.14}) & 58.36 \linebreak (\textit{10.18}) & 0.36& SS \\

\texttt{MSRC-21} & 563 & 77.52 \linebreak (\textit{12.27}) & 198.32 \linebreak (\textit{36.81}) & 121.80 \linebreak (\textit{25.20}) & 0.45& SS \\

\texttt{MSRC-21C} & 209 & 40.28 \linebreak (\textit{5.82}) & 96.60 \linebreak (\textit{16.44}) & 57.33 \linebreak (\textit{10.86}) & 0.35& SS \\

\texttt{MUTAG} & 188 & 17.93 \linebreak (\textit{4.58}) & 19.79 \linebreak (\textit{5.68}) & 2.86 \linebreak (\textit{1.30}) & -1.84& S \\

\texttt{NCI1} & 4110 & 29.76 \linebreak (\textit{13.55}) & 32.30 \linebreak (\textit{14.93}) & 3.62 \linebreak (\textit{1.98}) & -2.11& S \\

\texttt{NCI109} & 4127 & 29.57 \linebreak (\textit{13.55}) & 32.13 \linebreak (\textit{14.96}) & 3.64 \linebreak (\textit{1.98}) & -2.09& S \\

\texttt{PROTEINS} & 1113 & 39.05 \linebreak (\textit{45.76}) & 72.82 \linebreak (\textit{84.60}) & 34.84 \linebreak (\textit{40.56}) & -0.11& S \\

\texttt{REDDIT\linebreak-BINARY} & 2000 & 429.62 \linebreak (\textit{554.05}) & 497.75 \linebreak (\textit{622.99}) & 70.61 \linebreak (\textit{85.56}) & -1.81& S \\

\texttt{REDDIT\linebreak-MULTI-5K} & 4999 & 508.51 \linebreak (\textit{452.57}) & 594.87 \linebreak (\textit{566.77}) & 90.08 \linebreak (\textit{133.61}) & -1.73& S \\
\bottomrule
\end{tabularx}

\end{table}

\subsection{Full Test Results}\label{app:full-test}

\begin{table}[H]
    \caption{Classification accuracy of 10-fold cross-validation on benchmark datasets \textbf{without} labels. In the table, the accuracy is measured by percentage (\%); ``M'' stands for out of memory. The max memory is set to be 128GB.}
    \label{tab:acc-full}
    \centering
\begin{tabularx}{\textwidth}{XYYYYYYYY}
\toprule
Name & EH\linebreak (\textit{std.}) & GS\linebreak (\textit{std.}) & OS\linebreak (\textit{std.}) & SP\linebreak (\textit{std.}) & WL\linebreak (\textit{std.}) & $k$CD\linebreak (\textit{std.}) & TTE\linebreak (\textit{std.}) & TTW\linebreak (\textit{std.}) \\
\midrule
\texttt{AIDS} & 80.00\linebreak(\textit{2.44}) & 98.80\linebreak(\textit{0.68}) & 57.10\linebreak(\textit{37.47}) & 81.00\linebreak(\textit{2.95}) & 
\textbf{99.70}\linebreak(\textit{0.40})
&
80.10\linebreak(\textit{2.07}) & 93.40\linebreak(\textit{1.37}) & 90.50\linebreak(\textit{1.94}) \\

\texttt{BZR} & 78.78\linebreak(\textit{5.44}) & 57.54\linebreak(\textit{10.70}) & 76.55\linebreak(\textit{11.18}) & 78.74\linebreak(\textit{3.98}) & 
78.82\linebreak(\textit{6.08})
&
78.79\linebreak(\textit{6.69}) & 82.74\linebreak(\textit{7.67}) & \textbf{83.71}\linebreak(\textit{4.80}) \\

\texttt{BZR-MD} & 48.72\linebreak(\textit{7.44}) & 51.00\linebreak(\textit{8.82}) & 61.53\linebreak(\textit{12.77}) & 48.71\linebreak(\textit{7.13}) & 
\textbf{64.72}\linebreak(\textit{6.20})
&
48.68\linebreak(\textit{6.05}) & 48.13\linebreak(\textit{8.89})& 51.26\linebreak(\textit{4.95}) \\

\texttt{COIL-DEL} & 1.00\linebreak(\textit{0.33}) & \textbf{15.03}\linebreak(\textit{1.65}) & 14.72\linebreak(\textit{2.47}) & 1.00\linebreak(\textit{0.60}) 
& 
5.56\linebreak(\textit{0.72})
& 1.01\linebreak(\textit{0.51}) 
&
1.97\linebreak(\textit{0.66}) & 3.10\linebreak(\textit{1.03}) \\

\texttt{COX2} & 78.15\linebreak(\textit{3.49}) & 76.23\linebreak(\textit{3.85}) & 72.62\linebreak(\textit{8.60}) & 78.17\linebreak(\textit{4.33}) & 
78.19\linebreak(\textit{5.67})
&
78.16\linebreak(\textit{3.26}) & 78.17\linebreak(\textit{5.14}) & \textbf{78.21}\linebreak(\textit{7.69}) \\ 

\texttt{COX2-MD} & 47.81\linebreak(\textit{8.51}) & 48.85\linebreak(\textit{6.42}) & 51.83\linebreak(\textit{7.92}) & 48.20\linebreak(\textit{6.53}) & 
46.22\linebreak(\textit{10.83})
&
48.54\linebreak(\textit{6.21}) & \textbf{52.20}\linebreak(\textit{6.58}) & 47.20\linebreak(\textit{10.68}) \\ 

\texttt{DD} & 41.35\linebreak(\textit{4.41}) & 51.69\linebreak(\textit{8.87}) & M & M
& 
49.57\linebreak(\textit{9.05})
& M & \textbf{58.66}\linebreak(\textit{3.35}) & M \\ 

\texttt{DHFR} & 60.96\linebreak(\textit{3.86}) & 56.10\linebreak(\textit{6.61}) & \textbf{74.85}\linebreak(\textit{4.88}) & 61.00\linebreak(\textit{4.14}) & 
60.90\linebreak(\textit{4.56})
&60.99\linebreak(\textit{4.55}) & 58.86\linebreak(\textit{6.19}) & 60.86\linebreak(\textit{7.08}) \\ 

\texttt{DHFR-MD} & 67.90\linebreak(\textit{5.39}) & 60.25\linebreak(\textit{17.33}) & 66.13\linebreak(\textit{7.37}) & 67.00\linebreak(\textit{6.50}) & 
67.94\linebreak(\textit{5.42})
&67.90\linebreak(\textit{5.46}) & 67.94\linebreak(\textit{7.54}) & \textbf{67.95}\linebreak(\textit{7.40}) \\ 

\texttt{ENZYMES} & 16.00\linebreak(\textit{2.60}) & 21.67\linebreak(\textit{6.45}) & \textbf{22.00}\linebreak(\textit{6.74}) & 16.67\linebreak(\textit{6.32}) & 
17.33\linebreak(\textit{3.96})
&16.83\linebreak(\textit{5.40}) & 12.67\linebreak(\textit{3.00}) & 18.67\linebreak(\textit{6.05}) \\ 

\texttt{ER-MD} & 49.41\linebreak(\textit{6.96}) & 46.23\linebreak(\textit{10.77}) & 58.55\linebreak(\textit{11.63}) & 59.41\linebreak(\textit{9.95}) & 
56.04\linebreak(\textit{5.30})
&59.37\linebreak(\textit{8.55}) & 59.40\linebreak(\textit{5.18}) & \textbf{59.43}\linebreak(\textit{6.58}) \\ 

\texttt{FRANKEN\linebreak-STEIN} & 49.76\linebreak(\textit{5.78}) & 52.04\linebreak(\textit{4.43}) & 50.15\linebreak(\textit{5.46}) & 44.64\linebreak(\textit{2.02}) & 
55.66\linebreak(\textit{2.19})
&44.64\linebreak(\textit{2.74}) & \textbf{62.09}\linebreak(\textit{2.48}) & 62.07\linebreak(\textit{2.99}) \\

\texttt{IMDB\linebreak-BINARY} & 50.00\linebreak(\textit{4.98}) & 47.40\linebreak(\textit{3.38}) & 55.10\linebreak(\textit{4.16}) & 46.60\linebreak(\textit{3.61}) & 
48.60\linebreak(\textit{4.43})
&50.00\linebreak(\textit{6.00}) & 47.40\linebreak(\textit{2.33}) & \textbf{60.30}\linebreak(\textit{3.58}) \\

\texttt{IMDB\linebreak-MULTI} & 33.30\linebreak(\textit{4.33}) & 34.87\linebreak(\textit{3.53}) & 34.53\linebreak(\textit{2.93}) & 33.33\linebreak(\textit{3.20}) & 
32.20\linebreak(\textit{2.27})
&33.33\linebreak(\textit{4.39}) & 31.00\linebreak(\textit{3.42}) & \textbf{40.40}\linebreak(\textit{2.35}) \\

\texttt{MSRC-9} & 7.23\linebreak(\textit{4.15}) & \textbf{20.34}\linebreak(\textit{7.30}) & 14.9\linebreak(\textit{9.47}) & 9.51\linebreak(\textit{4.29}) & 
16.76\linebreak(\textit{7.91})
&10.00\linebreak(\textit{5.68}) & 5.85\linebreak(\textit{3.42}) & 5.40\linebreak(\textit{3.84}) \\

\texttt{MSRC-21} & 3.00\linebreak(\textit{2.49}) & \textbf{9.95}\linebreak(\textit{4.07}) & 8.52\linebreak(\textit{3.16}) & 4.09\linebreak(\textit{2.27}) & 
6.23\linebreak(\textit{3.03})
&5.33\linebreak(\textit{1.77}) & 4.09\linebreak(\textit{1.80}) & 4.61\linebreak(\textit{1.17}) \\

\texttt{MSRC-21C} & 8.67\linebreak(\textit{5.33}) & 18.14\linebreak(\textit{8.42}) & 13.38\linebreak(\textit{4.63}) & 10.00\linebreak(\textit{5.81}) & 
\textbf{19.12}\linebreak(\textit{9.49})
&8.57\linebreak(\textit{7.00}) & 11.45\linebreak(\textit{8.81}) & 10.55\linebreak(\textit{6.36}) \\

\texttt{MUTAG} & 66.46\linebreak(\textit{7.92}) & 84.53\linebreak(\textit{4.55}) & 83.51\linebreak(\textit{9.38}) & 36.26\linebreak(\textit{10.65}) & 
83.48\linebreak(\textit{3.84})
&31.46\linebreak(\textit{7.23}) & 78.30\linebreak(\textit{7.84}) & \textbf{85.18}\linebreak(\textit{10.43}) \\ 

\texttt{NCI1} & 49.95\linebreak(\textit{1.85}) & 57.71\linebreak(\textit{7.93}) & 56.11\linebreak(\textit{9.56}) & 49.95\linebreak(\textit{1.77}) & 
54.48\linebreak(\textit{7.05})
&49.95\linebreak(\textit{2.91}) & 61.85\linebreak(\textit{1.96}) & \textbf{62.06}\linebreak(\textit{2.03}) \\ 

\texttt{NCI109} & 49.63\linebreak(\textit{1.56}) & 51.74\linebreak(\textit{7.44}) & 60.67\linebreak(\textit{6.15}) & 49.62\linebreak(\textit{2.03}) & 
54.93\linebreak(\textit{5.82})
&49.62\linebreak(\textit{1.58}) & \textbf{62.73}\linebreak(\textit{3.29}) & 61.76\linebreak(\textit{1.99}) \\ 

\texttt{PROTEINS} & 40.43\linebreak(\textit{4.23}) & 49.41\linebreak(\textit{8.10}) & 44.93\linebreak(\textit{4.87}) & 40.44\linebreak(\textit{4.82}) & 
46.39\linebreak(\textit{8.35})
&40.43\linebreak(\textit{4.08}) & 65.41\linebreak(\textit{3.40}) & \textbf{71.16}\linebreak(\textit{3.49}) \\

\texttt{REDDIT\linebreak-BINARY} & 50.00\linebreak(\textit{3.35}) & M & M & M
& 48.70\linebreak(\textit{7.03})
& M & 62.50\linebreak(\textit{2.97}) & \textbf{67.35}\linebreak(\textit{3.83}) \\ 

\texttt{REDDIT\linebreak-MULTI-5K} & 20.01\linebreak(\textit{2.23}) & M & M & M
&28.32\linebreak(\textit{6.43})
& M & \textbf{43.89}\linebreak(\textit{2.08}) & M \\
\bottomrule
\end{tabularx}
\end{table}

\begin{table}[H]
\centering
\caption{Evaluation of Computational Runtime. The wall time is measured in seconds. The edge histogram kernel and the Weisfeiler--Lehman kernel are excluded ($< 0.1$s). }
\begin{subtable}[t]{\linewidth}
    \caption{Computation runtime of features (\texttt{.fit} function in \texttt{GraKeL}).}
    \label{tab:fit-time-full}
    \centering
\begin{tabularx}{\textwidth}{llXXXXXX}
\toprule
Name& Sparsity& GS & OS & SP & $k$CD & TTE & TTW \\
\midrule
\texttt{AIDS}& S & 9.46  & 19.41  & 4.79  & 11.01  & 2.65  & \textbf{1.59}  \\ 

\texttt{BZR}& S & 8.29  & 3.87  & 1.87  & 5.19  & \textbf{1.08}  & 1.21  \\ 

\texttt{BZR-MD}& D & 2388.48  & \textbf{0.79}  & 1.49  & 50.44  & 14.79  & 11.56  \\ 

\texttt{COIL-DEL}& SS & 1026.84  & 244.50  & \textbf{17.03}  & 77.47  & 25.78  & 34.13  \\

\texttt{COX2}& S & 10.37  & 5.09  & 3.71  & 7.90  & 1.50  & \textbf{1.20}  \\ 

\texttt{COX2-MD}& D & 9842.24  & \textbf{1.35}  & 2.50  & 110.06  & 15.42  & 27.31  \\ 

\texttt{DD}& SS & 8053.55  & M & M & M & \textbf{386.69}  & M \\ 

\texttt{DHFR}& S & 17.31  & 11.62  & 6.60  & 17.69  & \textbf{1.41}  & 2.57  \\ 

\texttt{DHFR-MD}& D & 4819.32  & \textbf{1.43}  & 2.73  & 113.35  & 25.23  & 23.94  \\ 

\texttt{ENZYMES}& S & 60.51  & 11.67  & 6.55  & 15.08  & \textbf{2.96}  & 6.35  \\ 

\texttt{ER-MD}& D & 5326.07  & \textbf{1.41}  & 3.22  & 89.17  & 21.58  & 19.51  \\ 

\texttt{FRANKENSTEIN}& S & 44.41  & 86.96  & 6.07  & 34.73  & \textbf{3.86}  & 7.78  \\

\texttt{IMDB-BINARY}& D & 8332.79  & \textbf{2.55}  & 3.49  & 69.06  & 10.22  & 21.69  \\ 

\texttt{IMDB-MULTI}& D & 12346.53  & \textbf{2.24}  & 2.90  & 111.16  & 10.98  & 11.83  \\ 

\texttt{MSRC-9}& SS & 110.96  & 3.26  & 3.91  & 9.41  & \textbf{2.16}  & 2.31  \\ 

\texttt{MSRC-21}& SS & 1222.51  & 50.79  & 36.71  & 111.15  & 24.60  & \textbf{15.29}  \\ 

\texttt{MSRC-21C}& SS & 104.49  & 3.69  & \textbf{1.75}  & 8.81  & 3.52  & 3.80  \\ 

\texttt{MUTAG}& S & 1.31  & 0.38  & \textbf{0.30}  & 1.10  & 0.89  & 0.96  \\ 

\texttt{NCI1}& S & 48.30  & 173.60  & 15.51  & 50.22  & \textbf{7.34}  & 10.86  \\ 

\texttt{NCI109}& S& 57.46  & 174.41  & 31.63  & 50.12  & 7.50  & \textbf{7.04}  \\ 

\texttt{PROTEINS}& S & 163.99  & 120.98  & 22.70  & 144.59  & 11.44  & \textbf{9.80}  \\ 

\texttt{REDDIT-BINARY}& S & M & M & M & M & \textbf{136.88}  & 148.57  \\ 

\texttt{REDDIT-MULTI-5K}& S & M & M & M & M & \textbf{374.67}& M \\
\bottomrule
\end{tabularx}
\end{subtable}
\smallskip

\begin{subtable}[b]{\linewidth}
    \caption{Computation runtime of full kernel matrices (\texttt{.transform} function in \texttt{GraKeL}).}
    \label{tab:transform-time-full}
    \centering
\begin{tabularx}{\textwidth}{llXXXXXX}
\toprule
Name& Sparsity& GS & OS & SP & $k$CD & TTE & TTW \\
\midrule
\texttt{AIDS}& S & 9.59  & 44.30  & 106.51  & 231.80  & \textbf{2.53}  & 175.45  \\

\texttt{BZR}& S & 8.57  & 6.49  & 3.63  & 11.81  & \textbf{0.73}  & 9.70  \\

\texttt{BZR-MD}& D & 2328.25  & 0.97  & \textbf{0.91}  & 63.16  & 19.21  & 1410.91  \\

\texttt{COIL-DEL}& SS & 939.85  & 684.63  & 1132.91  & 3631.45  & \textbf{30.89}  & 5584.44  \\ 

\texttt{COX2}& S & 10.59  & 7.66  & 10.20  & 21.16  & \textbf{1.29}  & 12.96  \\

\texttt{COX2-MD}& D & 9801.29  & 1.52  & \textbf{1.18}  & 132.11  & 23.93  & 3872.51  \\

\texttt{DD}& SS & 7730.73  & M & M & M & \textbf{392.03}  & M \\ 

\texttt{DHFR}& S & 17.27  & 22.72  & 30.14  & 118.15  & \textbf{1.24}  & 43.51  \\ 

\texttt{DHFR-MD}& D & 5078.20  & \textbf{1.62}  & 1.63  & 145.09  & 42.59  & 5970.10  \\ 

\texttt{ENZYMES}& S & 60.50  & 26.08  & 18.49  & 42.26  & \textbf{2.65}  & 147.04  \\

\texttt{ER-MD}& D & 5503.19  & \textbf{1.48}  & 2.78  & 119.88  & 31.80  & 5681.93  \\ 

\texttt{FRANKENSTEIN} & S & 43.33  & 262.12  & 714.15  & 2198.13  & \textbf{3.79}  & 1582.95  \\ 

\texttt{IMDB-BINARY}& D & 8637.97  & \textbf{4.19}  & 13.49  & 168.42  & 11.68  & 4152.41  \\ 

\texttt{IMDB-MULTI} & D & 12528.71  & \textbf{3.99}  & 20.94  & 388.92  & 12.59  & 6400.96  \\

\texttt{MSRC-9}& SS & 111.21  & 5.99  & 2.78  & 12.86  & \textbf{1.85}  & 31.72  \\ 

\texttt{MSRC-21}& SS & 1004.74  & 109.10  & 68.17  & 257.27  & \textbf{25.97}  & 1497.16  \\ 

\texttt{MSRC-21C}& SS & 104.52  & 6.48  & \textbf{1.30}  & 12.18  & 3.38  & 57.19  \\ 

\texttt{MUTAG}& S & 1.52  & 0.77  & 0.52  & 1.83  & \textbf{0.35}  & 2.64  \\ 

\texttt{NCI1}& S & 48.56  & 592.90  & 1675.78  & 4458.72  & \textbf{7.30}  & 934.91  \\ 

\texttt{NCI109}& S & 57.46  & 567.12  & 3309.95  & 4490.26  & \textbf{7.69}  & 994.78  \\ 

\texttt{PROTEINS}& S & 162.10  & 222.92  & 171.15  & 979.98  & \textbf{11.56}  & 648.14  \\ 

\texttt{REDDIT-BINARY}& S & M & M & M & M & \textbf{133.08}  & 9146.96  \\ 

\texttt{REDDIT-MULTI-5K}& S & M & M & M & M & \textbf{359.45} & M \\ 
\bottomrule
\end{tabularx}
\end{subtable}
\end{table}

\section{Additional Experimental Results on URN Datasets}\label{app:urn-detail}

\subsection{Construction of URN Datasets}\label{app:urn-construct}

\begin{table}[htbp]
    \centering
    \caption{Summary of URN datasets. In the table, $N$ is the number of graphs in a dataset; $R$ is global radius and $r$ is local radius, both measured in meters.}
    \small
    \begin{tabularx}{\linewidth}{Xp{3em}p{3em}p{3em}p{13em}X}
    \toprule
    Name    & $N$ &  $R$ & $r$ & Cities (Coordinates) & Patterns \\
    \midrule
    \texttt{URN2-1S}   & 400 & 1000 & 150 & Chicago: $(41.87, -87.63)$\newline Ufa: $(54.73, 55.95)$ & Gridiron\newline Chaotic \\

    \texttt{URN2-1M}  & 400 & 2000 & 300 &  Chicago: $(41.87, -87.63)$\newline Ufa: $(54.73, 55.95)$& Gridiron\linebreak Chaotic\\

    \texttt{URN2-2S} & 600 & 1000 & 150 & Wuhan: $(30.52, 114.35)$\newline London: $(51.50, -0.14)$ & Linear\linebreak Tributary \\

    \texttt{URN2-2M} & 600 & 2000 & 300 &Wuhan: $(30.52, 114.35)$\newline London: $(51.50, -0.14)$& Linear\linebreak Tributary\\

    \texttt{URN3-S}  & 600 & 1000 & 150 & Detroit: $(42.34, -83.05)$ \newline Omdurman: $(15.64, 32.48)$\newline Nanchang: $(28.68, 115.85)$ & Gridiron\linebreak Chaotic\linebreak Linear\\

    \texttt{URN3-M}  & 750 & 2000& 300 & Seattle: $(47.62, -122.35)$\newline Ufa: $(54.73, 55.95)$ \newline Tianjin: $(39.12, 117.17)$& Gridiron\linebreak Chaotic\linebreak Linear \\

    \texttt{URN4-S} & 800 & 1000 & 150 & Chicago: $(41.87, -87.63)$ \newline Omdurman: $(15.64, 32.48)$ \newline Shenyang: $(41.80, 123.43)$ \newline Paris: $(48.87, 2.29)$ & Gridiron\linebreak Chaotic\linebreak Linear\linebreak Tributary\\

    \texttt{URN4-M} & 800 & 2000 & 300 & Chicago: $(41.87, -87.63)$\newline Kano: $(12.01, 8.59)$ \newline Shenyang: $(41.80, 123.43)$ \newline Berlin: $(52.52, 13.40)$& Gridiron\linebreak Chaotic\linebreak Linear\linebreak Tributary \\

    \bottomrule
    \end{tabularx}
    \label{tab:urn-data}
\end{table}

\subsection{Confusion Matrices}\label{app:confusion}

\begin{figure}[H]
    \centering
     \begin{subfigure}{0.49\linewidth}
         \includegraphics[width=\linewidth]{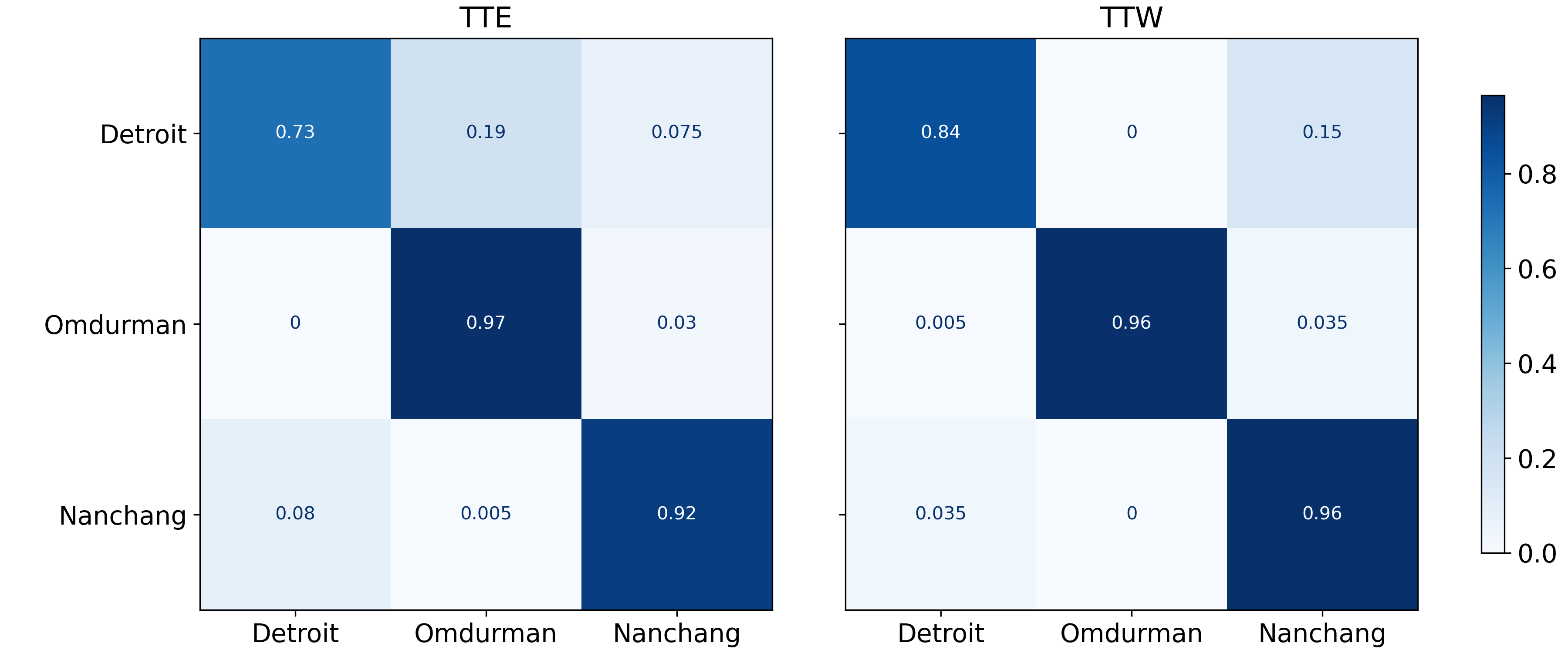}
         \caption{\texttt{URN3-S}}
     \end{subfigure}
    \begin{subfigure}{0.49\linewidth}
        \includegraphics[width=\linewidth]{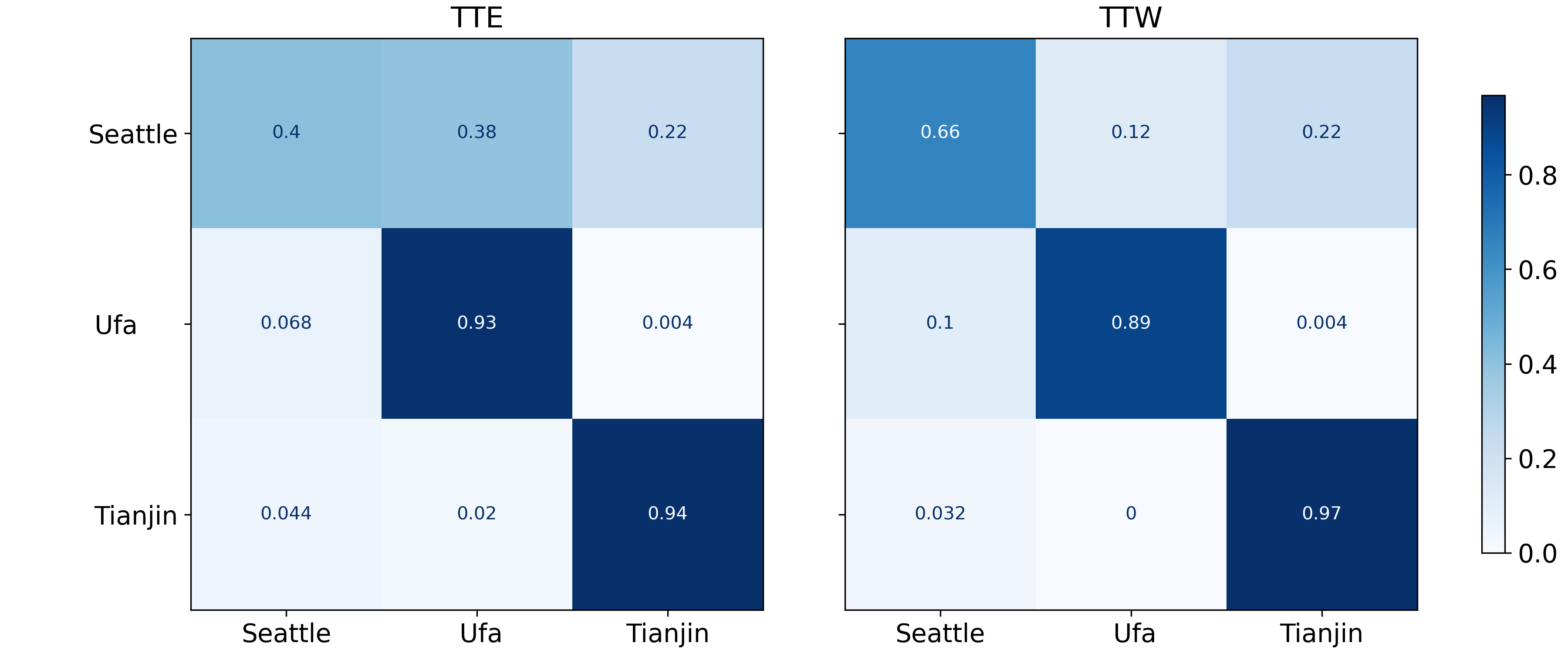}
        \caption{\texttt{URN3-M}}
    \end{subfigure}

    \begin{subfigure}{0.49\linewidth}
         \includegraphics[width=\linewidth]{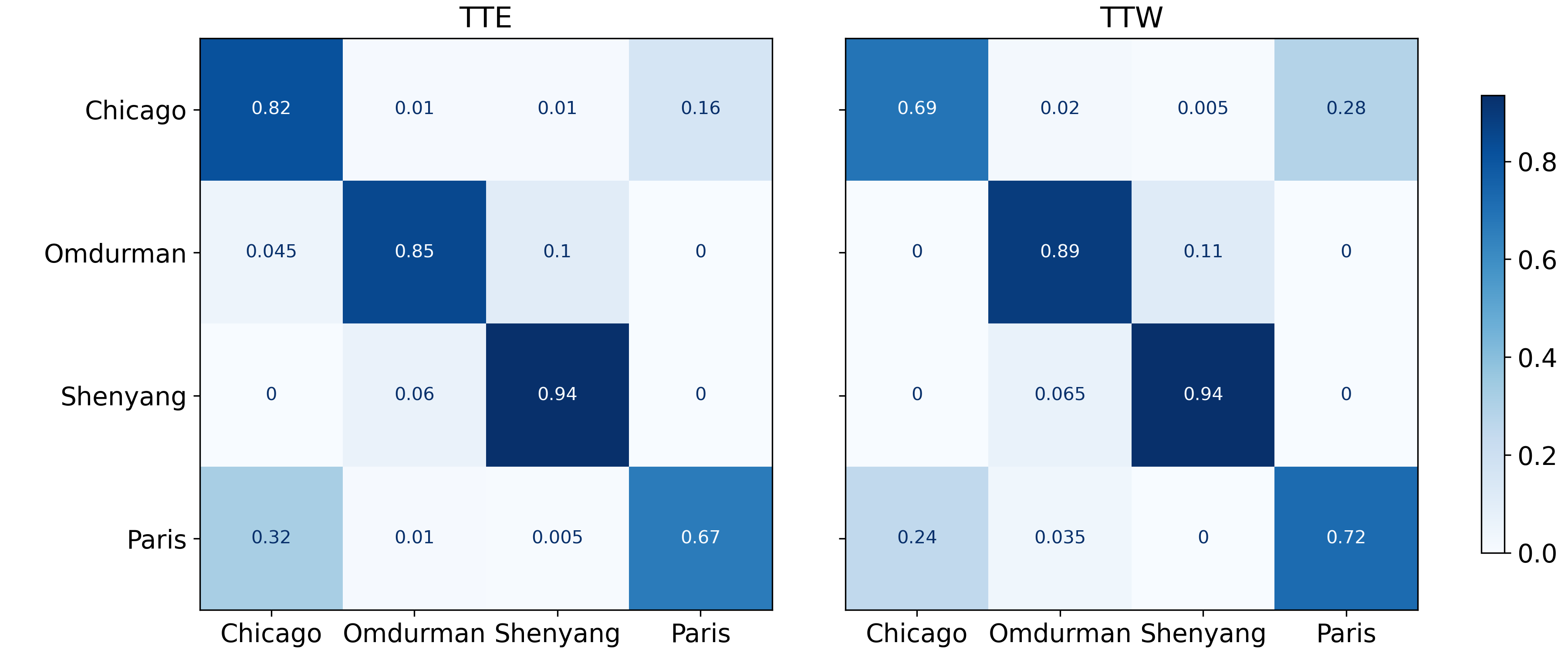}
         \caption{\texttt{URN4-S}}
     \end{subfigure}
    \begin{subfigure}{0.49\linewidth}
        \includegraphics[width=\linewidth]{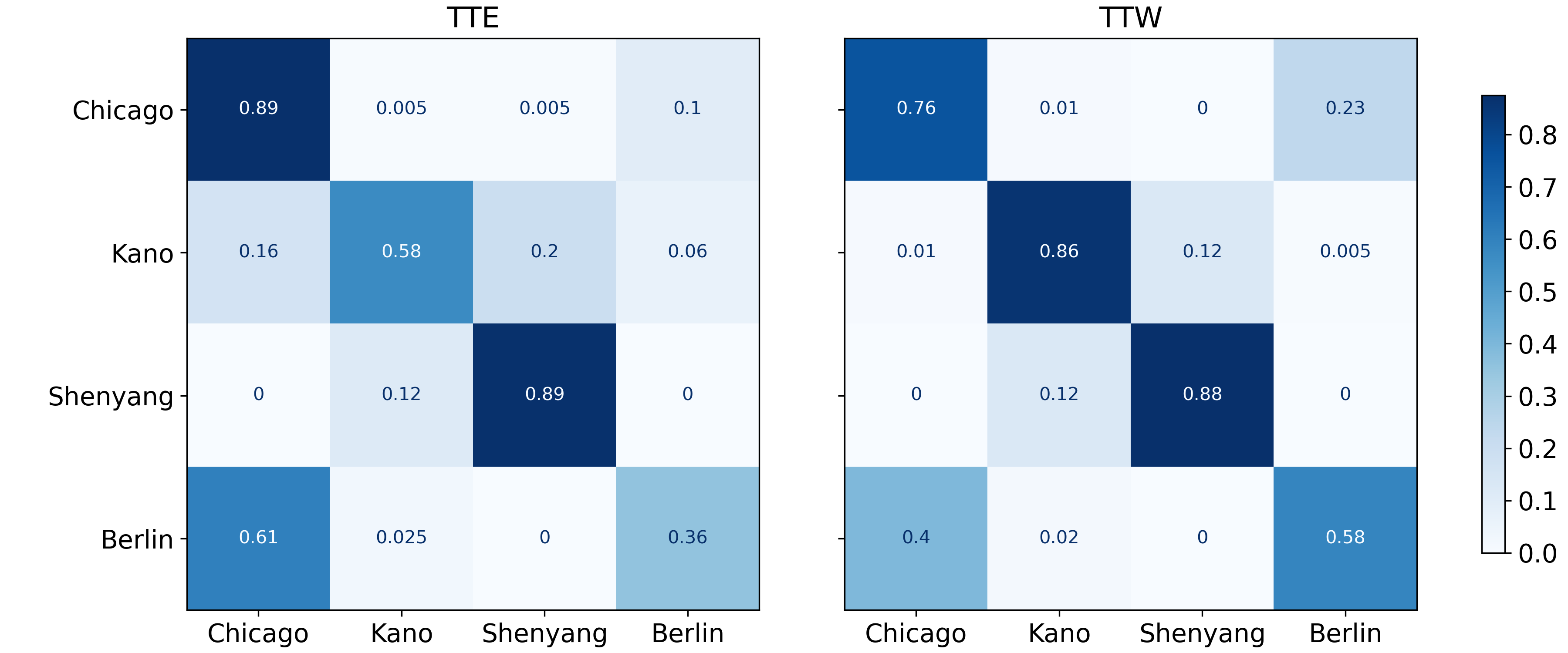}
        \caption{\texttt{URN4-M}}
    \end{subfigure}
        \caption{Confusion matrices of classification on \texttt{URN3-S} and \texttt{URN3-M}, \texttt{URN4-S} and \texttt{URN4-M}. The row indices are true labels and the column indices are predicted labels.}
    \label{fig:confusion-matrix-4}
\end{figure}

\newpage
\section{Additional Discussion on Limitations and Future Research}\label{app:discuss}

\textbf{``De-bridging'' a Graph.} The tropical Torelli map ignores bridge edges as they do not belong to cycles. More generally, this leads to the question of how to ``de-bridge" a graph $G$, i.e., modify $G$ to eliminate bridge edges in a reasonable way. Two potential criteria to consider would be to eliminate bridges incurring:  (1) a minimal change to its topology, for example, finding a new graph $G'$ whose persistent homology is similar to that of $G$ \citep{weinberger2011persistent}; or (2) a minimal change to its metric structure, for example, finding a $G'$ that is close to $G$ measured under the Gromov--Hausdorff distance \citep{memoli2012some}. Trees represent an extreme case of this problem, since all of their edges are bridges. Developing methods to de-bridge a graph while maintaining most of its original information is an important topic for future research.

\textbf{Further Real-World Examples.} Beyond road networks, we remark that our tropical Torelli framework for metric graphs are applicable to a wide range of domains where the geometry and topology of continuous structures are essential. Some examples include structural brain networks in neuroscience \citep{bullmore2009complex}; vascular networks (such as blood vessels) in biology \citep{scianna2013review}; and electric power grids in urban planning \citep{pagani2013power}. These are interesting application areas for future work.

\bigskip
\begin{table}[h]
    \centering
    \caption{Computational complexity for different graph kernels. In the table, \texttt{.fit} time is the time complexity to compute the features for all graphs in a dataset; \texttt{.transform} time is the time complexity to compute the full kernel matrix. $N$ is the number of graphs in a dataset.}
    \begin{tabularx}{\linewidth}{lXXX}
    \toprule
     Name & Parameters   & \texttt{.fit} Time & \texttt{.transform} Time \\
     \midrule
     EH & $m$: \# edges & $O(N\cdot m)$ &$O(N^2\cdot m)$\\
 
     GS & $B$: \# graphlet samples;\newline $G_k$: \# $k$-graphlets  & $O(N\cdot  B)$ & $O(N^2\cdot G_k)$\\

     OS & $n$: \# nodes; $d$: node degree;\newline $h$: subtree depth;\newline $T_h$: \# tree types of depth $h$ & $O(N\cdot nd^h)$ & $O(N^2\cdot T_h)$\\

     SP & $n$: \# nodes & $O(N\cdot n^2\log n)$ & $O(N^2\cdot n^2)$ (histogram)\newline $O(N^2\cdot n^4)$ (all pairs)\\

     WL & $m$: \# edges; $h$: \# iterations\newline
     $L$: \# distinct labels& $O(N\cdot hm)$ & $O(N^2\cdot L)$\\

     $k$CD & $n$: \# nodes; $m$: \# edges;\newline $C_k$: max core number & $O(N\cdot(n+m))$ & $O(N^2\cdot C_k)$\\

     TTE & $n$: \# nodes; $g$: genus & $O(N\cdot g^2n)$ & $O(N^2\cdot g^2)$\\

     TTW & $n$: \# nodes; $g$: genus & $O(N\cdot g^2n)$ & $O(N^2\cdot g^3)$\\
     \bottomrule
    \end{tabularx}
    \label{tab:complexity}
\end{table}


\end{document}